\documentclass[final,3p,times]{elsarticle}
\usepackage{graphicx}
\usepackage{amssymb}
\usepackage{amsmath}
\usepackage{amsthm}
\usepackage{stmaryrd}

\newtheorem{definition}{Assumption}
\usepackage{soul}
\usepackage{algorithm}
\usepackage[noend]{algpseudocode}
\usepackage{subfig}
\usepackage{booktabs}

\usepackage[pagewise]{lineno}

\usepackage{bm}

\newcommand{\set}[1]{\left\{#1\right\}}

\newcommand{\expnumber}[2]{{#1}\mathrm{E}{#2}}
\newcommand{\kernel}{\mathcal{K}}

\newcommand{\pspace}{\Lambda}
\newcommand{\qspace}{\mathcal{Q}}
\newcommand{\noisydata}{\mathcal{D}}
\newcommand{\filteredspace}{\mathcal{F}(\noisydata)}
\newcommand{\pmeas}{\mu_{\pspace}}

\newcommand{\pborel}{\mathcal{B}_{\pspace}}
\newcommand{\qborel}{\mathcal{B}_{\qspace}}

\newcommand{\preddens}{\pi_{\text{pred}}}
\newcommand{\obsdens}{\pi_{\text{obs}}}
\newcommand{\initdens}{\pi_{\text{init}}}
\newcommand{\updens}{\pi_{\text{update}}}

\newcommand{\predmeas}{\mathbb{P}_{\text{pred}}}
\newcommand{\obsmeas}{\mathbb{P}_{\text{obs}}}
\newcommand{\initmeas}{\mathbb{P}_{\text{init}}}
\newcommand{\upmeas}{\mathbb{P}_{\text{update}}}

\graphicspath{{PDF/}{Figures/}}

\newtheorem{remark}{Remark}

\def\w{\omega}
\def\k{\kappa}
\def\b{\beta}
\def\r{\rho}

\def\b{\beta}

\def\A{\mathcal{A}}
\def\Q{\mathcal{Q}}
\def\E{\mathcal{E}}

\def\D{\partial}
\def\u{\mathbf{u}}

\def\v{\mathbf{v}}

\def\.{\cdot}

\def\x{\mathbf{x}}

\newcommand{\ub}{\mathbf{u}}
\begin{document} 
\begin{frontmatter}

%% Title, authors and addresses

%% use the tnoteref command within \title for footnotes;
%% use the tnotetext command for the associated footnote;
%% use the fnref command within \author or \address for footnotes;
%% use the fntext command for the associated footnote;
%% use the corref command within \author for corresponding author footnotes;
%% use the cortext command for the associated footnote;
%% use the ead command for the email address,
%% and the form \ead[url] for the home page:
%%
%% \title{Title\tnoteref{label1}}
%% \tnotetext[label1]{}
%% \author{Name\corref{cor1}\fnref{label2}}
%% \ead{email address}
%% \ead[url]{home page}
%% \fntext[label2]{}
%% \cortext[cor1]{}
%% \address{Address\fnref{label3}}
%% \fntext[label3]{}

\title{From Displacements to Distributions: A Machine-Learning Enabled Framework for Quantifying Uncertainties in Parameters of Computational Models}

%% use optional labels to link authors explicitly to addresses:
%% \author[label1,label2]{<author name>}
%% \address[label1]{<address>}
%% \address[label2]{<address>}
\author[tr]{Taylor Roper}
\address[tr]{
University of Colorado Denver\\
Department of Mathematical and Statistical Sciences\\
1201 Larimer St\\
Denver, CO, 80204, USA
}
\ead{taylor.roper@ucdenver.edu}

\author[hh]{Harri Hakula}

\address[hh]{Aalto University\\
Department of Mathematics and System Analysis\\
P.O. Box 11100\\
FI--00076 Aalto, Finland
}
\ead{Harri.Hakula@aalto.fi}

\author[tb]{Troy Butler}
\address[tb]{
University of Colorado Denver\\
Department of Mathematical and Statistical Sciences\\
1201 Larimer St\\
Denver, CO, 80204, USA
}
\ead{troy.butler@ucdenver.edu}

\begin{abstract}
This work presents novel extensions for combining two frameworks for quantifying both aleatoric (i.e., irreducible) and epistemic (i.e., reducible) sources of uncertainties in the modeling of engineered systems.
The data-consistent (DC) framework poses an inverse problem and solution for quantifying aleatoric uncertainties in terms of pullback and push-forward measures for a given Quantity of Interest (QoI) map. 
Unfortunately, a pre-specified QoI map is not always available a priori to the collection of data associated with system outputs.
The data themselves are often polluted with measurement errors (i.e., epistemic uncertainties), which complicates the process of specifying a useful QoI.
The Learning Uncertain Quantities (LUQ) framework defines a formal three-step machine-learning enabled process for transforming noisy datasets into samples of a learned QoI map to enable DC-based inversion.
We develop a robust filtering step in LUQ that can learn the most useful quantitative information present in spatio-temporal datasets.
The learned QoI map transforms simulated and observed datasets into distributions to perform DC-based inversion.
We also develop a DC-based inversion scheme that iterates over time as new spatial datasets are obtained and utilizes quantitative diagnostics to identify both the quality and impact of inversion at each iteration.
Reproducing Kernel Hilbert Space theory is leveraged to mathematically analyze the learned QoI map and develop a quantitative sufficiency test for evaluating the filtered data.
An illustrative example is utilized throughout while the final two examples involve the manufacturing of shells of revolution to demonstrate various aspects of the presented frameworks. 
\end{abstract}

\begin{keyword}
inverse problems \sep uncertainty quantification \sep quantity of interest \sep data-consistent \sep push-forward measure \sep pullback measure
%% keywords here, in the form: keyword \sep keyword

%% MSC codes here, in the form: \MSC code \sep code
%% or \MSC[2008] code \sep code (2000 is the default)
\MSC 28A50 \sep  60-04 \sep 60-08
\end{keyword}

\end{frontmatter}

%%
%% Start line numbering here if you want
%%
%\linenumbers

\section{Introduction}
\label{sec:introduction}

Thin structures and shells in particular are common in engineering \cite{CB:Book}.
Despite their ubiquity, they remain challenging from the point-of-view of
computational mechanics. 
One of the reasons for this is their sensitivity
to perturbations in materials, kinematic constraints, and shell geometry itself 
due to manufacturing imperfections.

In the context of uncertainty analysis for structural design, \cite{BAI2016615} states that ``[a]ll variables in engineering structures are stochastic to a certain degree'' and that model uncertainty is ``due to imperfections and idealizations made in physical model formulations [$\ldots$] as well as in the choices of probability distribution types for the representation of uncertainties.''
The authors go on to present a method for uncertainty quantification (UQ) as a probabilistic quantity where one makes ``many'' observations and corresponding predictions of a particular scalar quantity of interest, forming the ratio of observations to predictions as a measure of discrepancy, and utilizing both the mean and variance of this ratio to perform the UQ analysis. 
While we consider a distinct measure-theoretic UQ framework, the perspective provided in \cite{BAI2016615} nonetheless proves prescient in this work.
At a high-level, we present novel developments within a machine learning-enabled data-to-distribution pipeline to transform (noisy) observed and predicted spatio-temporal data clouds into samples of (vector-valued) quantities of interest (QoI) that enables the construction of a data-consistent update to initially assumed distributions on model parameters via the ratio of (non-parametric) observed and predicted distributions of the QoI. 
% The rest of this introduction is organized and designed to be primarily conceptual and sufficiently self-contained so that readers who wish to skip mathematical details are able to go directly from the end of this introduction to the numerical examples.

In general, any UQ analysis for a computational model of engineered or physical system governed by principles of mechanics seeks to use principles of probability theory to identify, classify, and quantify sources of uncertainty between simulated predictions and observational data, e.g., see
\cite{ROY20112131} for a comprehensive framework analyzing the impacts of aleatoric (i.e., irreducible) and epistemic (i.e., reducible) sources of uncertainty on model predictions. 
To provide some context, suppose a manufacturing process is implemented to produce an ``ideal'' design of some engineered system for which uncertainty is inherent and irreducible in system inputs, e.g., structural design \cite{BAI2016615}, material science \cite{tran2021solving}, or thin elastic membranes \cite{BH20}.
Some common sources of input uncertainty in such systems include, but are not limited to, material impurities, e.g., see \cite{C3EE41328D, LIU2013143, Lu1987}, and mechanical tolerances in the machinery used to construct system components and sub-systems, e.g., see \cite{RAMESH20001235, RAMESH20001257, 10.1115/1.4005790}.
Complicating matters further is that the system inputs impacted by these uncertainties often define physical characteristics that have to be parameterized within the model and are hidden from direct observation. 

\begin{figure}[htbp]
\centering
\subfloat[]
{\label{fig:trommelA}\includegraphics[width=0.4\textwidth]{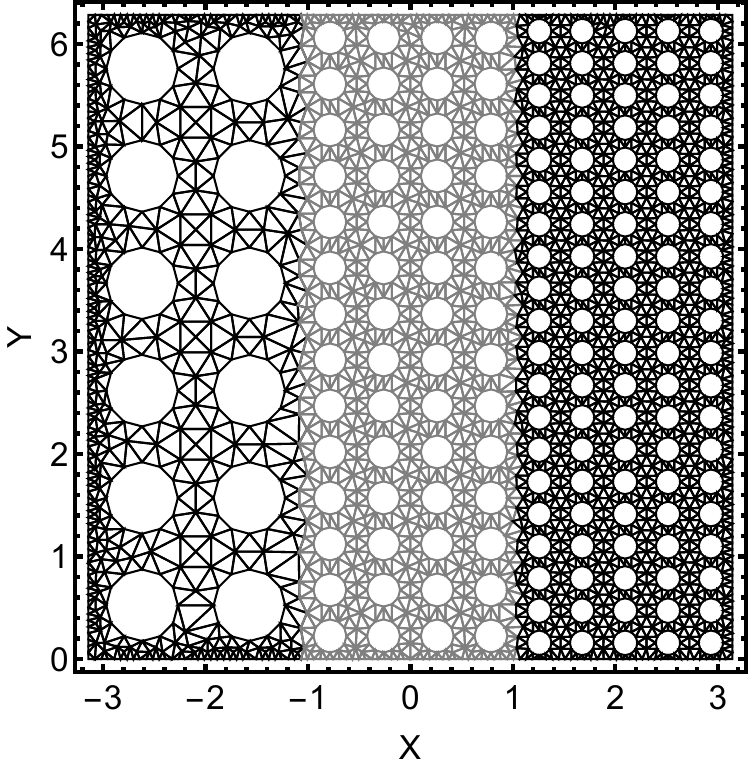}}
\quad
\subfloat[]
{\label{fig:trommelB}\includegraphics[width=0.4\textwidth]{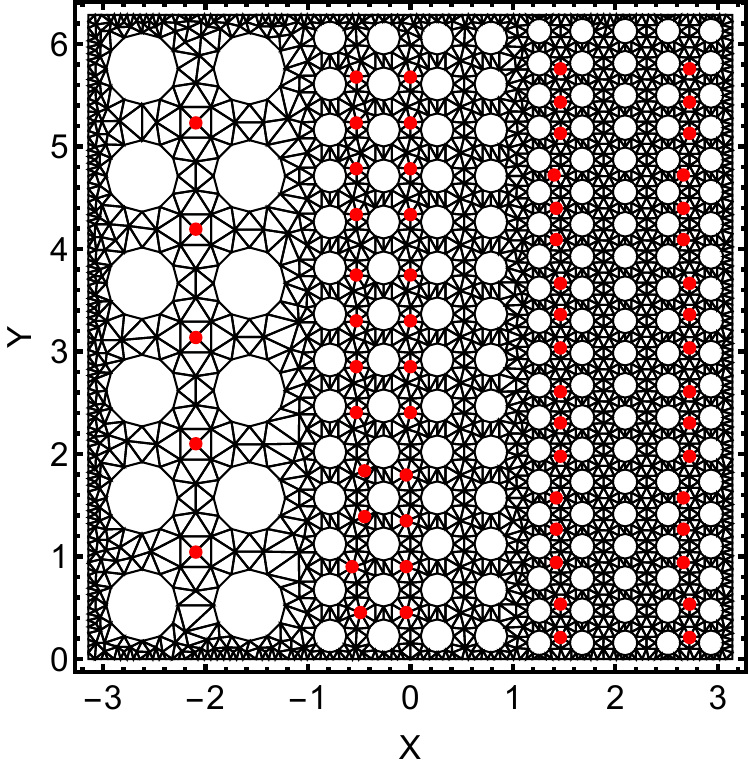}}
\caption{Trommel configuration, hole coverage 35\%. (a) Three different sections are indicated
with different graylevels. (b) The sensor locations are indicated with red colour.
The boundaries at $x = \pm \pi$ are clamped and $y=0$ and $y=2\pi$
are periodic.
}\label{fig:trommel}
\end{figure}
As an example, consider Young's modulus, which is a common parameter appearing in models of elastic materials that is a measure of material stiffness whose value is often inferred by exerting a known amount of force to the material and measuring its deformation, e.g., see \cite{KOSTIC2022100998}. 
Similar to such experimental settings, the impact of perturbing model parameters is usually observed indirectly via the simulation of model state data.
It follows that the UQ analysis requires the formulation and solution of a stochastic inverse problem utilizing discrepancies between (noisy) observational and simulated data.
One example considered in this work involves the construction and monitoring of trommel screens, which are mechanical screening machines often used to separate materials of distinct sizes in mineral and solid-waste applications \cite{Glaub1982DesignAU}. 
Due to harsh operating conditions, the structural health of a population of screens (defined by the distribution of Young's modulus) should be periodically analyzed.
In this work, we consider trommel screens with three sections that possess distinctive regular penetration patterns but constant 35\% hole coverage, see Figure~\ref{fig:trommelA} for the computational mesh utilized in simulations with shading indicating the distinct sections.
Each section possess a unique Young's modulus, which must be inferred from a combination of observed and simulated data at sensor locations indicated by the red circles shown in Figure~\ref{fig:trommelB}.

\subsection{Uncertainty Quantification and Stochastic Inverse Problems}

The type of stochastic inverse problem that is formulated and the UQ methods developed for solving such a problem are dictated by the assumptions made about the type of uncertainty that is to be quantified. 
The last few decades have led to significant advancements in UQ research and thousands of papers published across many computational science journals as well as several dedicated UQ journals. 
It is therefore impossible to provide a complete literature review of UQ and the types of inverse problems at the heart of many UQ methodologies.
Below, we provide a brief literature review that is intentionally chosen to (i) focus on the two types of uncertainties (epistemic and aleatoric) considered (often separately) in the literature that are at the heart of this work, and (ii) to help situate the specific contributions of this manuscript (summarized in Section~\ref{subsec:intro-contributions}) within the vast UQ literature on these topics.

In cases where the inverse problem arises from an initial assumption of epistemic uncertainty in data and parameters, e.g., assuming noisy data are observed for a single instance of a system associated with true, but unknown, parameter values, the Bayesian framework is often deployed, e.g., see \cite{Kapteyn2021, CKS14, Gelman2013, CDS10, KO2001, BMP+1994, Fitzpatrick1991}.
The solution to the resulting inverse problem within the Bayesian framework is known as a posterior, which is a conditional density defined by the product of a prior density on parameters and a data-likelihood function that is usually constructed from the differences in simulated and observed QoI data. 
The posterior is consequently interpreted as defining the relative likelihoods that any particular estimate for the parameters could have produced all of the observed (noisy) data.
A commonly used parameter estimate utilized in the literature is the maximum a posteriori (MAP) estimate, e.g., see \cite{Burger_2014, PMS+14, APS+16}.
Under a typical Bayesian setup and assumptions (such as an additive noise model on the data that follows a given distribution, usually assumed to be Gaussian), the Bernstein-von Mises theorem~\cite{AsymptoticStats1998} guarantees that the posterior becomes more ``spiked'' around the true parameter value and subsequently that the resulting uncertainty in a parameter estimate, such as the MAP point, is reduced.

In cases where the inverse problem arises from an initial assumption of aleatoric uncertainty in data and parameters, e.g., assuming natural variability in system outputs and inputs as discussed above in the context of an engineered system, the inferrential target is fundamentally a distribution on parameters that quantifies this intrinsic and irreducible variability.
Alternative Bayesian formulations exist to address such problems such as hierarchical Bayesian methods~\cite{Wikle1998} that begin with specifying prior distributions from a parametric family (e.g., normal distributions).
Some non-parametric, hierarchical Bayesian methods for modeling aleatoric uncertainty also exist, e.g., see \cite{Gelfand2002ACA, Teh2010}.
These are an active research area within the machine/statistical learning community, but they are currently more expensive and complex to implement than what we utilize in this work.

In this manuscript, we utilize a measure-theoretic data-consistent (DC) framework. In the DC framework, the inverse problem and its solution are defined in terms of pullback and push-forward measures \cite{BET+14, BJW18a, BBE24}. 
The DC solution has what is referred to as the data-consistency property in that its push-forward through the QoI map matches the observed probability measure. 
The density-based approximation of the DC solution, as derived in \cite{BJW18a} via the Disintegration Theorem \cite{Chang_Pollard}, has seen the most development, analysis, and application in recent years, e.g., see~\cite{Saghafi2023.04.18.537149, ZM2023, rumbell2023novel, rumbell2023sequential,  BWZ22, tran2021solving, BGW2020}.
It is worth noting that the density form of the solution perhaps first appeared in \cite{PR2000} where it was derived through heuristic arguments based on logarithmic pooling and referred to as ``Bayesian melding.'' 
The fundamental distinction in assumptions, form, and properties of the solution from the typical Bayesian framework led to a distinction of the terminology used in the DC framework in \cite{BJW18b} (which is a follow-up to \cite{BJW18a}).
In \cite{BJW18b} and many of the works that chronologically follow it, an initial and predicted density are used to describe the initial quantification of uncertainties on parameters and QoI, respectively, independent of any observed data.
The observed density describes the quantification of uncertainty for the observed QoI data.
An update to the initial density is then obtained via the product of the initial density with the ratio of observed to predicted densities evaluated on the outputs of the QoI map. 
The updated density serves as the DC solution.
Recently, \cite{PdCY+23} extended the DC framework to problems that focus on epistemic uncertainties and parameter estimation while \cite{BYW20, MSB+22} extended the DC framework to problems that simultaneously involve both aleatoric and epistemic uncertainties.
The work of \cite{MSB+22} is most closely related to the current work, which is discussed more below in the context of the primary contributions of this manuscript.

\subsection{Mathematical and Algorithmic Contributions}\label{subsec:intro-contributions}

% To properly frame the contributions of this work, we begin with a summary of challenges not yet addressed in the existing literature

The terms data and information are sometimes used interchangeably, but they are in fact distinct.
The distinction is critical in this work, and we refer to the commonly accepted definitions that data is a collection of facts whereas information is the organization and interpretation of those facts.
At a high-level, the interconnected goals of this work are as follows: (i) develop a robust framework that can learn the most useful quantitative information present in spatio-temporal data represented as a QoI map, (ii) utilize the learned QoI map to transform simulated and observed data into the predicted and observed distributions for DC-based inversion, and (iii) mathematically represent and analyze this learned QoI map to develop quantitative diagnostics that evaluate critical aspects of the framework. 

When data are strictly temporal in nature, goals (i) and (ii) are achieved with the Learning Uncertain Quantities (or LUQ\footnote{Pronounced as ``luck.''} for short) framework introduced in \cite{MSB+22}, which we briefly summarize here to help situate the contributions of this work.
The basic motivation behind LUQ is that individual measurement devices can easily produce $\mathcal{O}(10^4)$ (or more) time steps for which data are collected while the dynamical behavior is in fact characterized by some ``low-dimensional'' features. 
In other words, organizing and interpreting the data in terms of these low-dimensional features defines the useful information while the quantitative characterization of this information defines the QoI map.
This addresses goal (i).
Applying this QoI map to the data defines a transformation of samples of model parameters into numerical samples of the QoI for which (predicted and observed) distributions are estimable. 
This addresses goal (ii).
LUQ proceeds in three steps: (Step 1) filtering, (Step 2) clustering and classification, and (Step 3) feature extraction.
The filtering step utilizes predicted and observed data (both of which may include noise) to approximate the underlying temporal signal, which is then sampled to produce filtered data, i.e., the output of this step is a temporal dataset where epistemic uncertainty is reduced.
The clustering and classification step is an optional step.
It is useful in cases where a dynamical system presents strikingly different qualitative behaviors as model parameters are varied (e.g., due to bifurcations) or over different windows of time (e.g., due to transient or equilibrium behavior). 
In such cases, unsupervised learning approaches are first applied to classify (i.e., label) the different types of dynamics that are present in the predicted filtered data.
This is followed by the training and optimal selection of classifiers for this labeled data so that subsequently filtered observed temporal data are appropriately labeled by their prevalent dynamics. 
This leads to the final step of feature extraction that applies kernel-based principal component analysis (kPCA) \cite{pearson1901liii,scholkopf1997kernel,mika1999kernel,abdi2010principal}
to define, implicitly, the QoI maps that best describe the low-dimensional nature of the dynamics.

The main contributions of this work are both mathematical and algorithmic in nature with algorithmic developments encoded within the open-source software package \texttt{LUQ}~\cite{pyLUQ2.0.0}. 
The \texttt{LUQ} package \cite{pyLUQ2.0.0} underwent a complete refactoring to provide (a) an agnostic interface in terms of spatial or temporal data types and (b) advanced filtering options that apply to higher-dimensional data sets.
The refactored package also allows multiple points-of-entry within the workflow including offline construction of LUQ objects from pre-computed predicted data sets that can be subsequently applied to observed data sets as they are made available.
At a conceptual level, the new LUQ framework and software significantly extend prior work to apply to both spatial and spatio-temporal data. 
At a more specific algorithmic level, the filtering step in the prior LUQ framework only applied to temporal data associated with observing the evolution of a single state variable in a dynamical system since it utilized adaptively constructed splines to approximate the underlying temporal signal.
As discussed in \cite{MSB+22}, the choice of splines in the filtering step is motivated by a type of universal approximation theory for splines (see \cite{UFW17}) where it is possible to approximate the underlying dynamical response (assuming a particular regularity or finite number of discontinuities) to arbitrary pointwise accuracy if both a sufficiently high frequency for collecting data and number of knots are used. 
However, the use of splines prevents any practical application to spatial data in 2- or higher dimensions.
Motivated by the universal approximation theorem for neural nets (NNs) \cite{HORNIK1989359}, we provide a deep learning-inspired approach to filtering that utilizes radial basis functions (RBFs) in the hidden layer \cite{Jiang2022} which allows for direct implementation in arbitrary (but finite) dimension.

The contributions listed above address goals (i) and (ii).
To address goal (iii), we provide an explicit derivation and mathematical justification of the form of learned QoI maps obtained in the feature extraction step.
This utilizes theoretical results from reproducing kernel Hilbert spaces (RKHS) \cite{Aronszajn1950}.
This provides significant insight into the mathematical structure of the QoI map, which is then utilized to develop a quantitative scheme for numerically determining the sufficiency of filtered data produced in the first step of the LUQ framework. 

\subsection{Numerical Examples}

The main contributions of this work are contextualized in several numerical examples. 
First, to help illustrate various DC and LUQ concepts as they are presented throughout the text, we utilize a simple 2-dimensional wave model for a fluid where waves are excited by randomly located droplets.
The shape of the droplets and the choice of boundary conditions results in complicated wave shapes in space and time.
This serves to highlight the capabilities of the new filtering within \texttt{LUQ} as we seek QoI and DC solutions that attempt to recover the data-generating distribution of droplet locations.

The two main numerical examples are presented in Section~\ref{sec:numerical_experiments}.
These concern shells of revolution, have several engineering applications, and also serve to highlight various new and improved capabilities of the \texttt{LUQ} software package.  
Shells of revolution are 3D structures, but under assumptions on symmetries and properties of
loading, the problems can be reduced dimensionally to 2D and even 1D problems.
Consider a cylinder with constant radius $R = 1$ and thickness $d = 1/100$, say.
Many properties of the solutions of the shell problems, for instance, the
resulting displacement fields can be described as functions of the dimensionless thickness $t = d/R$. 
Under the assumptions of linear elasticity, the Young's modulus $E$, if isotropic over the whole structure, simply acts as the scaling parameter to the displacement field.

The first example in Section~\ref{sec:numerical_experiments} involves the manufacturing of approximately cylindrical shells subjected to persistent loading in the transverse direction as would occur, for instance, when such shells are utilized in the construction of wind turbine towers to be deployed in areas with strong directional prevailing winds.  
In this example, it is assumed that imperfections in materials impacts the Young's modulus and imperfections in the machining impacts the shape characteristics of each shell.
This example demonstrates how we can  utilize noisy deformation data to update distributions on Young's modulus and shape characteristics, which is useful in monitoring the overall structural integrity of the population of these shells.

The second example of Section~\ref{sec:numerical_experiments} involves the construction and monitoring of trommel screens as previously mentioned.
This example demonstrates how we can utilize spatial data obtained by different experiments at distinct times with distinct loads to sequentially learn QoI (of potentially distinct dimension) to learn the distributions of the Young's modulus associated with each of the distinctive spatial regions of the trommels.  

\subsection{Organization}

The rest of this paper is organized as follows.
Section~\ref{sec:DCI} provides a brief overview of the mathematics behind DC-based inversion including a critical quantitative diagnostic for verifying a key predictability assumption.
Section~\ref{sec:LUQ} summarizes the prior work in LUQ for temporal data.
Section~\ref{sec:LUQ_extensions} describes the extension of the LUQ framework to datasets with spatial attributes that require filtering.
Spatio-temporal datasets are then discussed in Section~\ref{sec:DCI_iterative} in the context of an iterative approach to DCI that utilizes the new spatial-filtering options within LUQ at each temporal iteration.
The mathematical analysis of learned QoI is presented in Section~\ref{sec:QoI_kernels}, which is subsequently utilized to develop a sufficiency diagnostic for filtered data.
The main numerical examples follow in Section~\ref{sec:numerical_experiments}.
Concluding remarks and future research directions are provided in Section~\ref{sec:conclusions}.
The appendices contain information for accessing all software and datasets utilized in this work as well as mathematical information related to the main numerical examples.

\section{Data-Consistent Inversion: Overview}
\label{sec:DCI}

We begin by introducing the minimal notation and terminology required for defining the stochastic inverse problem and its solution in the DC framework considered in this work.
Let $\pspace$ denote an uncertain model parameter taking values in the parameter space denoted by $\pspace\subseteq\mathbb{R}^p$.
Let $Q:\pspace\to\qspace\subseteq\mathbb{R}^q$ denote a quantities of interest (QoI) map from the parameter space to the space of observable QoI data associated with model outputs, which is denoted by $\qspace:=Q\left( \pspace\right)$.
Given an observed probability measure $\obsmeas$ on $\qspace$, the goal of Data-Consistent Inversion (DCI) is to compute a probability measure $\mathbb{P}_\pspace$ on $\pspace$ such that the push-forward of this measure through $Q$ matches $\obsmeas$. 
More formally, if $\mathcal{B}_\pspace$ and $\mathcal{B}_\qspace$ denote the Borel $\sigma$-algebras on $\pspace$ and $\qspace$, respectively, and $\obsmeas$ is defined on $\left( \qspace,\mathcal{B}_\qspace\right)$, then the DCI solution is given by a probability measure $\mathbb{P}_\pspace$ on $\left( \Lambda, \mathcal{B}_{\Lambda}\right)$ such that 
\begin{equation}\label{eq:dc_property_measure}
    \mathbb{P}_\pspace\left( Q^{-1}\left( E\right)\right) = \obsmeas\left( E\right) \: \: \forall E\in \mathcal{B}_\qspace, 
\end{equation} 
where $Q^{-1}$ is the measure-theoretic shorthand for the pre-image of $Q$, i.e., the map $Q$ is {\em not} assumed to be invertible. 
This introduces an immediate complication since it is possible (and in fact typical) for many DCI solutions to exist. 
To address this issue, we first note it is self-evident from~\eqref{eq:dc_property_measure} that DCI solutions are uniquely defined on the induced ``contour'' $\sigma$-algebra $\mathcal{C}_\pspace$ defined as $\mathcal{C}_\pspace=\left\{ Q^{-1}\left( E\right) \: | \: E\in \qborel\right\}$. 
For any measurable QoI map, $\mathcal{C}_\pspace\subseteq\pborel$.
However, the containment is usually proper since $\pborel$ often contains events not present in $\mathcal{C}_\pspace$. 
Since any event in $\pborel$ can be written as an intersection with the smallest contour event containing it\footnote{This can be defined in the usual measure-theoretic way of defining smallest sets in a $\sigma$-algebra with a specific property by taking the intersection of all such sets from the $\sigma$-algebra with the specific property.}, specifying different solutions to the DCI problem depends upon specifying different probabilities of $\pborel$-measurable subsets of events in $\mathcal{C}_{\Lambda}$. 

\subsection{A density-based solution}\label{sec:computing_solution}

% given below. 

% \begin{theorem}[Disintegration Theorem] Given a probability measure $P_{\Lambda}$ on $\left( \Lambda,\mathcal{B}_{\Lambda}\right)$ along with corresponding push-forward measure $P_{\mathcal{D}}$ on $\left( \mathcal{D},\mathcal{B}_{\mathcal{D}}\right)$, there exists a $P_{\mathcal{D}}$-almost everywhere uniquely defined family of conditional probability measures $\left\{ P_d\right\}_{d\in \mathcal{D}}$ on $\Lambda$ such that $$P_d\left( A\cap Q^{-1}\left( d\right)\right)=P_d\left( A\right) \: \forall \: A\in \mathcal{B}_{\Lambda}$$ and there exists the following disintegration of $P_{\Lambda}$, $$P_{\Lambda}\left( A\right)=\int_{\mathcal{D}}\int_{A\cap Q^{-1}\left( d\right)}dP_d\left( \lambda\right)dP_{\mathcal{D}}\left( d\right) \: \forall A\in \mathcal{B}_{\Lambda}.$$ 
% \end{theorem}

% We first specify an ansatz probability measure on $\Lambda$, usually a uniform distribution given by a scaled Lebesgue measure, called the initial distribution $P_{\text{init}}$. The only time a distribution other than a uniform one should be chosen is if information about variation in $\Lambda$ is already known (or perhaps learned) since the choice of $P_{\text{init}}$ will effect the DCI solution in directions uninformed by the data. The support of the initial distribution must contain the support of the true distribution of parameters. This along with the assumption that all observable measurements are contained within $Q\left( \Lambda\right)$ can be summarized in the following assumption.

In \cite{BJW18a}, a density-based approach to solving DCI is introduced. 
We summarize this approach using the terminology from \cite{BJW18b} as mentioned in the introduction.
This approach utilizes a disintegration theorem \cite{Chang_Pollard} and an initial density on $\pspace$, denoted by $\initdens$, to construct a specific solution from the space of all DCI solutions as
\begin{equation}\label{eq:dc_density}
    \updens(\lambda) :=\initdens(\lambda) \frac{\obsdens(Q(\lambda))}{\preddens(Q(\lambda))}. 
\end{equation}
Here, $\preddens$ denotes the \emph{predicted density} defined as the push-forward of $\initdens$ through the map $Q$ and $\obsdens$ denotes the density associated with $\obsmeas$. 
Note that in more general cases, specifying dominating measures on $(\pspace, \pborel)$ and $(\qspace, \qborel)$ admits the same form of solution but with the densities interpreted as Radon-Nikodym derivatives.
The solution, $\updens$, is referred to as an updated density (or simply as the update) because it serves to update $\initdens$ only in directions informed by the QoI map.
This is easily verified by specifying a fixed QoI value, $\bm{q}\in\qspace$, and restricting $\lambda$ to the ``contour'' defined by $Q^{-1}(\bm{q})$. 
For all such $\lambda$, $Q(\lambda)=\bm{q}$ and the ratio of observed to predicted densities in~\eqref{eq:dc_density} is a constant.
It follows that the initial likelihoods of the contours are updated while the initial conditional densities along the contours remain unchanged in the update.
To emphasize the role of the ratio of observed to predicted densities in updating the initial, we rewrite~\eqref{eq:dc_density} as
\begin{equation}\label{eq:dc_density_r}
    \updens(\lambda) = r(Q(\lambda)) \, \initdens(\lambda), \ r(Q(\lambda)):=\frac{\obsdens(Q(\lambda))}{\preddens(Q(\lambda))}.
\end{equation}

As shown in \cite{BJW18a}, $\updens$ is unique (up to the choice of $\initdens$), and is also stable in the total-variation metric with respect to perturbations in $\initdens$, $\obsdens$, or $\preddens$.
The last stability result is critical since even if $\initdens$ and $\obsdens$ are specified exactly, we must often estimate $\preddens$ in constructing an estimate of $\updens$.
In cases where the QoI map is itself approximated as part of the process, \cite{BJW18b, BWZ22} provides convergence analysis for the sequence of approximate updates. 

\subsection{Numerical approximation, predictability assumption, and a quantitative diagnostic}\label{sec:diagnostic}

It is often the case that $Q$ is nonlinear or there is a change in dimensions between $\pspace$ and $\qspace$.
In such cases, we expect $\preddens$ and thus $\updens$ to both be non-parametric even if $\initdens$ and $\obsdens$ are given exactly from some well-studied parametric families of distributions (e.g., as Gaussian distributions). 
In this work, for the sake of both simplicity and reproducibility of results, we estimate $\preddens$ with standard kernel density estimation (KDE) \cite{Devroye1985} techniques.
This requires us to address a key assumption involving $\preddens$ that allows for practical approximation and generation of random samples with standard techniques\footnote{By standard techniques, we refer to ``off the shelf'' approaches such as kernel density estimation and rejection sampling as utilized in \cite{BJW18a, BJW18b}.}.
In \cite{BJW18a}, this is referred to as the {\em predictability assumption} given by 
\begin{definition}\label{def:predictability} There exists a constant $C>0$ such that $\obsdens(\bm{q})\leq C \preddens(\bm{q}) \:\: \forall \bm{q}\in \qspace$.
\end{definition}

Under this predictability assumption, $\updens$ defined in~\eqref{eq:dc_density} or~\eqref{eq:dc_density_r} is a density. 
It follows that
\begin{equation}\label{eq:diagnostic}
    \mathbb{E}_\text{init}(r(Q(\lambda)) = \int_\pspace r(Q(\lambda))\, d\initmeas = \int_\pspace \initdens(\lambda)r(Q(\lambda))\, d\pmeas = \int_\pspace \updens(\lambda)\, d\pmeas = 1,
\end{equation}
where $\initmeas$ is the initial probability measure associated with the density $\initdens$ and $\pmeas$ is the Lebesgue measure\footnote{In the case of more general Radon-Nikodym derivatives, $\pmeas$ is a specified dominating measure on $\pspace$.} on $\pspace$.
We turn this into a quantitative diagnostic for verifying the predictability assumption as discussed in \cite{BJW18a}.
Specifically, given a set of independent identically distributed (iid) parameter samples drawn from the initial distribution, the corresponding
sample average of the ratio $r(Q(\lambda))$ should be approximately unity. 
To make this a computationally cheap diagnostic to compute in this work, we re-use the same parameter samples involved in estimating $\preddens$.
More precisely, we estimate $\preddens$ from a KDE applied to a set of QoI samples obtained by evaluating the QoI map on an iid set of parameters generated from the initial distribution. 
The quantitative diagnostic defined by the sample average of $r(Q(\lambda))$ is then estimated without any further model simulations.
It is the experience of the authors that values between $[0.9, 1.1]$ are expected when the predictability assumption is satisfied and sufficiently accurate estimates of the predicted and observed densities are utilized.

\subsection{Illustrative Example: Randomly Generated Waves (Part I)}
\label{sec:wave_method1}
% \st{\bf To-Do: Show the wave equation example from Taylor's PhD proposal and results associated with a fixed senor that generates a QoI associated with data taken at a particular moment in time for each experiment.}
As an illustrative example, we consider the random generation of 2-dimensional water waves in a square domain where the goal is to determine the distribution of initial droplets based on observations of an a priori defined QoI map.
For the mathematical model, we consider the prototypical wave equation $u_{tt}=\nabla^2 u$, where $u=u(x,y,t)$ denotes the heights of the waves on a physical (dimensionless) domain defined by $(0,5)^2\subset \mathbb{R}^2$ with homogeneous Dirichlet boundary conditions.
The model is solved with a standard centered finite difference scheme on a $101\times 101$ regular uniformly-spaced mesh ($99\times 99$ mesh when excluding the boundary) with $0.005$ sized time-steps\footnote{See~\ref{sec:software} on instructions for obtaining the code utilized to generate the results in this example.}.
The initial condition models a droplet of water at some uncertain location $(a,b)\in(0,5)^2$ given by 
\begin{equation*}
    u(x,y,0)=0.2\exp\left(-10\left( (x-a)^2+(y-b)^2\right)\right).
\end{equation*}

\begin{figure}[htbp]
\centering
\includegraphics[scale=0.45]{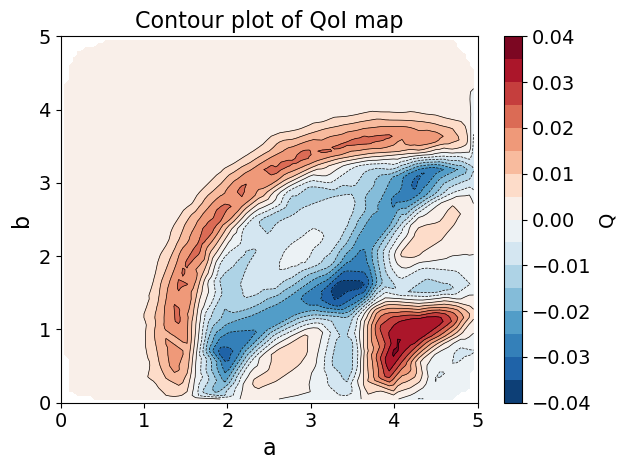}
\caption{Contour plot of the QoI map $Q$ over $[0,5]^2$ for illustrative example (part I) with the specified QoI map $Q(a,b)=u\left( 4.0,1.0,2.5;a,b\right)$ which represents the wave height at spatial location $(4.0,1.0)$ 2.5 time units after initial water droplet at $(a,b)$ occurs.}\label{fig:wave_method1_contour}
\end{figure}
In the more abstract notation given above, $\lambda=(\lambda_1,\lambda_2)=(a,b)$ defines the parameters for this problem with $\pspace=(0,5)^2$ denoting all physically possible values for these parameters.  
To emphasize the role of $(a,b)$ on the solution, we rewrite it as $u(x,y,t;a,b)$, and we define the QoI map as $Q\left( a,b\right)=u\left( 4.0,1.0,2.5;a,b\right)$. 
In other words, we assume that a single measurement of a wave height is obtained at $t=2.5$ at location $(x,y)=(4.0,1.0)$.
Note first that $Q$ maps from $\mathbb{R}^2$ into $\mathbb{R}$ so that the contour structure of $Q$ significantly influences the DCI solution. 
See Figure~\ref{fig:wave_method1_contour} for a contour plot of $Q$ over $\pspace$.

To construct the observed distribution, we utilize a \emph{data-generating (DG) distribution} defined as the exact distribution of $(a,b)$ values that leads to the observations.
Here, $a$ and $b$ are assumed independent with the DG distribution for $a$ (the horizontal location of a droplet) taken as a $\text{Beta}(2,5)$ distribution that is scaled and shifted to the interval $[1,2]$
%\footnote{In other words,  $a=2X+1$ for random variable $X\sim\text{Beta}(2,5)$.} 
while the DG distribution for $b$ (the vertical location of a droplet) is taken as a $N(2.5,0.5)$ distribution.
%\footnote{In other words, $b=Y$ for random variable $Y\sim N(2.5, 0.5)$.}.
% Since the calculation of a DCI solution requires an observed distribution, we must typically estimate this distribution from finite sampling, and the DCI solution will be data-consistent in the sense of calculating a distribution on the parameter space that matches the \emph{estimate} of the observed distribution when pushed forward through the model.
% In the context of this paper, we focus on calculating densities so that we perform a Kernel Density Estimation (KDE) of the observed density from the finite samples. 
% We therefore first explore differences between the exact DG distribution and a KDE of this distribution beforehand to help set the stage for comparing DCI solutions to exact true solutions in the examples presented.
We randomly generate 200 iid $(a,b)$ values from the joint DG distribution and numerically solve the model to simulate the associated QoI values defining the observations.

\begin{figure}[h]
\centering
\includegraphics[width=0.32\textwidth]{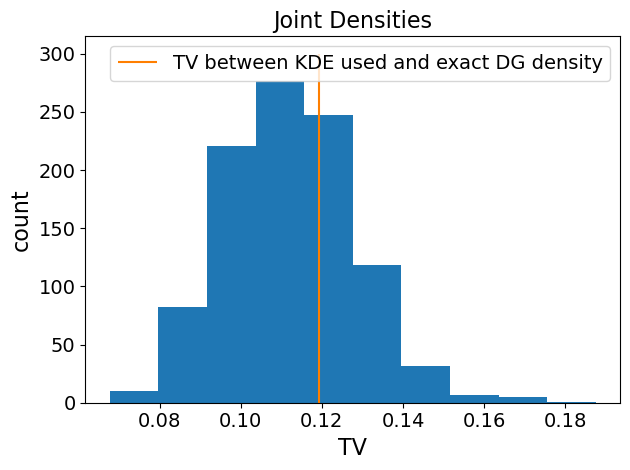}
\includegraphics[width=0.32\textwidth]{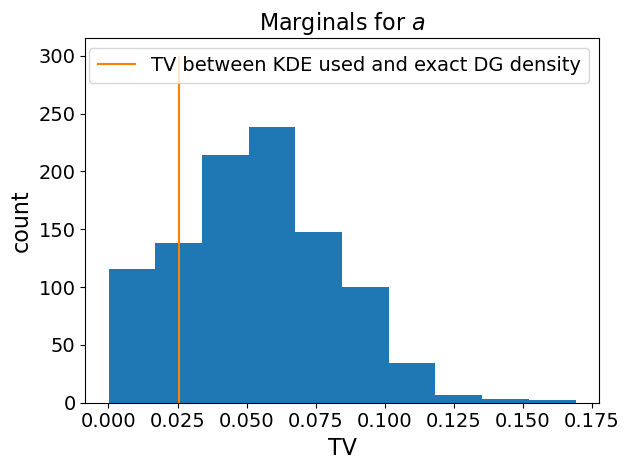}
\includegraphics[width=0.32\textwidth]{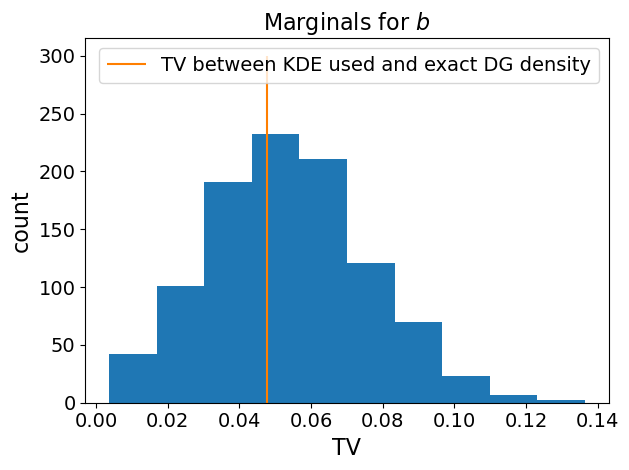}
\caption{Histogram of TV values between KDEs and exact joint DG densities using 1000 random sets of 200 random iid samples. On the left is the resulting histogram. The middle and right plots represent the resulting histograms for the horizontal and vertical marginals, respectively. The vertical lines in each plot represent the TV metric between the KDEs of 200 i.i.d. samples used for the illustrative example throughout the paper.}\label{fig:joint_TVs}
\end{figure}

To quantify the differences and similarities between densities, we make use of the total variation (TV) metric throughout this work.
In particular, we use the standard form of the TV metric for densities defined as one-half the $L^1$-distance between densities \cite{SGF+10}.
This produces values between $[0,1]$ with a value of $1$ obtained when the two densities have non-overlapping support.
To provide context on typical TV values obtained for KDE estimates of the DG distribution, we generate $\expnumber{1}{3}$ sets of 200 iid $(a,b)$ samples, compute the corresponding KDEs for each set of 200 samples, and then compute the TV distance between these KDEs and the exact DG distribution. 
These results are summarized in the histograms of Figure~\ref{fig:joint_TVs}, which illustrates that a ``typical'' TV distance for the joint KDE estimate is between approximately $0.08$ and $0.14$ from the exact joint density while the marginal estimates appear concentrated around a value of $0.05$. 
The vertical lines in these plots correspond to a particular set of 200 iid $(a,b)$ samples drawn from the DG distribution that we utilize throughout this work.
This particular set of samples from the DG distribution produces KDE estimates of the joint, $a$-marginal, and $b$-marginal densities that are TV distances from their exact counterparts of approximately $0.12$, $0.03$, and $0.05$, respectively.

% \begin{figure}[h]
% \centering

% \caption{Histogram of TV values between KDEs and exact marginal DG densities using 1000 sets of 200 random iid samples.}\label{fig:marginal_TVs}
% \end{figure}	

To construct the DCI solution, we take the uniform distribution on $[0,5]^2$ as the initial distribution for $(a,b)$, randomly generate $\expnumber{1}{4}$ iid samples from this initial distribution, simulate the associated QoI values, and then apply a standard KDE to estimate the predicted density. 
We then calculate the associated $r(Q(\lambda))$ values for these predicted QoI values based on the estimated observed and predicted densities. 
The sample mean of these $r$-values (i.e., the quantitative diagnostic as defined by~\eqref{eq:diagnostic}) is, to three-decimals, 0.961, which indicates the predictability assumption is verified and that the predicted and observed densities are reasonably approximated.

\begin{figure}[h]
\centering
% \subfloat[{Joint density}]{\includegraphics[width=0.45\textwidth]{graphics/wave_joint_1.png}
% }\quad
\subfloat[{Marginal for $a$.}]{%
\includegraphics[width=0.45\textwidth]{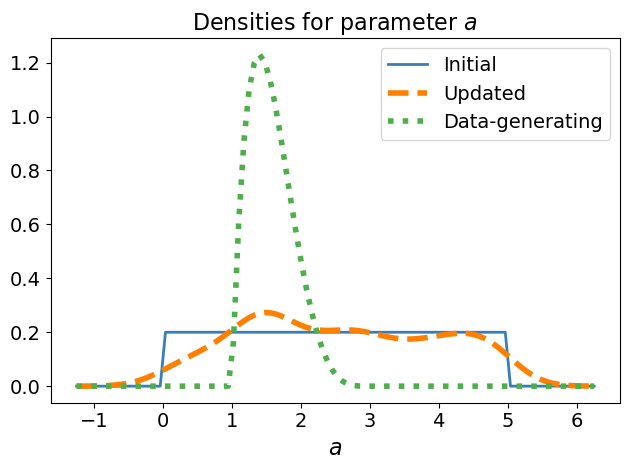}
}\quad
\subfloat[{Marginal for $b$.}]{%
\includegraphics[width=0.45\textwidth]{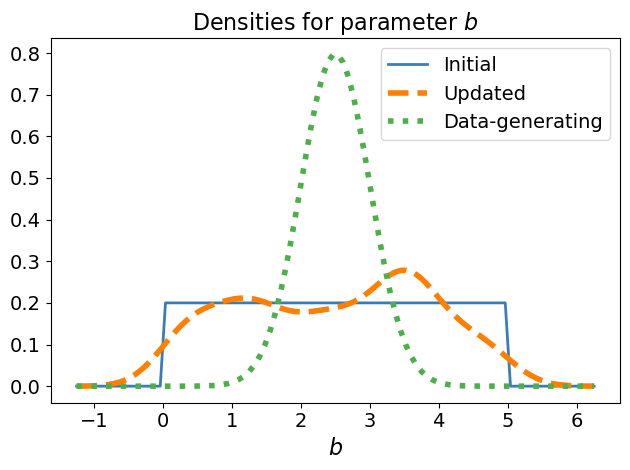}
}
\caption{(Left) Color plot of the DCI solution using $10E3$ uniform samples colored by the updated density for the illustrative example part I with data-generating samples overlaid in black. (Middle and Right) Updated density marginals of $a$ and $b$ for the illustrative wave equation example part I along with the original data-generating marginals. The TV metric between the initial predicted distribution and the data-generating distribution is 0.808 while the TV between the updated solution and the data-generating distribution is 0.718 showing minor improvements in learning the data-generating distribution.}\label{fig:wave_joint_1}
\end{figure}

We use a mixture of qualitative and quantitative methods for analyzing and comparing the various densities.
First, we utilize the $r$-values to visualize the updated density and its marginals in Figure~\ref{fig:wave_joint_1}. 
The left plot shows the $\expnumber{1}{4}$ initial samples colored by their associated $r$-values, which illustrates the impact of the contour structure of the QoI map as well as the uniform initial density.
Clearly, the updated density is no longer uniform although this initial distribution does essentially assign uniform likelihoods across the contours of the QoI map.
There is a clear concentration of probability in a contour event containing most of the samples generated from the DG distribution.
This is reflected in the TV metric computations as well where the TV metric between the exact uniform initial joint density and the KDE of the DG joint density is 0.808 while the TV metric is 0.718 between the KDEs of the updated and DG joint densities.
The updated marginals are computed as weighted KDE estimates with the $r$-values serving as the weights and are shown in the middle and right plots (for $a$ and $b$, respectively).
These are of limited utility and do not appear significantly updated from the uniform initial distributions due to the contour structure of the QoI map. 
This is reflected in the TV metric computations where we see the TV metrics between the exact uniform marginals and KDE of the DG marginals are 0.706 and 0.544 for $a$ and $b$ (resp.) while the TV metrics between the KDEs of the updated marginals and DG marginals are 0.651 and 0.543 for $a$ and $b$ (resp.). 

% The TV metrics between $\updens$ and its marginals from the associated DG distribution and its marginals are summarized in Table~\ref{tab:tv-example-1}.
% To provide context, Table~\ref{tab:tv-example-1} also summarizes the TV metrics between the exact DG distribution and its marginals from the KDE approximations computed from the 200 samples, which quantifies the amount of approximation error present from utilizing KDEs.
% Clearly, the updated density and its marginals are far away from a typical KDE estimate of the DG distribution. 

% \begin{table}
% \begin{center}
% \begin{tabular}{|c|c|}\hline
% TV between exact and KDE of DG & \\\hline 
% TV between $\pi_{\text{update}}$ and exact DG & 0.7678 \\\hline
% TV between $\pi_{\text{update}}$ and KDE of DG density & 0.7306 \\\hline
% TV between horizontal marginals of $\pi_{\text{update}}$ and exact DG densities & 0.6541 \\\hline
% TV between horizontal marginals of $\pi_{\text{update}}$ and KDE of DG densities & 0.6244 \\\hline
% TV between vertical marginals of exact and KDE of DG densities & 0.0753 \\\hline
% TV between vertical marginals of $\pi_{\text{update}}$ and exact DG densities & 0.5464 \\\hline
% TV between vertical marginals of $\pi_{\text{update}}$ and KDE of DG densities & 0.5103 \\\hline
% \end{tabular}
% \end{center}
% \caption{Various TV metrics between $\pi_{\text{update}}$ and DG densities for example Part I.}\label{tab:tv-example-1}
% \end{table}

The above qualitative and quantitative discrepancies between the updated and DG distributions highlights the deficiency in both a priori specifying and utilizing a single QoI.
Unfortunately, given the complexity in the contour structure of any particular measurement over the parameter space (e.g., as shown in Figure~\ref{fig:wave_method1_contour}), it is not at all obvious how one would a priori specify multiple QoI that lead to an updated density that is a reasonable approximation of the true data-generating distribution.
This issue is addressed in the next section.

% \begin{figure}[h]
% \centering
% \subfloat[{Marginal for $a$.}]{%
% \includegraphics[width=0.45\textwidth]{graphics/wave_marginal_a_1.png}
% }\quad
% \subfloat[{Marginal for $b$.}]{%
% \includegraphics[width=0.45\textwidth]{graphics/wave_marginal_b_1.png}
% }
% \caption{Updated density marginals for wave equation example part I along with the original data-generating marginals.}\label{fig:wave_marginals_1}
% \end{figure}	

\section{Learning Uncertain Quantities (LUQ) for Temporal Data}
\label{sec:LUQ} 

Let $\mathcal{M}(u;\lambda)=0$ define a model that admits a solution $u(\lambda)$ of state variables dependent upon model parameters $\lambda$ representing various uncertain inputs of the model taking values in parameter space $\Lambda \subset \mathbb{R}^n$. 
Note that in this work $u$ also depends on spatial or temporal variables, but we omit explicit dependence on those variables here both for simplicity in notation and to emphasize the dependence on the parameters. 
While the solution operator of the model defines a mapping from parameters, $\lambda$,  to state variables, $u$, in practice it is impossible to observe the full solution across all of space or time.
Instead, we collect a finite amount of spatial or temporal data on the solution $u$ that is then analyzed and transformed to the quantitative information useful for defining and solving a stochastic inverse problem.
This is the goal of the Learning Uncertain Quantities (LUQ) framework.

\subsection{The LUQ framework}

Originally introduced in \cite{MSB+22}, the LUQ framework defines a three-step data-to-distribution pipeline to transform noisy temporal data clouds for dynamical systems into distributions of learned QoI for the purposes of DCI. 
An open-source Python module \texttt{LUQ} is also available \cite{pyLUQ2.0.0}. 
We summarize the three steps below and refer the interested reader to \cite{MSB+22} for more details.

\subsubsection{Step 1: Filter Noisy Data}\label{subsec:filtering}
The filtering step occurs in two stages. 
The first step is to approximate the underlying dynamical signal responsible for the noisy data. 
Then, from this approximate signal, ``noise-free'' (i.e., filtered) data are produced.

% The choice of splines in the filtering step is motivated by a type of universal approximation theory for splines (see \cite{UFW17}) where it is possible to approximate the underlying dynamical response (assuming a particular regularity or finite number of discontinuities) to arbitrary pointwise accuracy if both a sufficiently high frequency for collecting data and number of knots are used. 

In \cite{MSB+22}, splines are utilized for filtering temporal measurements based on a universal approximation theorem and their ubiquity in inverse problems (see \cite{UFW17} for a thorough treatise of the subject).
For each set of temporal measurements obtained at a fixed point in space, piecewise linear splines are optimized to approximate the underlying dynamical response associated with the noisy temporal data. 
While the first and last knot are fixed to the beginning and end of the time-interval, the interior knot placements are optimized along with the values of the spline at each knot. 
The number of knots used is incrementally increased from a user-specified minimum until either the relative improvement in the sum square error falls below a threshold or a maximum number of knots, specified by the user, is reached. 
From the resulting piecewise linear splines, a specified number of usually equispaced data points are then calculated defining the filtered time series data.

\subsubsection{Step 2: Learn and Classify Dynamics}
The second step of LUQ, which is optional, is to perform unsupervised learning in order to label (i.e., define clusters) followed by supervised learning to classify the various types of dynamics present in the filtered data. 
This is an important step when different types of signals are present in the observed dynamics (e.g., due to bifurcations), which may be quantified by entirely distinct types and numbers of QoI. 

The \texttt{LUQ} module fully supports some of the most popular clustering schemes such as $k$-means \cite{arthur2006k}, Gaussian mixture modeling \cite{mclachlan1988mixture}, and spectral clustering routines \cite{ng2002spectral,von2007tutorial}.
All such methods begin with a specification of the number of labels (i.e., clusters) to learn from the predicted data.
While these schemes often implicitly contain a classifier model that can be applied to other data (e.g., the observed data), 
Support Vector Machines (SVMs) are a more robust class of supervised learning methods that can be used for, among other things, classification that has advantages in terms of operations in high dimensions and memory efficiency.
The \texttt{LUQ} module leverages the widely used library \verb|LIBSVM| \cite{chang2011libsvm} for SVMs to construct classifiers. 

Denote by $K$ the number of labels to learn from the predicted data that are then utilized to construct an SVM for classification.
This implicitly defines a partitioning of $\pspace$ into $K$ subsets\footnote{In general, the $K$ subsets may be both disconnected and non-convex.}, which we denote by $\pspace_k$ for $k\in\set{1,2,\ldots,K}$.
It is important to note that it is not technically necessary to explicitly identify $\pspace_k$ for any $k$, but it is computationally trivial to identify which initial subset of samples drawn from the initial distribution belong to each $\pspace_k$ by simply sorting the labels of the associated sample set of prediction data.
We make use of the notation $\pspace_k$ to describe the structure of the DCI solution at the end of Step 3.

\subsubsection{Step 3: Learn QoI via Feature Extraction and Compute the DCI solution}
The \texttt{LUQ} module makes use of kernel-based principal component analysis (kPCA) \cite{pearson1901liii,scholkopf1997kernel,mika1999kernel,abdi2010principal}, which is one of the most popular approaches to performing feature extraction.
Specifically, kPCA is performed with multiple kernels on each classification of predicted filtered time series data to learn the most beneficial QoIs for each type of dynamics. 
We provide an in-depth mathematical derivation of the QoI maps in terms of the chosen kernel in Section~\ref{sec:QoI_kernels}, which is subsequently utilized in a proposed quantitative method for evaluating the sufficiency of the filtered data for learning the QoI.

While the \texttt{LUQ} module technically ends after feature extraction is performed to learn the QoI, it is now possible to begin computing the DCI solution based on this learned QoI.
To that end, let $Q^{(k)}(\lambda)$ denote the learned QoI map defined on $\pspace_k$ and $\preddens^{(k)}$ denote a predicted density associated with a re-normalization of $\initdens$ restricted to $\pspace_k$. 
Similarly, let $\obsdens^{(k)}$ denote the observed density associated with the $k$th label of data and $w_k$ denote the {\em observed weights} associated with this label defined by the ratio of observed data with this label to all observed data of any label, i.e., $0\leq w_k\leq 1$, $\sum_{k=1}^K w_k=1$, and
\begin{equation}
    \sum_{k=1}^K w_k\obsdens^{(k)}(Q^{(k)}(\lambda))\mathbb{I}_{\pspace_k}(\lambda)
\end{equation}
defines the observed density over $\pspace$ where $\mathbb{I}_{\pspace_k}$ denote the indicator function over $\pspace_k$. 
With this notation, the DCI solution is given by
\begin{equation}\label{eq:updated-cluster}
	\updens(\lambda) = \sum_{k=1}^K w_k\left(\initdens(\lambda)\frac{\obsdens^{(k)}(Q^{(k))}(\lambda)}{\preddens^{(k)}(Q^{(k)}(\lambda))}\right)\mathbb{I}_{\pspace_k}(\lambda).
\end{equation}

While using the same number of principal components (PCs) as the number of desired QoIs will typically explain the majority of variance in the data, the resulting QoIs may be heavily correlated leading to a poor construction of the updated density. 
To decide whether a set of desired QoI are to be used for DCI, we check the diagnostic discussed in \ref{sec:diagnostic} for each label (i.e., we compute the diagnostic for each ratio of $\obsdens^{(k)}$ to $\preddens^{(k)}$ shown in~\eqref{eq:updated-cluster}) and systematically reduce the number of QoI extracted for each type of dynamic if the diagnostic deviates too far from unity.

\subsection{Illustrative Example: Randomly Generated Waves (Part II)}
\label{sec:wave_method2}
% \st{\bf To-Do: Return to the wave equation exaple but now consider that the fixed sensor location records data over a time interval so that we learn QoI and the distribution associated with temporal data to then construct a data-consistent solution. Results demonstrate a need to expand LUQ for both spatial and spatio-temporal data.}

Returning to the illustrative example from Section~\ref{sec:wave_method1} with the same DG and initial distributions as well as the same random samples from these distributions, we now consider temporal data recorded every 0.5 time units up to $t=7.0$ at the spatial location $(4.0,1.0)$ producing 14 measurements of wave heights for every experiment\footnote{See~\ref{sec:software} on instructions for obtaining the code utilized to generate the results in this example.}. 
In the interest of simplicity and to illustrate limitations in the analysis of only temporal data in the previous LUQ framework, we avoid any introduction of noisy data. 
This allows us to skip Step 1.
Moreover, clustering is not found to be useful in this case, so we also skip Step 2. 
Ideally, we seek a QoI map of the same dimension as the parameter space to attempt to recover the DG distribution. 
Hence, the goal is to learn a 2-dimensional QoI map for the DCI solution. 
To this end, we apply Step 3 and perform kernel PCA on the 14-dimensional data to learn the 2-dimensional QoI map for DCI.
The sample average of the resulting $r$-values (i.e., the quantitative diagnostic as defined by~\eqref{eq:diagnostic}) for this QoI map is approximately 1.031 indicating that the predictability assumption is satisfied and the observed and predicted densities are well approximated. 

\begin{figure}[h]
\centering
\subfloat[{Joint density.}]{
\includegraphics[width=0.3\textwidth]{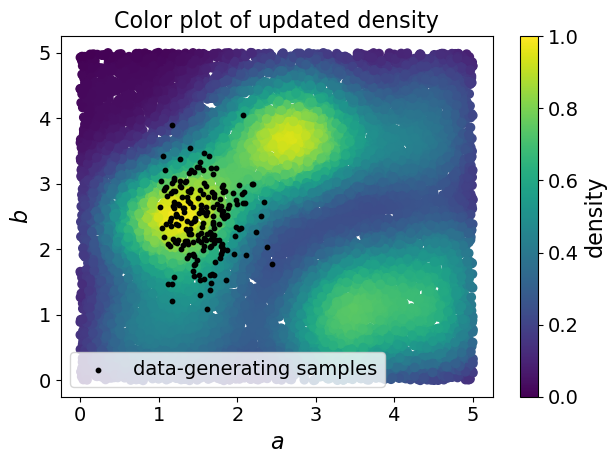}
}\quad
\subfloat[{Marginal for $a$.}]{%
\includegraphics[width=0.3\textwidth]{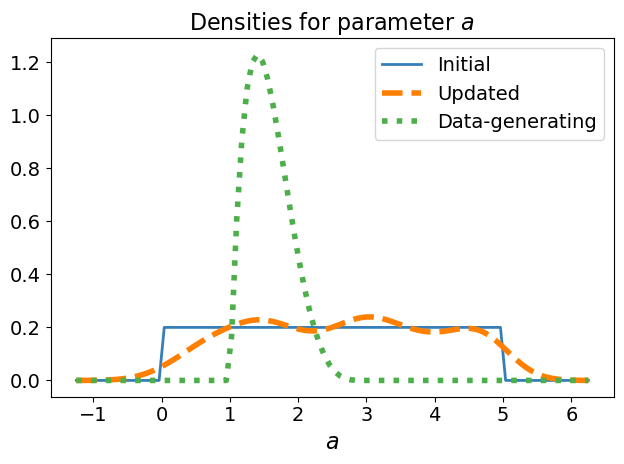}
}\quad
\subfloat[{Marginal for $b$.}]{%
\includegraphics[width=0.3\textwidth]{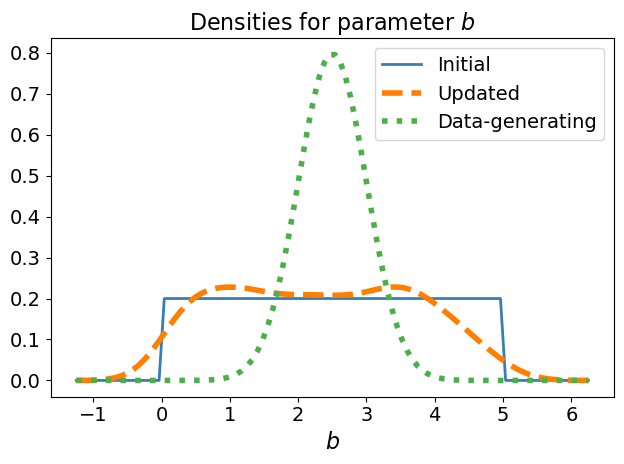}
}
\caption{(Left) Color plot of the DCI solution using $10E3$ uniform samples colored by the updated density for the illustrative example part II with data-generating samples overlaid in black. (Middle and Right) Updated density marginals of $a$ and $b$ for the illustrative wave equation example part II using purely temporal data along with the original data-generating marginals. The TV metric between the updated solution and the data-generating distribution is 0.731. For part I of this example, the resulting TV metric was 0.718. While there is an increase in TV metrics, the increase is within expected error tolerances with the quadrature used to compute the TV metric integral suggesting that there are minor changes in the results from parts I and II.}\label{fig:wave_joint_2}
\end{figure}

The QoI map learned is still producing a coarse contour structure for the updated density as evidenced by Figure~\ref{fig:wave_joint_2}. 
While there appears to be improved concentration of probability near the samples drawn from the DG distribution, there is no improvement in terms of the TV metric.
Here, the TV metric between the updated joint density and the exact DG density is approximately 0.731.
By comparison, the TV metric between the updated joint density and the exact DG density is 0.718 using the QoI from Section~\ref{sec:wave_method1}. 
It is self-evident from Figure~\ref{fig:wave_joint_2} that the marginals are not improved from those seen in Figure~\ref{fig:wave_joint_1}. 
% This exhibits that there is very little difference in the quality of the DCI solution whether a single measurement is taken at the location $(4.0,1.0)$ or 14 measurements every 0.5 time units at the same location. 
When using strictly temporal data from different spatial locations, similar results occur. 
This highlights the need to extend the LUQ framework to admit spatio-temporal data rather than simply temporal data.

% \begin{figure}[h]
% \centering
% \subfloat[{Marginal for $a$.}]{%
% \includegraphics[width=0.45\textwidth]{graphics/wave_marginal_a_2.png}
% }\quad
% \subfloat[{Marginal for $b$.}]{%
% \includegraphics[width=0.45\textwidth]{graphics/wave_marginal_b_2.png}
% }
% \caption{Updated density marginals for wave equation example part II along with the original data-generating marginals.}\label{fig:wave_marginals_2}
% \end{figure}	

\section{Extending LUQ for Spatial and Spatio-Temporal Data}
\label{sec:LUQ_extensions}

Utilizing linear splines or even higher order splines to filter spatial or spatio-temporal data introduces technical difficulties due to the increased dimension that the spatial domain can introduce as well as the complexity in defining a mesh of potential spline knots. 
At the same time, the filtering step is essential in LUQ since we cannot reasonable expect a kPCA to learn the QoI maps that best explain perturbations in parameters that are useful for DCI from data with significant noise levels as this results in much of the variation in the data being explained by the noise itself. 
We therefore extend both the LUQ framework and module \texttt{LUQ} to learn a mixture of polynomial and radial basis functions from noisy data for the purpose of filtering data in other dimensions. 
This filtering of data utilizes a variety of user-specified functions and is described in general terms through the use of a Neural Network (NN) \cite{SCHMIDHUBER201585}. 

\subsection{A deep-learning based perspective for filtering}
\label{subsec:NNs} 

\begin{figure}[h]
\centering
\includegraphics[scale=0.27]{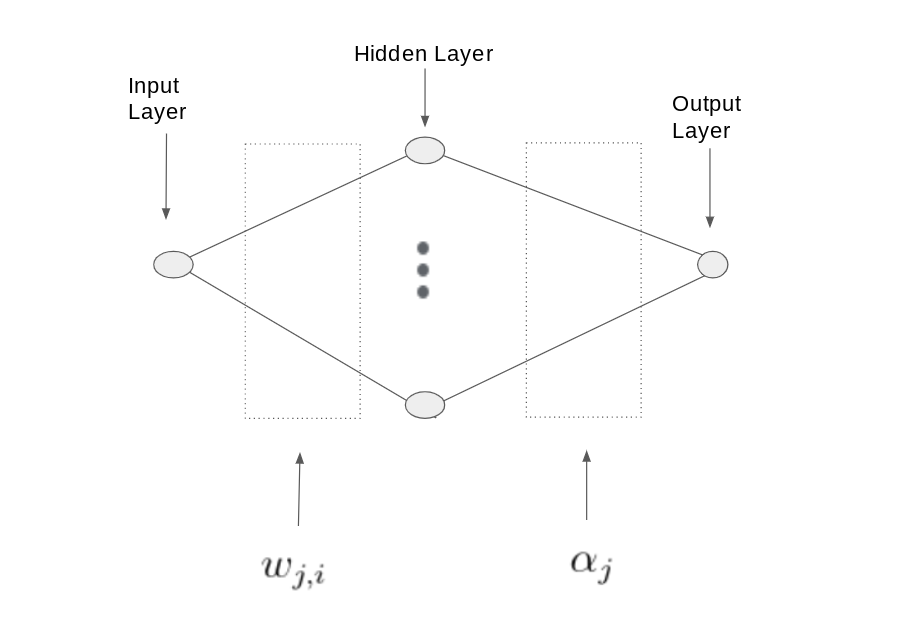}
\caption{Neural network schematic illustrating location of hyperparameters $w_{j,i}$ and $\alpha_j$ in~\eqref{eq:NN-output} and~\eqref{eq:NN-output-simple}.}
\label{fig_schematic}
\end{figure}

Much like with the use of splines for filtering temporal data, the motivation for using NNs comes from a Universal Approximation Theorem (UAT) for NNs that states that there exists a NN with a single hidden layer that can approximate a continuous function defined on a compact subset of $\mathbb{R}^m$ to arbitrary accuracy in the sup-norm metric \cite{HORNIK1989359}. 
Assuming a scalar valued function is to be approximated on an $m$-dimensional input space using a NN with a single hidden layer of width $n$, then such a NN's output, denoted by $N(x)$ is given by a function of the form
\begin{equation}\label{eq:NN-output}
     N(x) = \mathcal{A}\left(\alpha_0 + \sum_{j=1}^n \alpha_j \phi_j\left(w_{j,0} + \sum_{i=1}^m w_{j,i} x_i\right)\right).
\end{equation}
Here, $\mathcal{A}$ is the activation function of the neuron on the output layer while $\phi_j$ is the activation function associated with the $j$th neuron in the hidden layer.
The $\alpha_j$, for $0\leq j\leq n$, define the hyperparameters between the hidden and output layer, while the $w_{j,i}$, for $0\leq i\leq m$ for each $0\leq j\leq n$, define the hyperparameters between the input and hidden layer. 
See Figure~\ref{fig_schematic} for a schematic. 
All of these hyperparameters are learned via backpropagation \cite{SCHMIDHUBER201585}. 
When $\mathcal{A}$ is chosen to be the identity, as we initially utilize for simplicity, the equation for $N(x)$ simplifies to
\begin{equation}\label{eq:NN-output-simple}
    N(x) = \alpha_0 + \sum_{j=1}^n \alpha_j \phi_j\left(w_{j,0} + \sum_{i=1}^m w_{j,i} x_i\right).
\end{equation}

While the UAT establishes the existence of a NN achieving a specified level of accuracy (assuming significant data are provided to train the NN), it does not provide any insight as to the necessary width (i.e., the number of neurons) required to achieve such accuracy. 
Some insight is still generally required to setup a range of potentially useful NNs.
Below, we describe how such insight may be obtained in the context of utilizing NNs in the filtering step for LUQ.

First, consider a NN with $n$ neurons where each neuron is activated via rectified linear unit (ReLU) activation functions \cite{Bach2017}. 
In the case of temporal data, the input to the $j$th neuron in this layer is defined by $w_{j,1} t + w_{j,0}$. 
The activation function of the $j$th neuron is then given by $\phi_j(w_{j,1} t + w_{j,0}) = \max\left\{ 0,w_{j,1} t + w_{j,0}\right\}$. 
The output of the NN (assuming the identity activation output layer) is then equivalent to utilizing linear splines where the values of $\{w_{j,0},w_{j,1}\}_{j=1}^n$ are associated with node placement while the values of $\alpha_j$ for $0\leq j\leq n$ are associated with the value of the spline through the nodes. 
In this case, identifying a range of neurons to use in the hidden layers of potential NNs is equivalent to choosing a range of knot values to use in linear spline approximations.
A pre-analysis of noisy data (e.g., using visualization or other techniques) or domain-specific knowledge of solutions to models and their general features, then allows the user to specify a range of neurons necessary for approximating the underlying temporal signal. 

We adapt this conceptual perspective to the use of radial basis functions (RBFs).
Specifically, we consider the ubiquitous Gaussian-type functions of the form $\phi_j (x) = e^{-\lVert x-r_j\rVert ^2/\sigma_j}$ where $x$ is the $m$-dimensional location variable.
While this can be rewritten using the $w_{j,i}$ notation for hyperparameters, it is conceptually simpler to explain the interpretation of the hyperparameters in the notation of $r_j$ and $\sigma_j$. 
The hyperparameters $r_j$ are interpreted as the $m$-dimensional vectors defining the centers of the activation functions and the $\sigma_j$ denotes a scaling factor associated with the width of the activation.
If $\sigma_j$ is ``small,'' then the associated $j$th neuron is generally associated with a localized feature present in the function near the $r_j$ location whereas if $\sigma_j$ is ``large,'' then the associated $j$th neuron is generally associated with a non-localized trend (such as an average value) present in the function ``around'' the $r_j$ location.
This perspective and interpretation is valuable in user-provided guidance in specifying the depth of the NN required to approximate data with a NN instead of linear splines. 
Moreover, for moderate levels of complexity in the function response (i.e., for functions with a few localized features such as those resulting from many smoothly modeled phenomena), we can use $n$ values that permit direct optimization of the loss function associated with training a NN, which is often equivalent to minimizing a sum square error.
We discuss this more below.

\subsection{RBF-based filtering in \texttt{LUQ}}

One of the redesigns of the \texttt{LUQ} package produced as part of this work is the inclusion of an RBF fitting algorithm using a trust-region reflective least squares algorithm \cite{TR-least-squares} available within the \texttt{scipy} Python package for the filtering step. 
This allowed us to circumnavigate the need to tune training parameters in NNs and to speed up the optimization by not using the more robust optimization algorithms standard in large-scale machine-learning libraries. 
We note, however, that the RBF fitting technique we deploy is mathematically equivalent to fitting a specific NN. 

Before fitting the data, we perform some standard pre-processing of the data by first centering the data according to its mean. 
The RBF fitting algorithm we implement allows for centering the data over a linear or quadratic trend if desired. 
With the data centered, we then fit a user-specified number $n$ of Gaussian-type RBFs and an optional polynomial $p(x)$ of user-specified degree as $$\sum_{i=1}^n w_ie^{-\left\lVert x-r_i\right\rVert ^2/\sigma_i}+p(x)$$ 
where for each $i=1,\dots,n$, the hyperparameters that are fitted are the weights $w_i$, centers $r_i$, scales $\sigma_i$, and any coefficients that define the optionally specified polynomial $p$. 
It is worth noting that the options to center the data over a linear or quadratic trend as well as the option to include a polynomial to learn as part of the RBF fitting serves to generally reduce the number $n$ of RBFs required to approximate the underlying function since more global trends in the data are approximated by these polynomial objects, which are typically more well-suited to such tasks than RBFs. 

\subsection{Filtered data from disparate datasets}

It is worth noting an additional feature of the LUQ framework that is fully incorporated within the new \texttt{LUQ} module developed as part of this work. 
The predicted data is often simulated and therefore straightforward to define on a spatial grid utilized within the simulation model. 
The observed data may be more irregularly spaced and not aligned to any grid.
The \texttt{LUQ} module allows for such disparities between datasets by allowing the user to specify precisely the coordinates where (predicted or observed) data are recorded and identifying the unifying set of filtered coordinates for which the filtered data are obtained.
This flexibility allows for functionality of the framework even in situations where the predicted and observed datasets are of different dimensions.
Iterating through the filtering method within \texttt{LUQ} for each individual sample associated with a specific predicted or observed stream of data further allows the user to specify data coordinates for each sample, which is useful in situations where data coordinates or number of data differ from one iteration to the next (e.g., due to adaptive mesh refinement/coarsening in the predicted case or accounting for sensor outages in an observation network). 
The filtering step is particularly crucial in such situations to ensure a common dimensionality of the data vectors analyzed in either the clustering or feature extraction steps.

\subsection{Illustrative Example: Randomly Generated Waves (Part III)}\label{sec:wave_method3} 

\begin{figure}[h]
\centering
\includegraphics[scale=0.45]{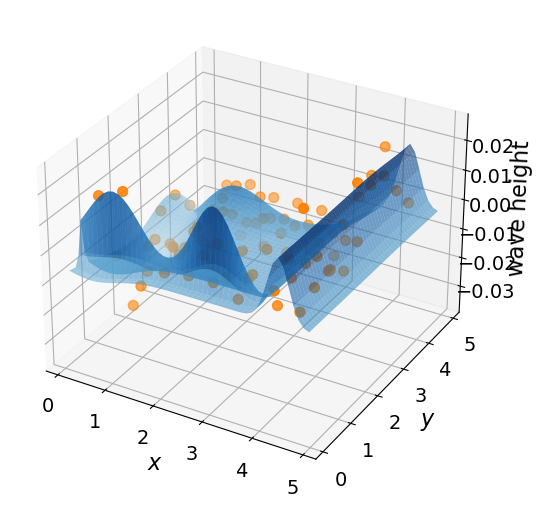}
\caption{Random sample taken from the 200 i.i.d. observed samples consisting of exact wave height measurements over the spatial domain $[0,5]$ along with the fitted surface computed by fitting a number of Gaussians up to 7 total until a relative tolerance is reached from the mean-centered noisy wave height data. The fitting learns the locations, scales, and weights of each Gaussian. The sample used in this plot was fitted using 6 Gaussians.}
% sample_num = 20
\label{fig:filter_step}
\end{figure}

Revisiting the illustrative example, we now assume we have sensors at 81 spatial locations given by $\left( 0.5i, 0.5j\right)$ for $i,j=1,\dots,9$ that record the wave height at $t=2.5$. 
For the sake of illustrating the filtering step for spatial data, we add ``white'' noise simulated by a normal distribution centered at $0.0$ with standard deviation of $\expnumber{2.5}{-3}$, which is a significant amount of noise given that the wave height measurements are on the order of $\expnumber{1}{-2}$). 
We specify a range of 1-7 RBFs to approximate the wave's deviation from its approximate mean value computed from the noisy data.
Beginning with a single RBF, the number is incrementally refined until either the relative error between fittings reaches an error tolerance of 0.01 or the a priori specified maximum number of 7 RBFs is reached. 
A representative plot is shown in Figure~\ref{fig:filter_step} for an RBF-based surface that is learned for a single randomly generated sample of observation data.
Note that the predicted data exist at each time step on the computational mesh of $99\times 99$ spatial coordinates.
However, we choose to evaluate the RBF-based surface on a different mesh defining the filtered data coordinates given by the mesh consisting of 2401 spatial locations given by $\left(0.1i,0.1j\right)$ for $i,j=1,\dots,49$.
This is not an arbitrary choice for the filtered coordinates as we show in Section~\ref{sec:QoI_kernels} where we provide both a mathematical and statistical justification for choosing these filtered coordinates.
For now, we simply note that for each sample from the DG distribution there are 81 noisy data points, for each sample from the initial distribution there are 9801 data points, but for each sample, regardless of the distribution it comes from, there are 2401 filtered data points. 
% Thus, the fitted function for the filtered observed data is subsequently evaluated at the same spatial locations as the simulated predicted data so that the relatively sparse number of noisy observations are transformed into a filtered data set that matches the structure of the predicted data. 

\begin{figure}[h]
\centering
\subfloat[{Joint density.}]{\includegraphics[width=0.3\textwidth]{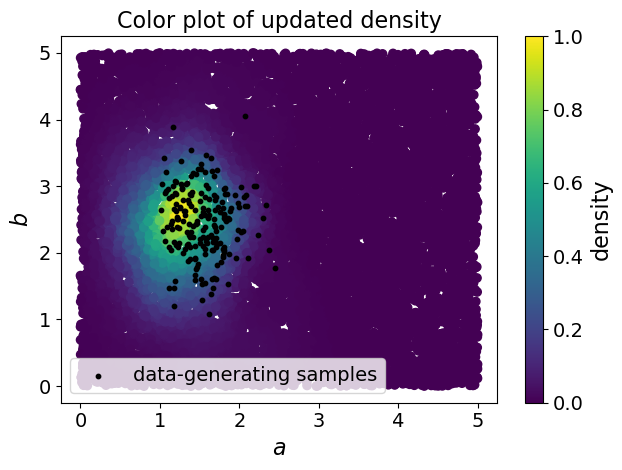}
}
\subfloat[{Marginal for $a$.}]{%
\includegraphics[width=0.3\textwidth]{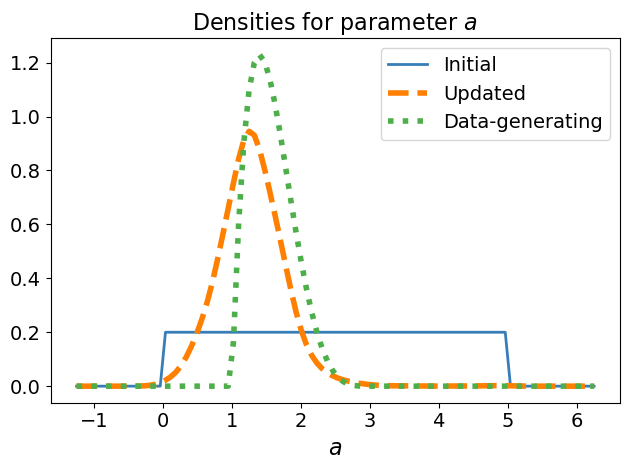}
}\quad
\subfloat[{Marginal for $b$.}]{%
\includegraphics[width=0.3\textwidth]{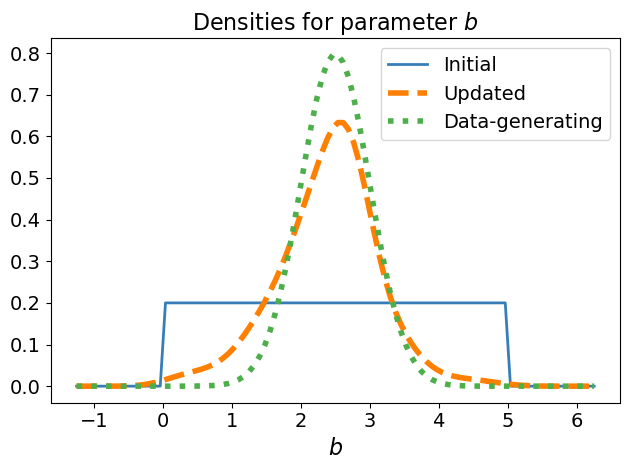}
}
\caption{(Left) Color plot of the DCI solution using $10E3$ uniform samples colored by the updated density for the illustrative example part III with data-generating samples overlaid in black. (Middle and Right) Updated density marginals of $a$ and $b$ for the illustrative wave equation example part III using noisy spatial data along with the original data-generating marginals. The TV metric between the updated solution and the data-generating distribution is about 0.390 showing a major improvement from parts I and II of this example that produced TV metrics of about 0.718 and 0.731, respectively. The TV metric between the KDE of the data-generating distribution and the exact data-generating distribution is about 0.119 suggesting that the DCI solution has room for improvement which is achieved in part IV by utilizing a combination of spatial and temporal data.}\label{fig:wave_joint_3}
\end{figure}

The learned QoI map is obtained via kPCA on the predicted spatial data, and the remaining steps are analogous to the prior illustrative example parts. 
The resulting updated density along with the marginals of the updated density compared to the data-generating densities are shown in Figure~\ref{fig:wave_joint_3}.
The resulting TV between the updated density and the exact DG density is about 0.390 which is a significant improvement from the values of 0.718 and 0.731 obtained in Part I using a single measurement at the spatial location $(4.0,1.0)$ and Part II using temporal data at the same spatial location, respectively.
These qualitative and quantitative results demonstrate a significant improvement over the results of Parts I and II of this illustrative example in terms of the updated density providing a better approximation to the DG density, though the value obtained here is still not within the expected range of TV values between KDEs of the DG density and the exact DG density which is within about 0.01 and 0.13.

% \begin{figure}[h]
% \centering
% \subfloat[{Marginal for $a$.}]{%
% \includegraphics[width=0.45\textwidth]{graphics/wave_marginal_a_3.png}
% }\quad
% \subfloat[{Marginal for $b$.}]{%
% \includegraphics[width=0.45\textwidth]{graphics/wave_marginal_b_3.png}
% }
% \caption{Updated density marginals for wave equation example part III along with the original data-generating marginals.}\label{fig:wave_marginals_3}
% \end{figure}	

\section{Temporal Iterations of Spatially Filtered Data to Update Distributions}
\label{sec:DCI_iterative}

\subsection{Iteratively obtained QoI and DCI}

Once the spatial data are filtered at a particular time as discussed in the previous section, we learn the QoI maps to use for DCI. 
If the data are purely temporal or spatial, then QoI maps are learned and utilized as described in Steps 2 and 3 of Section~\ref{sec:LUQ}. 
For spatio-temporal data, we propose an iterative approach to calculating a DCI solution. 
At a conceptual level, after the first updated density is computed in the first iteration, the initial density at each subsequent iteration is defined as the updated density of the prior iteration. 

Mathematically, given the $i$th iteration that produces an updated density denoted by $\updens^{(i)}$, the $i+1$ iteration for $i\geq 1$ starts with an initial density defined by $\initdens^{(i+1)}:=\updens^{(i)}$.
We then attempt to learn $p=\text{dim}\left(\Lambda\right)$ number of QoI and denote the learned map at this iteration as $Q^{(i+1)}$ (not to be confused by the notation used to denote any potential part of the QoI map associated with a cluster), and compute the weights $r^{(i+1)}(Q^{(i+1)}(\lambda))$ at this iteration using the ratio of observed to predicted densities constructed and evaluated from this learned QoI map as described in Section~\ref{sec:DCI}.
%$\left(\lambda\right)=\frac{\pi_{\text{obs}}\left(Q^{(i)}\left(\lambda\right)\right)}{\pi^{(i)}_{\text{pred}}\left(Q^{(i)}\left(\lambda\right)\right)}$ where $\pi^{(i)}_{\text{pred}}$ is the push-forward of $\pi^{(i)}_{\text{init}}$ through $Q^{(i)}$ so that the $i^{\text{th}}$ iteration's solution is 
This results in the $i+1$ update defined as $\updens^{(i+1)}(\lambda):=r^{(i+1)}(Q^{(i+1)}(\lambda))\initdens^{(i+1)}(\lambda)$. 
Denoting $\initdens$ as the original initial density for the first iteration, it follows that
\begin{equation}
    \updens^{(i+1)}(\lambda) = \left(\prod_{j\leq i+1} r^{(j)}(Q^{(j)}(\lambda))\right)\initdens(\lambda).
\end{equation}
%$$\pi^{(i)}_{\text{init}}=r_{i-1}\pi^{(i-1)}_{\text{init}}=\left(\prod_{k<i}r_k\right)\pi_{\text{init}}; \:\:\: \pi^{(i)}_{\text{update}}=\left(\prod_{k\leq i}r_k\right)\pi_{\text{init}}.$$ 
This is summarized in Algorithm~\ref{alg:iterative_DCI} below where we emphasize how the DCI diagnostic is utilized at each iteration to systematically determine the number of components of each $Q^{(i+1)}$ map to utilize.
Note the possibility that no components are utilized.
This is a feature of the algorithm that avoids two undesirable situations.
One situation involves a violation of the predictability assumption, which can occur, for example, due to an inability to sufficiently filter noise at a particular time step.
Another situation involves data that are insensitive to parameter value perturbations, which can result, for example, from dynamics converging to steady-state or equilibrium values resulting in data for which it is fundamentally not possible to quantitatively learn a useful QoI map for DCI. 

\begin{algorithm}
\caption{$i+1$ Iteration of Iterative DCI}\label{alg:iterative_DCI}
\begin{algorithmic}[1]
% \State initialize $r_{\text{old}}(Q(\lambda))$ array for initial/predicted samples, $tol$ parameter, and $d$=\# of QoI to learn
% \For {each time index} 
\State apply kPCA to spatial data at current time step to learn new $d$-dimensional QoI map $Q^{(i+1)}(\lambda)$
\State evaluate $r = \prod_{j\leq i}r^{(j)}(Q^{(j)}(\lambda))$ at initial parameter samples
\State estimate $\obsdens$ from learned QoI samples and $\preddens$ as a weighted KDE of learned QoI samples weighted by $r$
\State evaluate $r^{(i+1)}(Q^{(i+1)}(\lambda))=\obsdens(Q^{(i+1)}(\lambda))/\preddens(Q^{(i+1)}(\lambda))$ at learned predicted QoI samples
\If {$| 1-\text{mean}(r^{(i+1)}r)| > tol$}
\State restart using $(d\leftarrow d-1)$ or discard time step if $d=1$
\EndIf
% \EndFor
% \State new updated density is $\pi^{(i+1)}_{\text{update}}=r^{(i+1)}r\pi_{\text{init}}$ as weighted KDE of initial samples
\end{algorithmic}
\end{algorithm} 

While the DCI diagnostic in this iterative approach informs us when it is possible to utilize data from a particular time step to update the distribution for the next iteration, it does not state whether an iterative update is computationally worthwhile. 
The impact of a potential $i+1$ iteration to improve the overall solution requires the contour structure of $Q^{(i+1)}$ to differ from that of $Q^{(i)}$. 
This naturally leads to questions involving optimal experimental design (OED) for learning QoI maps.
While defining any formal OED metrics based on quantification of geometric structures is outside the scope of this work, it is the subject of ongoing research in DCI.
Here, we note that the iterative approach does permit, a simple statistic that quantifies, in a sense, the degree of change in a proposed updated density in terms of the standard deviation of the $r$-values. 
It is straightforward to verify that the variance of the $r$-values for a given QoI map $Q$ with respect to $\initmeas$ is given by 
\begin{align*}
\mathbb{V}_{\text{init}}\left(r(Q(\lambda))\right) := \mathbb{E}_{\text{init}}\left(r^2(Q(\lambda))\right)-\mathbb{E}_{\text{init}}^2\left(r(Q(\lambda))\right) =\mathbb{E}_{\text{update}}\left(r(Q(\lambda))\right)-1
\end{align*}
where $\mathbb{E}_{\text{init}}$ and $\mathbb{E}_{\text{update}}$ are the expected value operators with respect to $\initmeas$ and $\upmeas$, respectively. 
We note that $\mathbb{V}_{\text{init}}\left(r(Q(\lambda))\right)=\mathbb{E}_{\text{update}}\left(r(Q(\lambda))\right)-1$ implies that $\mathbb{E}_{\text{update}}\left(r(Q(\lambda))\right) \geq 1$.
Moreover, $\mathbb{E}_{\text{update}}\left(r(Q(\lambda))\right)$ quantifies the expected weight given to samples of $\predmeas$ that are aligned with $\obsmeas$.

At a computational level, if the $i+1$ iteration results in $\mathbb{V}_{\text{init}}\left(\prod_{j\leq i+1}r^{(j)}(Q^{(j)}(\lambda))\right)>\mathbb{V}_{\text{init}}\left(\prod_{j\leq i}r^{(j)}(Q^{(j)}(\lambda))\right)$, then the $i+1$ iteration is providing a statistically substantial change to the iterative DCI solution.
%whereas if $\text{Var}_{\text{init}}\left(r_i\right)\leq \text{Var}_{\text{init}}\left(r_{i-1}\right)$, then the $i^{\text{th}}$ iteration is not learning anything statistically different from the previous iterations. 
Additionally, since $\mathbb{V}_{\text{init}}\left(r(Q(\lambda))\right)=\mathbb{E}_{\text{init}}\left(r^2(Q(\lambda))\right)-1$, we can simply use the $r^2$-values of iid samples drawn from the initial distribution and compute the sample average analogous to the DCI diagnostic that computes the sample average of the $r$-values. 
That is, an optional ``significance'' check can be added to Algorithm~\ref{alg:iterative_DCI}, which we summarize in Algorithm \ref{alg:iterative_DCI_r2} below.

% \begin{algorithm}
% \caption{Iterative DCI with Significance Check}\label{alg:iterative_DCI_r2}
% \begin{algorithmic}[1]
% \State initialize $r_{\text{old}}$
% \For {each time index} 
% \State extract spatial data at time index
% \State learn QoI using kernel PCA
% \State transform observed and predicted data
% \State calculate KDE of $\pi_{\text{obs}}$ and $\pi_{\text{pred}}$ using transformed samples
% \State evaluate $r=\dfrac{\pi_{\text{obs}}}{\pi_{\text{pred}}}$ at transformed samples
% \State define $r_{\text{new}}=r_{\text{old}}r$
% \If {$\left| 1-\text{mean}\left(r_{\text{new}}\right)\right| > \text{tol}$}
% \State restart at 4 using less QoI or set $r_{\text{new}}=r_{\text{old}}$
% \EndIf
% \If {$\text{mean}\left(r_{\text{new}}^2\right)\leq \text{mean}\left(r_{\text{old}}^2\right)$}
% \State disregard time step and set $r_{\text{new}}=r_{\text{old}}$
% \EndIf
% \EndFor
% \State calculate $\pi_{\text{update}}=r_{\text{new}}\pi_{\text{init}}$ as weighted KDE of initial samples
% \end{algorithmic}
% \end{algorithm} 

\begin{algorithm}
\caption{$i+1$ Iteration of Iterative DCI with Significance Check}\label{alg:iterative_DCI_r2}
\begin{algorithmic}[1]
% \State initialize $r_{\text{old}}(Q(\lambda))$ array for initial/predicted samples, $tol$ parameter, and $d$=\# of QoI to learn
% \For {each time index} 
\State apply kPCA to spatial data at current time step to learn new $d$-dimensional QoI map $Q^{(i+1)}(\lambda)$
\State evaluate $r = \prod_{j\leq i}r^{(j)}(Q^{(j)}(\lambda))$ at initial parameter samples
\State estimate $\obsdens$ from learned QoI samples and $\preddens$ as a weighted KDE of learned QoI samples weighted by $r$
\State evaluate $r^{(i+1)}(Q^{(i+1)}(\lambda))=\obsdens(Q^{(i+1)}(\lambda))/\preddens(Q^{(i+1)}(\lambda))$ at learned predicted QoI samples
\If {$| 1-\text{mean}(r^{(i+1)}r)| > tol$}
\State restart using $(d\leftarrow d-1)$ or discard time step if $d=1$
\ElsIf {$\text{mean}((r^{(i+1)}r)^2)\leq \text{mean}(r^2)$}
\State discard time step
\EndIf
% \EndFor
% \State new updated density is $\pi^{(i+1)}_{\text{update}}=r^{(i+1)}r\pi_{\text{init}}$ as weighted KDE of initial samples
\end{algorithmic}
\end{algorithm}

% \begin{algorithm}
% \caption{Iterative DCI with Significance Check}\label{alg:iterative_DCI_r2}
% \begin{algorithmic}[1]
% \State initialize $r_{\text{old}}(Q(\lambda))$ array for initial/predicted samples, $tol$ parameter, and $d$=\# of QoI to learn
% \For {each time index} 
% \State apply kPCA to spatial data at time index to learn new $d$-dimensional QoI map $Q_\text{new}(\lambda)$
% \State estimate $\obsdens$ and $\preddens=r_{\text{old}}\pi_{\text{init}}$ for learned (observed and predicted) QoI samples
% \State evaluate $r(Q_\text{new}(\lambda))=\obsdens(Q_\text{new}(\lambda))/\preddens(Q_\text{new}(\lambda))$ at learned predicted QoI samples
% \State evaluate $r_{\text{new}} = r_{\text{old}}r$ at initial parameter samples
% \If {$\left| 1-\text{mean}\left(r_{\text{new}}\right)\right| > tol$}
% \State restart at line 4 using $d-1$ components of learned QoI map for $Q_{\text{new}}$ or set $r_{\text{new}}=r_{\text{old}}$ if $d=1$
% \ElsIf {$\text{mean}\left(r_{\text{new}}^2\right)\leq \text{mean}\left(r_{\text{old}}^2\right)$}
% \State disregard time step and set $r_{\text{new}}=r_{\text{old}}$
% \EndIf
% \State set $r_{\text{old}}=r_{\text{new}}$
% \EndFor
% \State calculate $\pi_{\text{update}}=r_{\text{new}}\pi_{\text{init}}$ as weighted KDE of initial samples
% \end{algorithmic}
% \end{algorithm} 

An important feature of the iterative DCI framework is that it allows immediate utilization of new observed data to update the DCI solution without requiring any re-computation of prior information. 
In other words, once an iterative solution is computed, the only values required to store in memory are the original samples of $\initdens$, the current $r$-values at these samples, and any model information associated with the initial samples required to generate predicted data at the next potential iteration. 
It is also possible to setup either algorithm to utilize \texttt{LUQ} with different filtering options (in terms of centering data, utilizing different depths of NNs, and/or including polynomial trend fitting) at each iteration as the complexity of the underlying spatial signal at each time evolves. 
% With new observed data at time $t=T$, we push the $\pi_{\text{init}}$ samples through the model to compute the predicted spatial data at $t=T$ if the data at time $t=T$ has yet to be computed and proceed with the iterative DCI approach to calculate a new iterative update. {\bf Reference numerical examples that touch on this.}

% {\bf Add info about LUQ handling spatio-temporal data that might have different spatial structure at each time step.}

\subsection{Illustrative Example: Randomly Generated Waves (Part IV)} 
\label{sec:wave_method4}

% \st{\bf Re-word what is currently below and put in the introduction when discussing the outline of the paper in terms of where the different parts of this illustrative example show up in the narrative of the paper. Method 1 is in Section 2. Method 2 is in Section 3. Method 3 is in Section 4, and Method 4 is shown here.}

\begin{figure}
\centering
\includegraphics[scale=0.45]{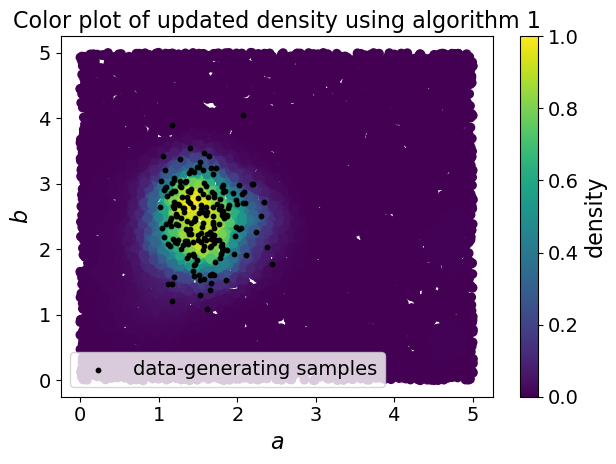}
\includegraphics[scale=0.45]{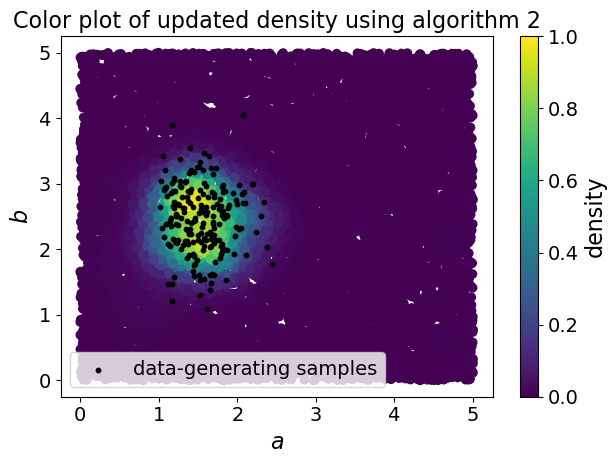}
\caption{Color plots of the DCI solutions using $10E3$ uniform samples colored by the updated densities for part IV of the illustrative example with data-generating samples overlaid in black. The left plot is from using Algorithm \ref{alg:iterative_DCI} that does not utilize the $r^2$ values, and the right plot is from using Algorithm \ref{alg:iterative_DCI_r2} that does utilize the $r^2$ values. The resulting TV metrics are 0.161 for Algorithm \ref{alg:iterative_DCI} and 0.160 for Algorithm \ref{alg:iterative_DCI_r2}. For reference, the TV between the initial uniform distribution and the exact data-generating distribution is about 0.808 while the TV between the KDE of the data-generating distribution and the exact data-generating distribution is about 0.119.}\label{fig:wave_joint_4}
\end{figure}

\begin{figure}
\centering
\subfloat[{Algorithm \ref{alg:iterative_DCI}}]{%
\includegraphics[width=0.45\textwidth]{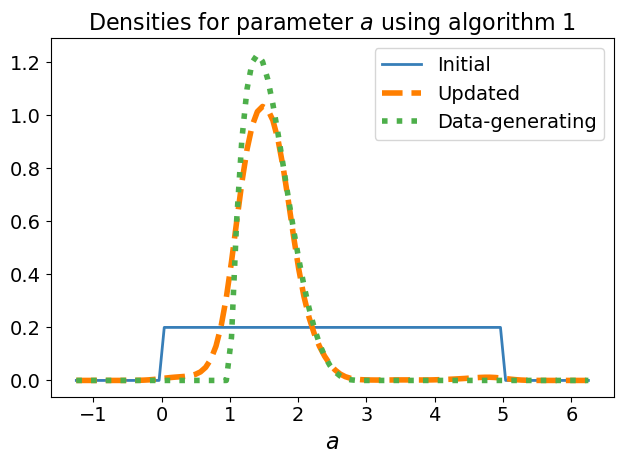}
\includegraphics[width=0.45\textwidth]{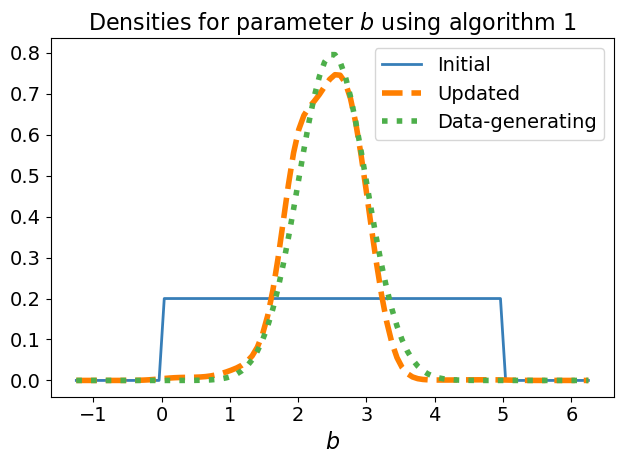}
}\quad
\subfloat[{Algorithm \ref{alg:iterative_DCI_r2}}]{%
\includegraphics[width=0.45\textwidth]{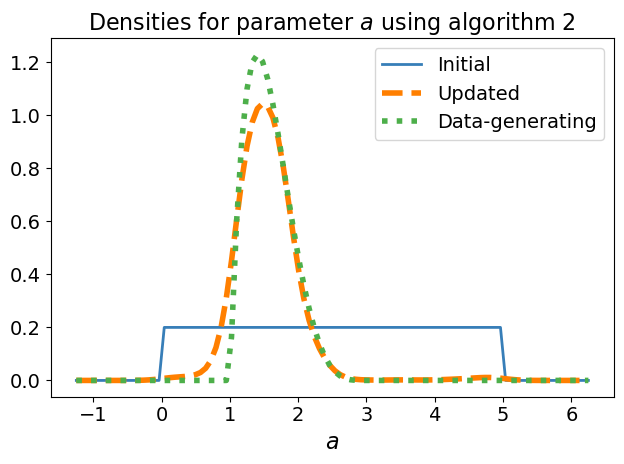}
\includegraphics[width=0.45\textwidth]{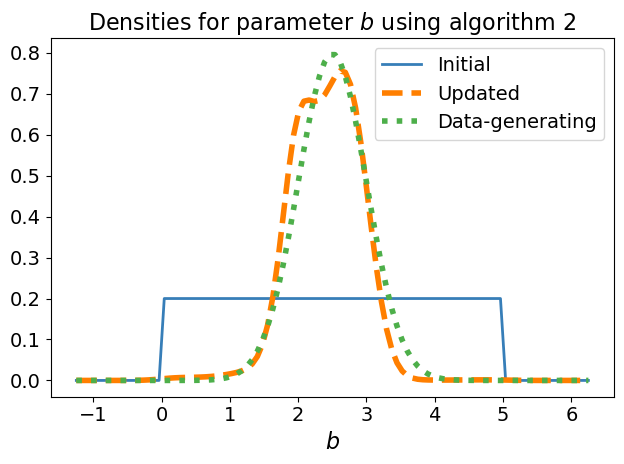}
}
\caption{Updated density marginals for wave equation example part IV along with the original data-generating marginals. Top is from using Algorithm \ref{alg:iterative_DCI}; bottom is from using Algorithm \ref{alg:iterative_DCI_r2}. The resulting TV metrics for parameter $a$ are 0.034 and 0.037 for Algorithms \ref{alg:iterative_DCI} and \ref{alg:iterative_DCI_r2}, respectively; for parameter $b$, the metrics are 0.071 and 0.037 for Algorithms \ref{alg:iterative_DCI} and \ref{alg:iterative_DCI_r2}, respectively. These values are within the expected range of metric values of about 0.01 to 0.08 for parameter $a$ and 0.01 to 0.10 for parameter $b$ as seen in Figure \ref{fig:joint_TVs}.}\label{fig:wave_marginals_4}
\end{figure}

Revisiting the wave equation example, we show how spatio-temporal data can be fully utilized. 
Suppose that there are sensors at each $(0.5i, 0.5j)$ for $i, j = 1, . . . , 9$ that record the wave height every 0.5 time units from $t = 2.5$ to $t = 7.0$. 
As in Part III of this illustrative example, ``white'' noise is simulated by a normal distribution cetered at 0.0 with standard deviation of 2.5E-3.
The spatial data at each time step are filtered by iteratively fitting 1-7 RBFs to approximate the wave's deviation from its approximate mean value by starting with 1 RBF and increasing to 7 RBFs stopping if a relative error of less than 0.01 is reached.
% Since the simulated predicted data are computed on a finer grid consisting of spatial locations $\left(0.1i,0.1j\right)$ for $i,j=1,\dots,49$, the fitted functions for the observed data is re-evaluated on this finer grid so that the dimensionalities of the observed and predicted data match.
We use the same filtered data coordinates as in Part III of this illustrative example. 
At each time step, kPCA is performed and the corresponding predicted and observed densities are estimated from the transformed samples corresponding to the first two principal components. 
% If the sample mean of the resulting $r$-values deviate more than 0.095 away from unity, then we re-attempt using only the first PC to define the QoI map. 
% If the new $r$-values still deviate more than 0.095 away from unity, then the spatial data at this time step is not utilized to update the initial distribution. 
We consider both Algorithms~\ref{alg:iterative_DCI} and \ref{alg:iterative_DCI_r2} with $tol$ to $0.095$.
% Note that the optional impact check can also be performed as in Algorithm~\ref{alg:iterative_DCI_r2}. 
% The updated density is than calculated as a weighted KDE of the initial density using the product of the $r$-values from each kept iteration as the weights. 
The resulting updated joint densities using both algorithm are compared in Figure~\ref{fig:wave_joint_4}, and the marginals are compared in Figure~\ref{fig:wave_marginals_4}. 

\begin{figure}
\centering{%
\includegraphics[width=0.45\textwidth]{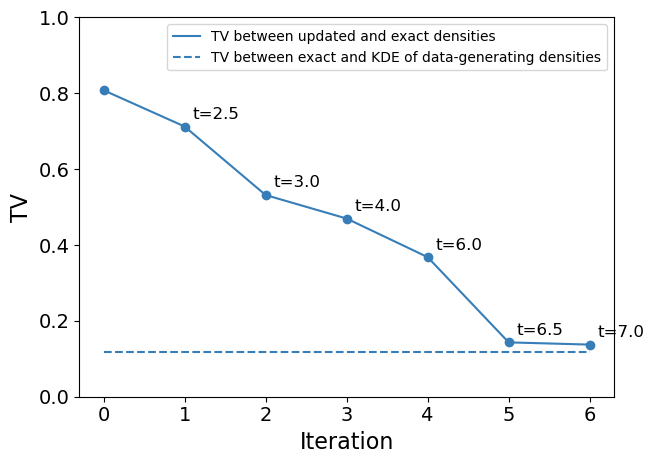}
}\quad{%
\includegraphics[width=0.45\textwidth]{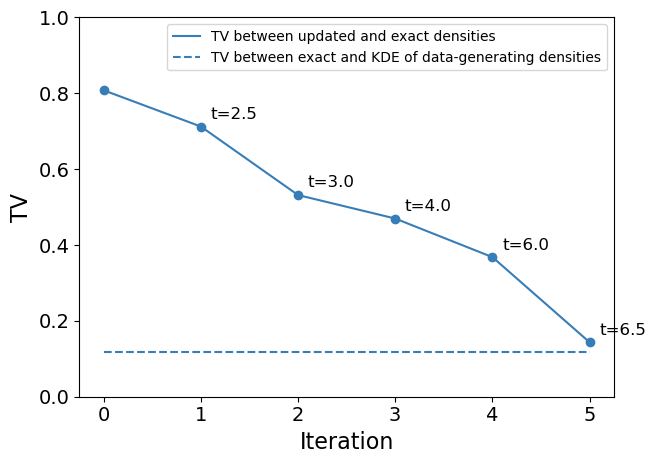}
}
\caption{TV metrics between updated iterative densities and the exact data-generating distribution for the illustrative wave equation example part IV. Left is from using Algorithm \ref{alg:iterative_DCI}; right is from using Algorithm \ref{alg:iterative_DCI_r2}. The time of each iteration used is annotated for each data point. The dotted lines show the TV between the KDE of the data-generating distribution and the exact data-generating distribution.}\label{fig:wave_TV_joint}
\end{figure}

\begin{figure}
\centering{%
\includegraphics[width=0.45\textwidth]{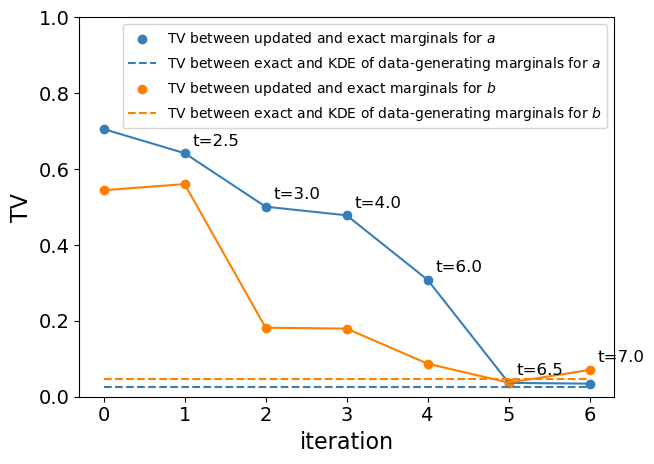}
}\quad{%
\includegraphics[width=0.45\textwidth]{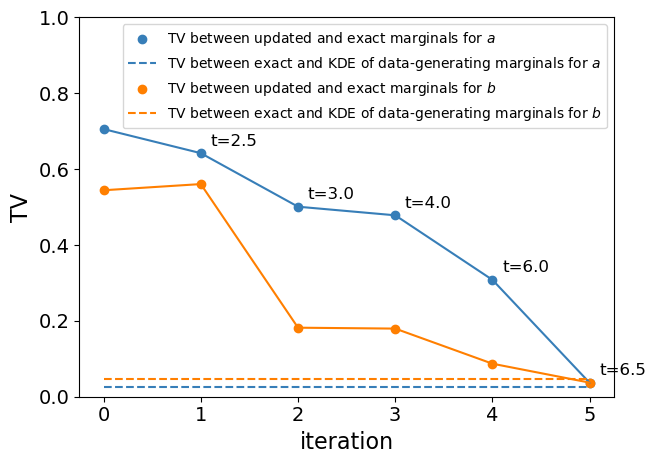}
}
\caption{TV metrics between updated iterative marginal densities and the exact data-generating distribution for the illustrative wave equation example part IV. Left is from using Algorithm \ref{alg:iterative_DCI}; right is from using Algorithm \ref{alg:iterative_DCI_r2}. The time of each iteration used is annotated for each data point. The dotted lines show the TV metrics between the KDEs of the data-generating marginal distributions and the exact data-generating marginal distributions.}\label{fig:wave_TV_marginal}
\end{figure}

The TV metrics between the final updated density and the exact DG density is about 0.161 using Algorithm \ref{alg:iterative_DCI} and about 0.160 using Algorithm \ref{alg:iterative_DCI_r2}. 
For the marginal densities, the TV metrics between the marginals for $a$ are about 0.034 and 0.037 and for $b$ are about 0.071 and 0.037 when using Algorithms \ref{alg:iterative_DCI} and \ref{alg:iterative_DCI_r2}, respectively.
Plots of the TV metrics at each iteration for both algorithms are shown in Figure~\ref{fig:wave_TV_joint} (for the joint densities) and Figure~\ref{fig:wave_TV_marginal} (for the marginals). 
Note that the final values are within the range of TV values expected from a KDE estimate of the DG density and its marginals as shown in Figure~\ref{fig:joint_TVs}.
% TVs between the exact DG density and the KDE using samples of the DG density are 0.121 for the joint, 0.101 for the horizontal marginal, and 0.020 for the vertical marginal. 
% This can be observed in the above mentioned plots that include dotted lines representing these TV values.

% \begin{figure}[h]
% \centering
% \subfloat[{Marginal for $a$.}]{%
% \includegraphics[width=0.45\textwidth]{graphics/wave_marginal_a_iter.png}
% }\quad
% \subfloat[{Marginal for $b$.}]{%
% \includegraphics[width=0.45\textwidth]{graphics/wave_marginal_b_iter.png}
% }
% \caption{}\label{fig:}
% \end{figure}	

\section{Learned QoI as Functions of Kernels and Numerical Sufficiency of Filtered Data}\label{sec:QoI_kernels}

In this section, we derive a closed-form representation of a learned QoI map as a function of the kernel utilized within the kPCA approach applied in the final step of the LUQ framework.
We then apply the mathematical insight provided by this representation to develop a quantitative procedure for identifying when a sufficient number of filtered data are computed for learning the QoI. 

\subsection{QoI maps as functions of kernels}

At a high-level, a kernel function denoted by $\kernel(\cdot,\cdot)$ is a measure of similarity, i.e., $\kernel(a, b) > \kernel(a, c)$ if objects $a$ and $b$ are considered ``more similar'' than objects $a$ and $c$ \cite{zhang2007local}.
In the context of this work, the objects we consider are the filtered datasets.
Let $\noisydata$ denote the space of unfiltered data (either observed or predicted) and $\filteredspace$ the space of filtered data obtained by the filtering step of LUQ.
With this notation, we assume that $\kernel:\filteredspace\times\filteredspace\to[0,\infty)$ is symmetric and positive definite. 
With these assumptions, the kernel is both a measure of similarity between filtered data vectors as well as a representation of an inner product in the so-called reproducing kernel Hilbert space (RKHS) with reproducing kernel $\kernel$. 
This Hilbert space, denoted by $\mathcal{H}$, is the span of the collection $\left\{ \kernel \left(\cdot ,\bm{d}\right)\right\}_{\bm{d}\in \filteredspace}$. 
In this context, for every $\bm{d}\in \filteredspace$, $\kernel \left(\cdot, \bm{d}\right)$ represents the unique element in $\mathcal{H}$ such that for every $h\in\mathcal{H}$, 
\begin{equation}\label{eq:reproducing_property}
    h(\bm{d})=\langle \kernel\left(\cdot,\bm{d}\right),h(\cdot)\rangle_{\mathcal{H}}
\end{equation}
where $\langle \cdot,\cdot\rangle_{\mathcal{H}}$ is the inner product of $\mathcal{H}$.

% This Hilbert space, denoted by $\mathcal{H}$, consists of functions $h:\filteredspace\to \mathbb{R}$ with inner product $\langle \cdot,\cdot\rangle_{\mathcal{H}}$ such that for each $h\in \mathcal{H}$ and $\bm{d}\in\filteredspace$, 
% \begin{equation}\label{eq:reproducing_property}
%     h(\bm{d}):=\langle \kernel\left(\cdot,\bm{d}\right),h(\cdot)\rangle_{\mathcal{H}}.
% \end{equation}
The above equation is referred to as the reproducing property of $\kernel$.
The Moore-Aronszajn theorem \cite{Aronszajn1950} implies that for each symmetric positive definite kernel $\kernel$ defined on $\filteredspace$, there exists a unique RKHS $\mathcal{H}$ with reproducing kernel $\kernel$. 
Working directly with the kernel in a RKHS has advantages in many contexts.
For instance, \cite{SMOLA1998637} presents analysis to clarify the ``dilemma'' that directly working with the kernel ``seemingly def[ies] the curse of dimensionality'' in the context of SVMs, ridge regression, and regularization networks.
The application of the kernel to significantly reduce computational complexity for kernel ridge regression is  analyzed in \cite{NIPS2017_61b1fb3f}.
Below, we work in the RKHS as a means of deriving the closed-form representation of learned QoI maps as functions of the kernel utilized in a kPCA. 

Given a set of $N$ iid parameter samples drawn from the initial distribution, let $\bm{d}^{(1)},\dots,\bm{d}^{(N)}\in \filteredspace$ denote the associated filtered data vectors.
Given an arbitrary error function $E:\left(\filteredspace \times \mathbb{R}\right)^N\to \mathbb{R}\cup \left\{\infty\right\}$, consider the optimization problem with a regularization constraint, 
\begin{align*}
    \text{minimize: } & E\left(\left(\bm{d}^{(1)},h\left(\bm{d}^{(1)}\right)\right),\dots,\left(\bm{d}^{(N)},h\left(\bm{d}^{(N)}\right)\right)\right) \\
\text{subject to: }&\lVert h\rVert_{\mathcal{H}} \leq c
\end{align*}
where $c$ is some positive constant and $\lVert h\rVert_{\mathcal{H}}=\sqrt{\langle h, h\rangle_{\mathcal{H}}}$. 
Consider the decomposition of $h\in\mathcal{H}$ as $h=h_{\parallel}+h_{\perp}$ where $h_{\parallel}$ is in the span of $\kernel\left(\cdot,\bm{d}^{(1)}\right),\dots,\kernel\left(\cdot,\bm{d}^{(N)}\right)$ and $h_{\perp}$ is in the orthogonal complement of this span. 
By the reproducing property, $h\left(\bm{d}^{(i)}\right)=\left\langle \kernel\left(\cdot,\bm{d}^{(i)}\right),h(\cdot)\right\rangle_{\mathcal{H}}$ for each $1\leq i\leq N$.
It follows that $h\left(\bm{d}^{(i)}\right)=h_{\parallel}\left(\bm{d}^{(i)}\right)$ for each $1\leq i\leq N$.
Subsequently, $E$ is unaffected by $h_{\perp}$. 
Furthermore, since $\lVert h_{\parallel}\rVert_{\mathcal{H}} \leq \lVert h\rVert_{\mathcal{H}}$, the optimal feasible solution, denoted by $h^*\in\mathcal{H}$, is in the span of $\set{\kernel\left(\cdot,\bm{d}^{(1)}\right),\dots,\kernel\left(\cdot,\bm{d}^{(N)}\right)}$.
In other words, the optimal feasible solution is given by choosing $h_\perp$ to be the zero function and determining the unique constants $\alpha_1,\dots,\alpha_N\in\mathbb{R}$ such that
\begin{equation*}
    h^*=\sum_{i=1}^N\alpha_i \kernel\left(\cdot,\bm{d}^{(i)}\right)\in\mathcal{H} \ \Longrightarrow \ h^*(\bm{d}) = \left\langle \kernel(\cdot,\bm{d}), \sum_{i=1}^N \alpha_i\kernel(\cdot, \bm{d}^{(i)})\right\rangle_\mathcal{H}, \ \forall \bm{d}\in\filteredspace.
\end{equation*}  
The reproducing property of $\kernel$ implies that $\kernel\left(\bm{d},\bm{d}^{(i)}\right):=\left\langle \kernel\left( \cdot,\bm{d}\right),\kernel \left(\cdot, \bm{d}^{(i)}\right)\right\rangle$ for all $\bm{d}\in \filteredspace$ so that $h^*$ can be constructed by computing the $\alpha_i$'s without working in the RKHS directly.
Furthermore, the argument above implies that while $\mathcal{H}$ may be infinite dimensional, we are only concerned with the finite dimensional subspace spanned by $\left\{\kernel\left(\cdot,\bm{d}^{(1)}\right),\dots,\kernel\left(\cdot,\bm{d}^{(N)}\right)\right\}$ allowing us to perfom linear PCA in $\mathcal{H}$.
We can therefore interpret a kPCA as applying a standard PCA on the data vectors after they are mapped to $\mathcal{H}$ via the so called feature map $\phi$ defined as $\phi\left(\bm{d}\right):=\kernel\left(\cdot,\bm{d}\right)$.
% Thus, one can construct $h^*$ by computing the $\alpha_i$'s which forgoes working in the RKHS directly, which is usually referred to as the feature space since we can define the feature map $\phi:\filteredspace\to\mathcal{H}$ as $\phi\left(\bm{d}\right):=\kernel\left(\cdot,\bm{d}\right)$.
% We can interpret a kPCA as applying a standard PCA on the data vectors after they are mapped to $\mathcal{H}$ via a feature map $\phi$.
With this interpretation, the kPCA seeks the unit vector pointing in the direction that maximizes the variance of the transformed \emph{centered} data.
Denote this vector by $h^*$.
This vector subsequently defines the first component of the learned QoI map.
We make this mathematically precise below. 

Define $D$ as the $N\times \text{dim}\left(\filteredspace\right)$ matrix $D:=\left[ \bm{d}^{(1)},\dots,\bm{d}^{(N)}\right]^{\top}$ and $\phi\left(D\right):=\left[ \phi\left(\bm{d}^{(1)}\right),\dots,\phi\left(\bm{d}^{(N)}\right)\right]^{\top}$. 
The transformed centered data is denoted as $\tilde{\phi}\left(D\right):=\phi\left(D\right)-\frac{1}{N}\pmb{1}\sum_{i=1}^N\phi\left(\bm{d}^{(i)}\right)$ where $\pmb{1}$ denotes the $N\times \text{dim}\left(\filteredspace\right)$ matrix of all ones. 
With this notation, the kPCA seeks to solve the following optimization problem:
\begin{align*}
\text{maximize: } & \mathbb{V}\left(\tilde{\phi}\left(D\right)h\right) \\
\text{subject to: } & \lVert h\rVert_{\mathcal{H}}\leq 1
\end{align*}
Note that the variance operator $\mathbb{V}\left(\tilde{\phi}\left(D\right)h\right)=\langle h,\mathcal{C}h\rangle_{\mathcal{H}}$ where $\mathcal{C}:=\frac{1}{N-1}\tilde{\phi}\left(D\right)\tilde{\phi}\left(D\right)^{\top}$ denotes the sample covariance matrix in the finite dimensional subspace of $\mathcal{H}$ spanned by $\left\{\kernel\left(\cdot,\bm{d}^{(1)}\right),\dots,\kernel\left(\cdot,\bm{d}^{(N)}\right)\right\}$. 
Since the variance will increase with scaling of $h$, we take $\lVert h\rVert_{\mathcal{H}}=1$ and by formulating the Lagrangian of the above optimization problem in order to set the partial derivatives of the Lagrangian with respect to $h$ and Lagrange multiplier $\ell$ equal to zero, one obtains the eigenvalue equation 
\begin{equation*}
    \mathcal{C}h=\ell h
\end{equation*} 
and $\mathbb{V}\left(\tilde{\phi}\left(D\right)h\right)=\ell$. 
The above optimization problem is therefore equivalent to solving the following least squares projection problem:
\begin{align*}
\text{minimize: }&\left\lVert \tilde{\phi}\left(D\right)-\tilde{\phi}\left(D\right)hh^{\top}\right\rVert_{\mathcal{H}}^2 \\
\text{subject to: }&\lVert h\rVert_{\mathcal{H}}\leq 1
\end{align*}
From the discussion above, we can rewrite the solution $h^*$ as $h^*=\sum_{i=1}^N\alpha_i\tilde{\kernel}\left(\cdot,\bm{d}^{(i)}\right)$ where $\tilde{\kernel}$ is the centered kernel defined as 
\begin{equation*}
    \tilde{\kernel}\left(\bm{d}^{(i)}, \bm{d}^{(j)}\right) := \left\langle \phi\left(\bm{d}^{(i)}\right)-\frac{1}{N}\sum_{n=1}^N\phi\left(\bm{d}^{(n)}\right),\phi\left(\bm{d}^{(j)}\right)-\frac{1}{N}\sum_{n=1}^N\phi\left(\bm{d}^{(n)}\right)\right\rangle_{\mathcal{H}}.
\end{equation*} 
Substituting this expression of $h^*$ into the eigenvalue equation $\mathcal{C}h^*=\ell h^*$, we obtain the matrix equation $\tilde{K}^2\bm{\alpha}=\left(N-1\right) \ell \tilde{K}\bm{\alpha}$ where $\tilde{K}$ is the centered Gram matrix, i.e. the $N\times N$ matrix with $\tilde{K}_{i,j}:=\tilde{\kernel}\left(\bm{d}^{(i)}, \bm{d}^{(j)}\right)$, and $\bm{\alpha}=\left[\alpha_1,\dots,\alpha_N\right]^{\top}$. 
Since $\tilde{K}$ is symmetric positive definite, then the matrix equation reduces to \begin{equation*}
    \tilde{K}\bm{\alpha}=\left(N-1\right) \ell \bm{\alpha}.
\end{equation*} 
Note that $\lVert h\rVert_{\mathcal{H}}=1$ is equivalent to $\lVert \bm{\alpha}\rVert_2=\frac{1}{\sqrt{(N-1)\ell}}$. 
Subsequently, solving the eigenvalue equation $\tilde{K}\bm{\alpha} = \left(N-1\right) \ell \bm{\alpha}$ for $\bm{\alpha}$ and rescaling according to 
\begin{equation*}
    \bm{\alpha}^{(1)} \mapsfrom \dfrac{1}{(N-1)\ell}\dfrac{\bm{\alpha}}{\lVert \bm{\alpha}\rVert_2}
\end{equation*} 
produces the vector of constants defining $h^*=\left[ \kernel\left(\cdot,\bm{d}^{(1)}\right),\dots, \kernel\left(\cdot,\bm{d}^{(N)}\right)\right] \bm{\alpha}^{(1)}$. 
% Projecting each transformed data point $\tilde{\phi}\left(\bm{d}^{(i)}\right)$ onto $h^*$ gives
% \begin{equation*}
%     \left\langle \tilde{\phi}\left(\bm{d}^{(i)}\right),h^*\right\rangle_{\mathcal{H}}=\tilde{K}_i
% \end{equation*} 
% where $\tilde{K}_i$ denotes the $i$th row of $\tilde{K}$. 
We have thus derived the first component of the learned QoI map associated with this kernel: \begin{equation*}
    Q_1\left(\cdot\right):=\sum_{j=1}^N\bm{\alpha}^{(1)}_j\tilde{\kernel}\left(\cdot,\bm{d}^{(j)}\right).
\end{equation*} 
For the other components of $Q$, we organize the similarly rescaled eigenvectors $\bm{\alpha}^{(i)}$ of the eigenvalue equations $\tilde{K}\bm{\alpha}=(N-1)\ell \bm{\alpha}$ in descending order of corresponding eigenvalues $\ell_i$ (which is equivalent to the descending order of variance).
Utilizing the first $q$ of these eigenvalues and eigenvectors, we construct the learned QoI map $Q=\left[Q_1,\dots,Q_q\right]^{\top}$ where 
\begin{equation}\label{eq:learned_QoI}
    Q_i\left(\cdot\right)=\sum_{j=1}^N\bm{\alpha}_j^{(i)}\tilde{\kernel}\left(\cdot,\bm{d}^{(j)}\right), \ 1\leq i\leq q.
\end{equation}

% The $N\times N$ Gram matrix, denoted by $K$, is defined componentwise as $K_{i,j}:=\kernel\left(\bm{d}^{(i)},\bm{d}^{(j)}\right)$ for $1\leq i,j\leq N$.

% Note that $Q$ is a function $Q:\mathcal{D}\rightarrow \mathbb{R}^N$, but if we introduce the model evaluation $\mathcal{M}$ that takes parameters in $\Lambda$ and maps them to evaluations in $\mathcal{D}$, i.e., $\mathcal{M}:\Lambda\rightarrow \mathcal{D}$, then we may define $Q:\Lambda \rightarrow \mathbb{R}^N$ as $$Q\left(\lambda\right)\coloneqq Q\left(\mathcal{M}\left(\lambda\right)\right).$$ The DCI diagnositc defined as the sample average of $$r\left(\lambda\right)=\frac{\pi_{\text{obs}}\left(Q\left(\lambda\right)\right)}{\pi_{\text{pred}}\left(Q\left(\lambda\right)\right)}$$ evaluated at the original samples for $\pi_{\text{pred}}$ is then calculated, and if the diagnositc deviates too far from 1, then $Q$ is reduced to $Q=\left[Q_1,\dots,Q_{N-1}\right]^{\top}$. This process proceeds until the diagnostic is within a tolerance of 1 or until only one component is kept. Let $m$ denote the resulting number of components kept, i.e., $Q=\left[Q_1,\dots,Q_m\right]^{\top}$ and $Q:\Lambda \rightarrow \mathbb{R}^m$ with $m\leq N$.

\subsection{Numerical Sufficiency Test for Filtered Data}\label{sec:how_much_data}

% {\bf Troy: outlining points to make. Need to flesh out later.}
% \begin{itemize}
%     \item The dimension of $\bm{\alpha}^{(i)}$ in~\eqref{eq:learned_QoI} only depends on the number of initial samples. It is entirely independent of the dimension of the space $\filteredspace$.
%     \item We therefore compare the QoI learned from different filtered data sets associated with the same set of initial samples in terms of these vectors since they are always the same dimension. 
%     \item As more filtered data are incorporated, we expect (and assume) that the QoI map should converge component-by-component. Let $\bm{\alpha}^{(i),m}$ denote the $i$th $\bm{\alpha}$ vector obtained from the $m$th refinement of filtering. We treat the $m-1$ filtered data set as defining the predictors $\bm{\alpha}^{(i),m-1}_j$ and the $m$ filtered data set as defining the response $\bm{\alpha}^{(i),m}_j$ for $1\leq j\leq N$ and we apply linear regression looking for a slope and $r^2$ values both close to $1$ to determine if sufficient filtered data are utilized to learn the components of the QoI map. We demonstrate this in the final illustrative example. 
% \end{itemize}

In the above analysis, the data vectors $\bm{d}^{(1)},\ldots,\bm{d}^{(N)}\in\filteredspace$ are associated with the $N$ iid parameter samples drawn from the initial distribution. 
From~\eqref{eq:learned_QoI}, we see that the $\bm{\alpha}^{(i)}\in\mathbb{R}^N$ for each $1\leq i\leq q$.
In other words, the vectors learned from the kPCA that define each component of the QoI map have dimension that is equal to the number of samples drawn from the initial distribution.
Consequently, the dimension of these vectors is invariant to the dimension of $\filteredspace$. 
This observation is key to developing a quantitative diagnostic for analyzing the sufficiency of filtered data.

Let $A$ and $B$ denote two potential sets of coordinates for which filtered data are constructed by evaluation of the learned filtered functions in Step 1 of LUQ. 
In the context of how the following ideas are applied, it is helpful to think of $B$ as a potential refinement of $A$.
For instance, if $A$ is defined by evaluating the filtered functions on an $n\times n$ grid of spatial coordinates, suppose $B$ is defined by evaluating the filtered functions on an $m\times m$ grid with $m>n$. 
Denote by $\filteredspace^{(A)}$ and $\filteredspace^{(B)}$, respectively, the associated potential filtered data spaces for these two sets of coordinates.
Finally, for a given component of the QoI vector, denote by $\bm{\alpha}^{(A)}$ and $\bm{\alpha}^{(B)}$, respectively, the associated learned $\alpha$ vectors from the kPCA analysis as given by~\eqref{eq:learned_QoI}. 

Viewing the components of $\bm{\alpha}^{(A)}$ as predictors for the components of $\bm{\alpha}^{(B)}$, we can use simple linear regression on these components to quantify what is learned from refining the filtered coordinate system. 
However, since the eigenvalues $\ell$ associated with the Gram matrix that are used to rescale the $\bm{\alpha}$ vectors may vary significantly from the one filtered coordinate system to another, we first normalize the $\bm{\alpha}$ vectors with respect to the standard $2$-norm in $\mathbb{R}^N$ before applying regression.
For a given QoI component learned from the two filtered coordinate systems $A$ and $B$, the filtered coordinates of $A$ are deemed sufficient if the associated slope and coefficient of determination are both near unity.
In other words, the additional filtered data obtained by the $B$ coordinate system do not significantly alter the QoI component to justify the additional computational burden of increasing the dimension.
Below, we illustrate the application of the concept in the context of evaluating sufficiency of filtered data for the example of randomly generated waves.
A future work will incorporate this perspective into a more comprehensive OED framework that evaluates optimal unfiltered data collection in conjunction with optimally learning the QoI from such data.

\subsection{Illustrative Example: Randomly Generated Waves (Part V)}\label{sec:example_V}

Revisiting the wave equation example from Part III in Section~\ref{sec:wave_method3}, recall that the original mesh over the spatial domain is a uniformly spaced $99\times 99$ grid at spacial locations $\left(0.05i,0.05j\right)$, $i,j=1,\dots,99$. 
% In this section, we apply the framework discussed in Section~\ref{sec:how_much_data} to explore how much spatial data is actually needed in order to learn the QoI map via kPCA.
We investigate potential filtered coordinates defined by sub-meshes of the computational mesh.
Specifically, we consider grid of sizes $4\times 4$, $9\times 9$, $19\times 19$, $49\times 49$, and finally the original $99\times 99$ mesh. 
While we focus our attention on uniformly spaced meshes that coincide with the original mesh for computational simplicity in filtering the predicted data, this is not necessary due to the filtering method allowing the data to be evaluated at any set of data coordinates.

\begin{figure}
\centering{%
\includegraphics[width=0.45\textwidth]{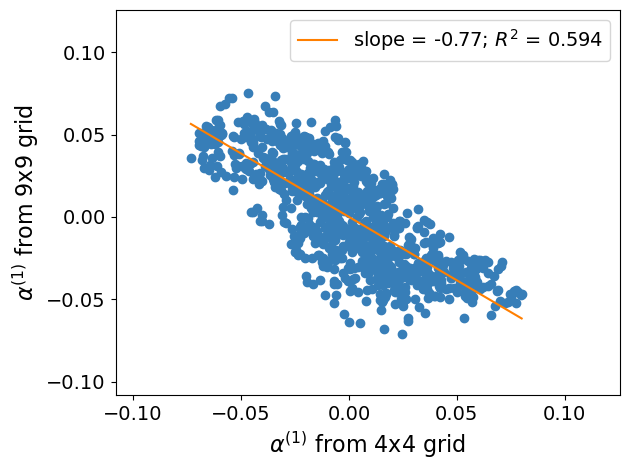}
\includegraphics[width=0.45\textwidth]{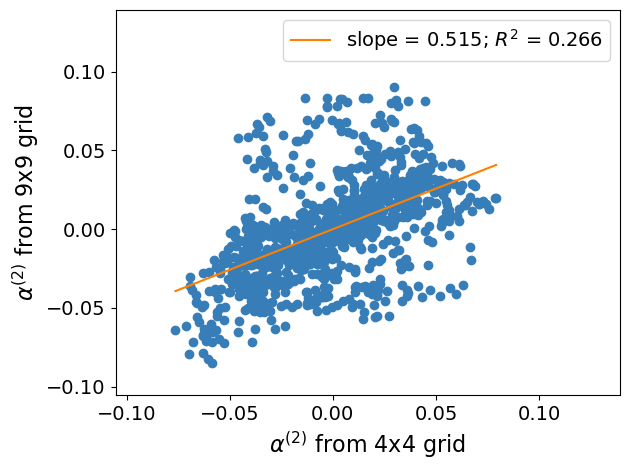}
% }\quad{%
% \includegraphics[width=0.45\textwidth]{graphics/9x9_19x19_1.png}
% \includegraphics[width=0.45\textwidth]{graphics/9x9_19x19_2.png}
% }\quad{%
% \includegraphics[width=0.45\textwidth]{graphics/19x19_49x49_1.png}
% \includegraphics[width=0.45\textwidth]{graphics/19x19_49x49_2.png}
}\quad{%
\includegraphics[width=0.45\textwidth]{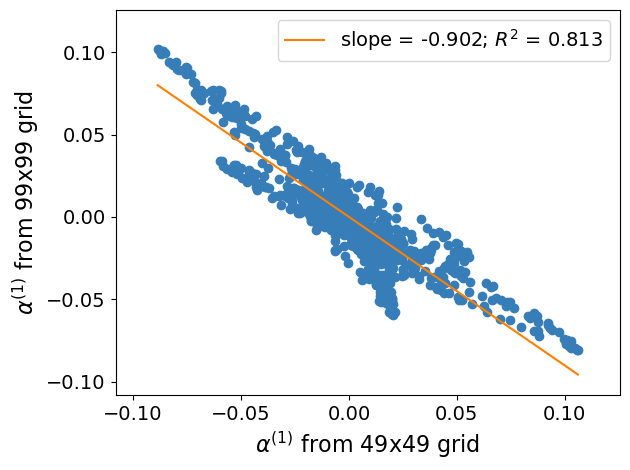}
\includegraphics[width=0.45\textwidth]{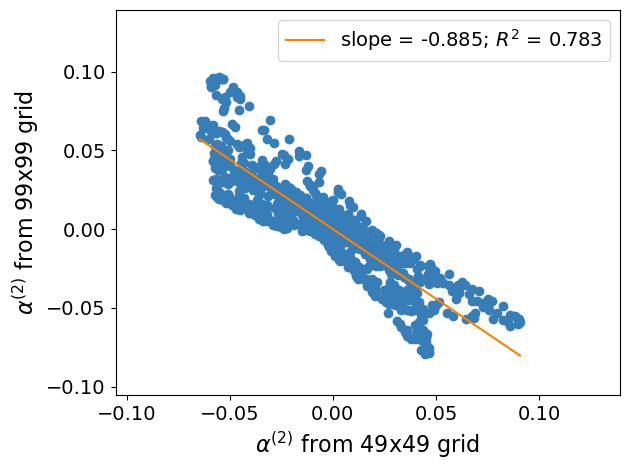}
}
\caption{Plots of normalized $\pmb{\alpha}$ vectors from performing kernel PCA on the 4x4 and 9x9 grids (top) and from 49x49 and 99x99 grids (bottom). For each row, the left plots are of $\pmb{\alpha}^{(1)}$ from each grid and the right plots are of $\pmb{\alpha}^{(2)}$ from each grid.}\label{fig:wave_partV_alphas}
\end{figure}

Starting with the two coarsest potential filtered coordinate sets, we compute the desired QoI maps, normalize the $\bm{\alpha}$ vectors for each component, and perform the regression analysis as described above in Section~\ref{sec:how_much_data} to quantitatively compare the coefficient of determination denoted by $R^2$ and the absolute value of the slope against values of unity. 
In the interest of space, we only show results for comparing the $4\times 4$ grid with the $9\times 9$ grid and comparing the $49\times 49$ grid with the $99\times 99$, which are shown in Figure \ref{fig:wave_partV_alphas}.
The interested reader can follow the instructions in~\ref{sec:software} to obtain all the data and scripts required to generate the intermediate comparisons.
Note that going from the $4\times 4$ grid (i.e., using $16$ data points) to the $9\times 9$ grid (i.e., using $81$ data points) produces significantly different QoI components with coefficients of determination 0.496 and 0.243 and slopes -0.705 and -0.493 for each component, respectively.
Note that the sign of the slopes is arbitrary since multiplying a given $\alpha$ obtained in the kPCA analysis by $-1$ does not alter the QoI learned from this kPCA.  
By comparison, once we use the $49\times 49$ grid (i.e.,using $2401$ data points) of potential filtered coordinates, we see that going to a $99\times 99$ grid (i.e., using $9801$ data points) produces minimal change in QoI components with coefficients of determination 0.813 and 0.783 and slopes -0.902 and -0.885 for each component, respectively. 
In other words, there is strong statistical evidence for using the $49\times 49$ filtered coordinate system to learn the QoI map since any further refinement of this mesh does not significantly alter the learned $\bm{\alpha}$ vectors defining the QoI components.  
% We therefore see that learning the QoI map via kPCA using 9801 data points from the 99x99 grid will produce nearly the same QoI map, up to scaling, as using 2401 data points from the 49x49 grid. 
This does not imply that using one of the coarser filtered data coordinate sets will provide a lower quality QoI map for DCI than using the data points from the $49\times 49$ grid, but it is clear that there is a limit to how much filtered data obtained at more data coordinates, either in space or time, can effect the kPCA components.
In fact, many of the updated densities shown in prior parts of this illustrative example are similar both qualitatively and quantitatively if we utilize the $9\times 9$ grid for the filtered coordinates, which the interested reader can easily verify by adjusting the filtered data coordinates variables within the provided data and scripts. 
At a practical level, the analysis discussed in this section is useful for establishing a numerical method for identifying an upper bound on the amount of potentially useful filtered data for learning the QoI. 

\section{Numerical Experiments}
\label{sec:numerical_experiments}

% \begin{table}
% \centering
% \begin{tabular}{lll}
% \toprule
% Experiment & Dimension & Parameters\\ \hline
% Random Profile and Material & 1D & $a_0 \sim \text{Beta}(2,6)$ (Young's Modulus),\\
%  & & $a_1 \sim \text{Beta}(3,4)$ (Profile)\\
% Trommel Screen & 2D & 
% $a_1 \sim \text{Beta}(2,6)$ (Left Section), \\
%  & & $a_2 \sim \text{Beta}(3,4)$ (Mid Section), \\
%  & & $a_3 \sim \text{Beta}(2,4)$ (Right Section)\\
% \bottomrule
% \end{tabular}
% \caption{Description of parameters used in the experiments.}\label{tbl:experiments}
% \end{table}

Our numerical examples concern shells of revolution.
Specifically, we seek to use displacement data from a population of shells to infer distributions of parameters defining material properties of the shells.
All structures are considered to be thin with
dimensionless thickness of $1/100$ and a Poisson number $\nu = 0.3$ approximating steel.
All realizations are solved using $hp$-FEM.
For details on the underlying
elasticity equations see~\ref{sec:shell_models}. 

\subsection{Shell of Revolution: Random Profile and Material}
%\subsection{Classification of Shell Types Based on Gaussian Curvature}
%
%{\bf Rewrite this from the perspective of a single shell type and its dependence on a parameter like Young's modulus?}

\begin{figure}
\centering
\subfloat[{Parabolic: $f(x) = 1$.}]{%
\includegraphics[width=0.3\textwidth]{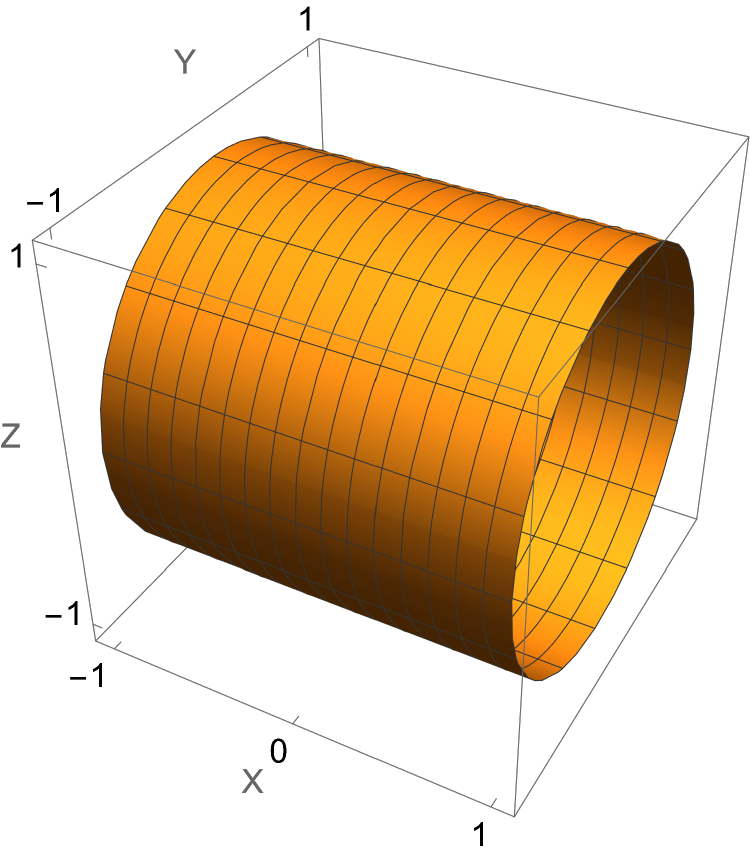}
}\quad
\subfloat[{Elliptic: $f(x) = 1 - x^2 / 4$.}]{%
\includegraphics[width=0.3\textwidth]{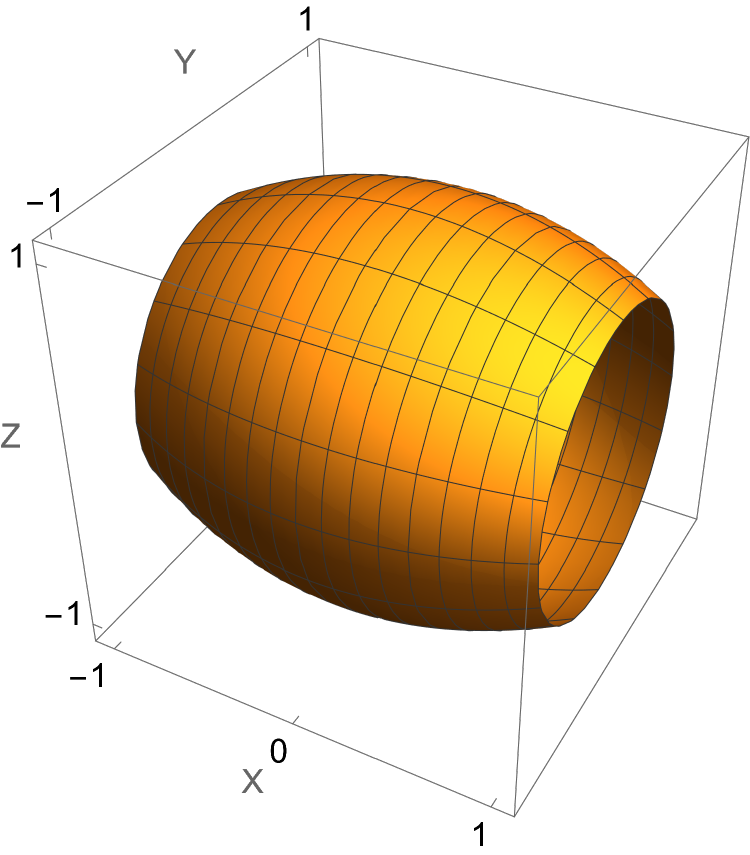}
}\quad
\subfloat[{Hyperbolic: $f(x) = 1 + x^2 / 4$.}]{%
\includegraphics[width=0.3\textwidth]{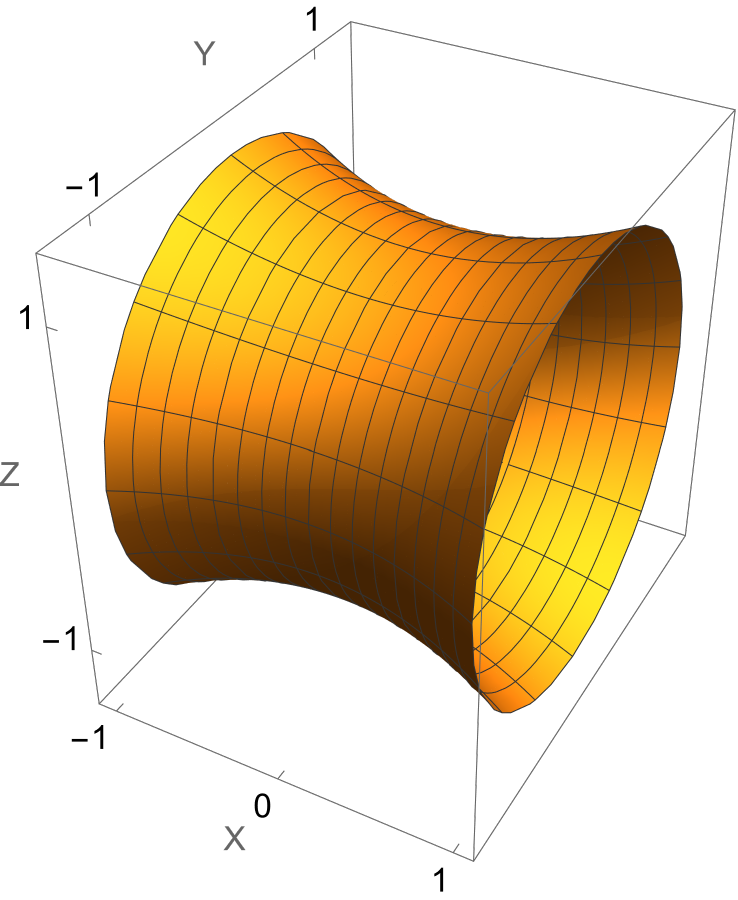}
}
\caption{Three shells of revolution: $x \in [-1,1]$.}\label{fig:shellgeometry}
\end{figure}

Mechanical characteristics of shells of revolution are strongly dependent on their shapes, or more precisely, types of midsurface geometry.
Given an interval $I$, we define a positive profile function $f(x) > 0$, $x \in I$, which revolves around the $x$-axis. 
In this example, we take $I=[-1,1]$.
The different kinds of shells of revolution formed in this way can be classified in terms of Gaussian curvature (see for instance \cite{doCarmo}) which can be described in terms of the derivatives of the profile function $f(x)$. 
Assuming that the second derivative has a constant sign over the whole interval $I$, one has
\begin{itemize}
\item \emph{Parabolic [Zero Gaussian curvature shells]} if
$f''(x) = 0$ (see left plot in Figure~\ref{fig:shellgeometry}), 

\item \emph{Elliptic [Positive Gaussian curvature shells]} if $f''(x) < 0$ (see mid plot in Figure~\ref{fig:shellgeometry}),

\item \emph{Hyperbolic [Negative Gaussian curvature shells]} if $f''(x) > 0$ (see right plot in Figure~\ref{fig:shellgeometry}).

\end{itemize}

%\begin{enumerate}
%%
%
%\item[1 --] \emph{Parabolic [Zero Gaussian curvature shells].} We have
%%
%\begin{equation}
%\label{parcyl}
%f''(x) = 0 , \qquad \forall\ x \in I \
%\end{equation}
%
%
%\item[2 --] \emph{Elliptic [Positive Gaussian curvature shells].} We have
%%
%\begin{equation}
%\label{ellcyl}
%f''(x) < 0 , \qquad \forall\ x \in I 
%\end{equation}
%
%
%\item[3 --] \emph{Hyperbolic [Negative Gaussian curvature shells].} We have
%%
%\begin{equation}
%\label{hypcyl}
%f''(x) > 0 , \qquad \forall\ x \in I 
%\end{equation}
%
%\end{enumerate}

In this example, both the Young's modulus $E = a_0$ and profile function $f(x)$ are considered uncertain.
We assume based on the types of materials used in the manufacturing as well as the machining of the shells that it is known that $a_0$ should be within $[0.8,1.2]$ (dimensionless) and $f(x)=1+a_1(x-1)(x+1)$ with $a_1$ varying within $[-0.2,0.2]$ to yield shells that vary from elliptic/positive Gaussian curvature to hyperbolic/negative Gaussian curvature for $x\in[-1,1]$. 
It follows that $\pspace=[0.8,1.2]\times[-0.2,0.2]$ in this example.

In the context of this example, it is useful to consider three classes of shapes defined as ``clearly elliptic'', ``almost parabolic with mild deformation'', or ``clearly hyperbolic'' that are associated, respectively, with $a_1$ belonging to the interval $[-0.2,-0.075]$, $(-0.075, 0.075)$, or $[0.075,0.2]$.

Loading is a periodic unit ``wind load'' acting on the tranverse direction.
In the notation of \ref{sec:shell_models}, this implies that the $w$-component is formally given by $f_w(x,y) = \cos y$.
This is intended to simulate the strong directional winds that wind turbine towers are subjected to in a wind farm. 
As further described in \ref{sec:shell_models}, in the reduced 1D formulation, the load  becomes $f_w(x) = 1$, with the angular wave number $k = 1$.

\subsubsection{Example highlights}

This experiment demonstrates several features of the LUQ and DCI frameworks not previously highlighted with the illustrative examples.
First, we demonstrate that updated distributions can reasonably approximate DG distributions with relatively few samples from both the DG and initial distributions. 
In doing this, we instantiate/re-use the same \texttt{LUQ} object on several distinct observation data sets.
This exploits a mathematical feature of the updated density obtained within the DCI framework where any observed density satisfying the predictability assumption immediately defines an updated density via direct substitution into the numerator of either~\eqref{eq:dc_density_r} or \eqref{eq:updated-cluster} once the predicted density is estimated for a given initial density.
In other words, it is trivial to solve multiple stochastic inverse problems for the same initial density once a predicted density is estimated. 
To properly exploit this fact, we designed the \texttt{LUQ} software to allow for the instantiation and saving of a \texttt{LUQ} object from initial/predicted sample data in cases where either observable data are not yet available or multiple observed data sets are to be analyzed. 
%and predicted density associated with a fixed set of initial samples that are subsequently utilized to filter and classify different observable data sets to produce distinct estimates of various DG distributions associated with potentially distinct observable data sets.
Additionally, to demonstrate the flexibility of the filtering step in the \texttt{LUQ} code, we assume that the observable and predicted data are both obtained via experiments with differing levels of signal-to-noise (SNR) ratios defined by the ratio of the variance of ``exact'' data (the signal) to variance in the noise and emphasize that it is unnecessary to specify the SNR value to perform the filtering as this step is agnostic to such details.

\subsubsection{DCI and LUQ Setup}

\begin{table}
\begin{center}
\begin{tabular}{|c|c|}\hline
Input & Details \\ \hline\hline 
DG distributions & \begin{tabular}{l|l} 
 $a_0\sim\text{Beta}(2,6)$ on $[0.8,1.2]$  & $a_1\sim\text{Beta}(3,4)$ on $[-0.2,0.2]$ 
\end{tabular} \\ \hline
Init.~distributions & \begin{tabular}{l|l}
$a_0\sim\mathcal{U}([0.8,1.2])$ &  $a_1\sim\mathcal{U}([-0.2,0.2])$ \end{tabular} \\ \hline 
DG sample sets & 3 sets of 75 samples from DG distributions \\ \hline
Init.~sample sets & 1 set of 200 samples from init.~distributions \\ \hline
Predicted data labels & $0$ if $a_1\in [-0.2, -0.075]$, $1$ if $a_1\in (-0.075, 0.075)$, and $2$ if $a_1\in [0.075,0.2]$ \\ \hline
Observable data & 901 noisy data with SNR of 5 for each DG sample \\ \hline
Predicted data & 901 noisy data with SNR of 10 for each init.~sample\\ \hline
\# Filtered data & 60 filtered data for each observation of prediction sample \\ \hline
\end{tabular}
\end{center}
\caption{Summary of setup details for DCI and LUQ frameworks in shell of revolution example.}\label{tab:ex-1-inputs}
\end{table}

\begin{figure}
\centering
\includegraphics[width=0.45\textwidth]{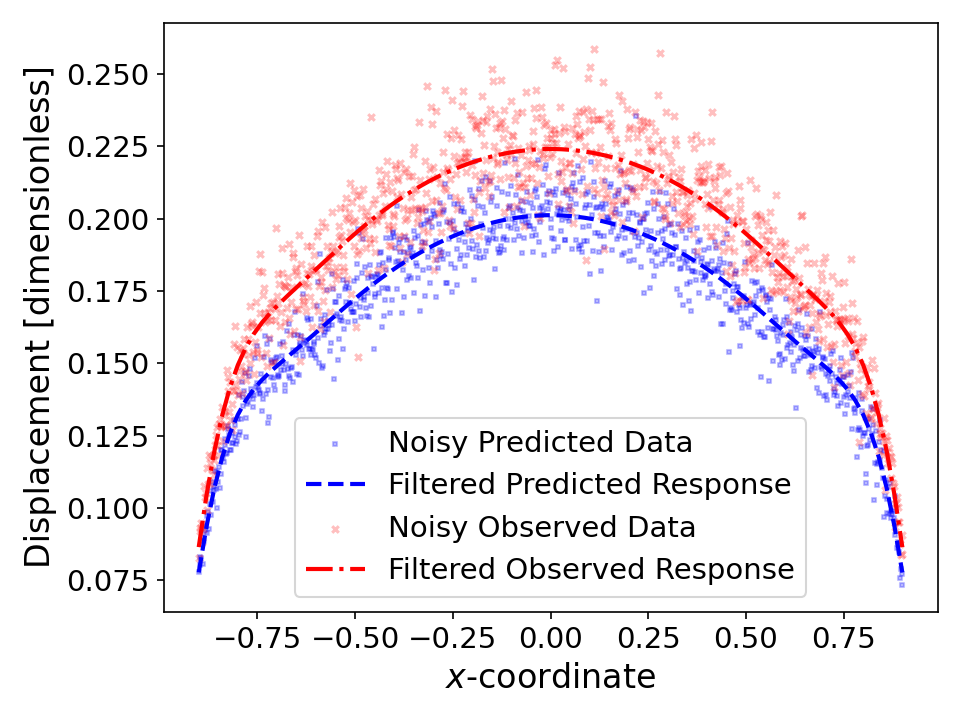}
\caption{Noisy data for shell of revolution example with different SNR levels for predicted and observed data. The filtering step in \texttt{LUQ} can flexibly approximate the underlying response even if each sample within a data set has its own unique SNR level.}\label{fig:ex1_filtered_data}
\end{figure}

Table~\ref{tab:ex-1-inputs} summarizes the relevant setup details for both the DCI problem and LUQ frameworks in this example.
We assume (noisy) displacement data are obtained via an electronic ``strip'' equipped with 901 equally spaced sensors that is placed on the opposite side of each shell from the direction of forcing for $x\in[-0.9,0.9]$.
The noise is assumed to be additive Gaussian and iid at each measurement location. 
The goal is to learn the distributions on $a_0$ and $a_1$ from this noisy observed displacement data.
For the purposes of this example, the DG distributions are taken to be $a_0\sim \text{Beta}(2,6)$ scaled and shifted to $[0.8,1.2]$ and $a_1\sim\text{Beta}(3,4)$ scaled and shifted to $[-0.2,0.2]$ while the initial distributions on $a_0$ and $a_1$ are assumed uniform across $\pspace$.
To simulate the difference between controlled experiments versus field studies, we assume samples taken from the initial distribution have measurement data producing a SNR of ten while samples taken from the observed distribution have measurement data producing a SNR of five. 
In other words, the ``field'' data is twice as noisy as the ``predicted'' data.
Figure~\ref{fig:ex1_filtered_data} illustrates representative results for the different levels of noise in these types of data and the estimated filtered responses from \texttt{LUQ} utilizing splines.
See~\ref{sec:software} for details on obtaining the full scripts and datasets utilized for creating this and the remaining figures in this example.

We use 200 initial samples and three different sets of 75 samples from the DG distribution to demonstrate the simplicity of solving multiple stochastic inverse problems and the \texttt{LUQ} code as mentioned in the example highlights.
The initial samples are drawn from a set of $\expnumber{1}{4}$ simulations associated with the initial distribution while the three sets of 75 samples are drawn from a set of $\expnumber{3}{3}$ simulations associated with the DG distribution, and the interested reader can change both the number of samples and how they are drawn from these sets in the provided code.
The provided code also considers two different LUQ setups that utilize either two or three classes based on the shape characteristics (i.e., the $a_1$ values).
In the interest of space, we only present the results for the three classes case since the diagnostics obtained from a pilot study involving the noise-free data indicated better performance than in the two classes case (these pilot studies are included in the code).
As mentioned above, we define the three classes as being for $a_1$ in the intervals $[-0.2,-0.075]$, $(-0.075, 0.075)$, and $[0.075,0.2]$, which roughly correspond to the shapes being either ``clearly elliptic'', ``almost parabolic with mild deformation'', or ``clearly hyperbolic''. 
The interested reader can alter the ranges of these intervals or the number of classes defined and use the diagnostics discussed within this example and provided within the code as guides for assessing the quality of their choices. 

\subsubsection{Results}

\begin{table}
\begin{center}
\begin{tabular}{|c|c|}\hline
Output & Details \\ \hline\hline 
Diagnostic: $\mathbb{E}_\text{init}(r(Q(\lambda)))\approx 1$ &
$0.934$ (DG set 1) $\quad$ $0.937$ (DG set 2) $\quad$ $0.928$ (DG set 3)
\\ \hline
TV between $a_0$ update/DG marginals & 0.086 (DG set 1) $\quad$ 0.096 (DG set 2) $\quad$ 0.099 (DG set 3) \\ \hline
TV between $a_1$ update/DG marginals & 0.083 (DG set 1) $\quad$ 0.028 (DG set 2) $\quad$ 0.060 (DG set 3) \\ \hline
\end{tabular}
\end{center}
\caption{Summary of output details for DCI in shell of revolution example.}\label{tab:ex-1-outputs}
\end{table}

Table~\ref{tab:ex-1-outputs} summarizes the relevant outputs details for DCI in this example.
To obtain these results requires the filtering of data, training of the SVM classifier, and learning of QoI in the LUQ framework.
To accomplish those steps, the predicted dataset generated from the initial samples is used to instantiate a \texttt{LUQ} object where the predicted data are subsequently filtered, utilized to learn/train an SVM classifier for the three-clusters associated with $a_1$ values, and learn the 2-dimensional QoI maps.
While 901 noisy data are utilized in the filtering, a total of 60 filtered data are produced in the filtering step to learn the QoI maps.
The sufficiency of 60 filtered data is verified in the code by comparing to 30 filtered data in the manner described in Section~\ref{sec:how_much_data} and demonstrated on the illustrative example in Section~\ref{sec:example_V}.
In this case, we observe both slopes and $R^2$ values from the regression analysis are all above 0.95 for each component of the QoI map across all three clusters indicating sufficiency in the number of filtered data utilized to learn the QoI.

\begin{figure}
\centering
\subfloat[{Data-generating set 1.}]{%
\includegraphics[width=0.3\textwidth]{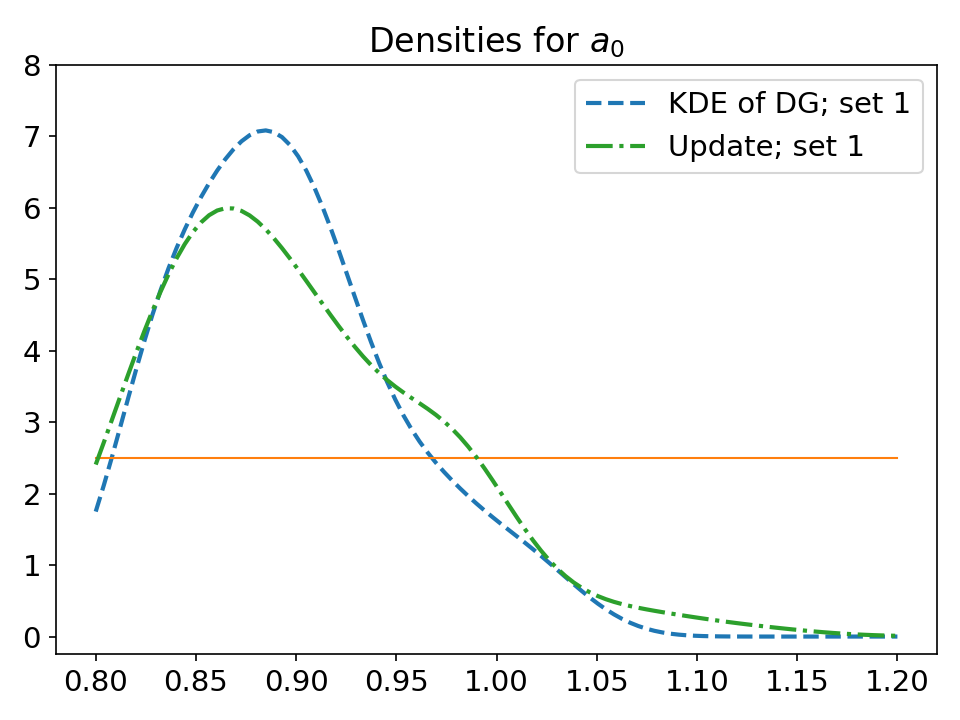}
}\quad
\subfloat[{Data-generating set 2.}]{%
\includegraphics[width=0.3\textwidth]{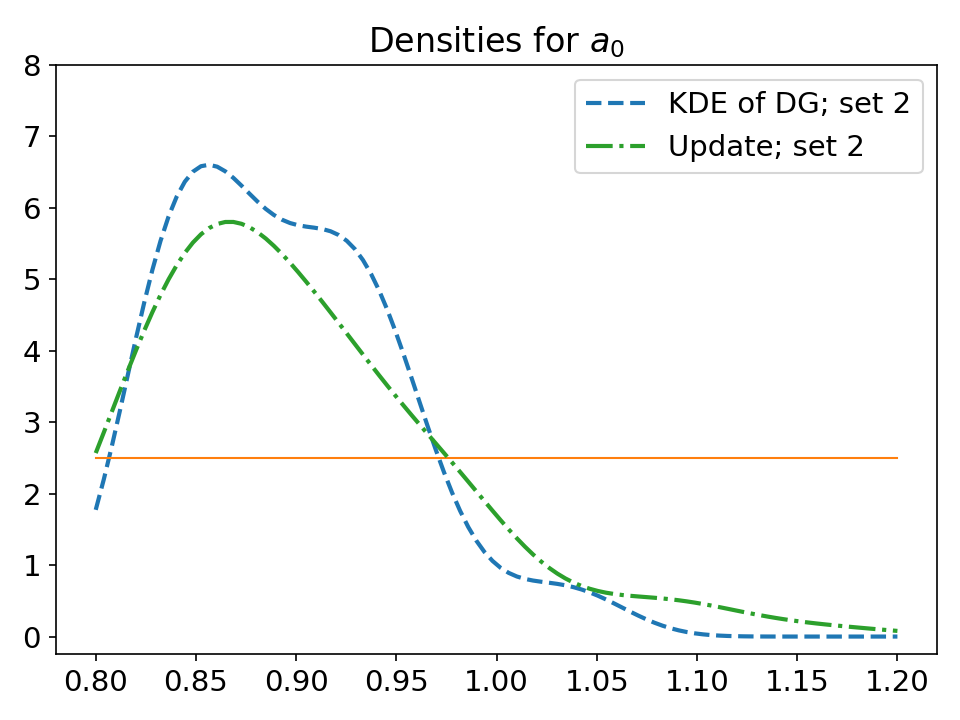}
}
\quad
\subfloat[{Data-generating set 3.}]{%
\includegraphics[width=0.3\textwidth]{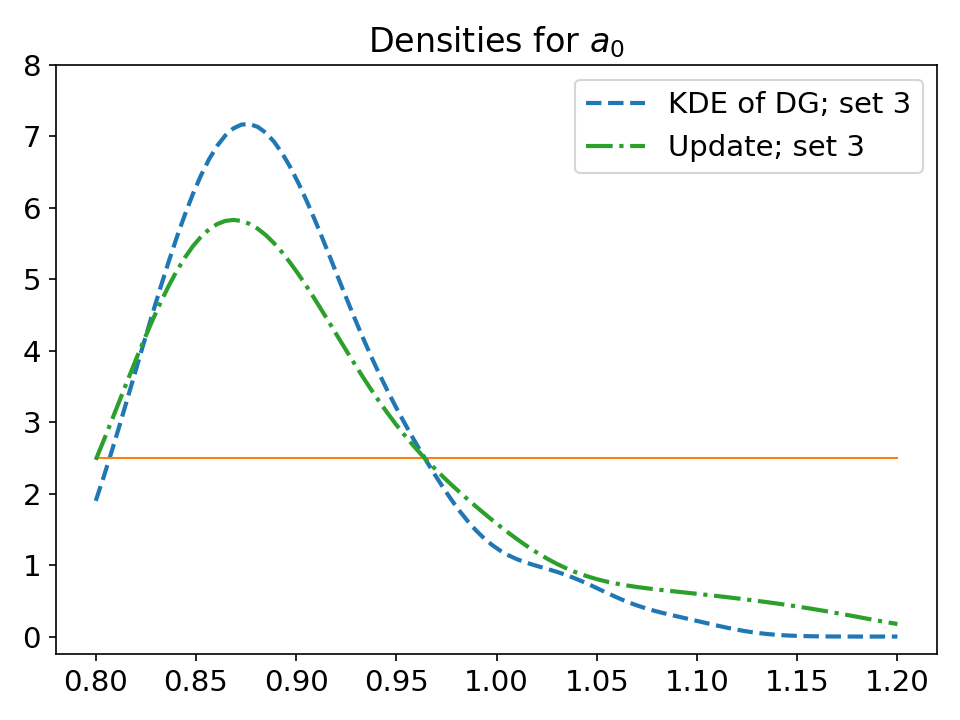}
}
\caption{Marginal densities for $a_0$ for shell of revolution example from three different data-generating sets of 75 samples. The horizontal line indicates the uniform initial density, the dashed line indicates the KDE of the DG distribution computed from the samples associated with the given observation data set, and the dashed-dotted line indicates the KDE of the updated density.}\label{fig:ex1_a0_dens}
\end{figure}

\begin{figure}
\centering
\subfloat[{Data-generating set 1.}]{%
\includegraphics[width=0.3\textwidth]{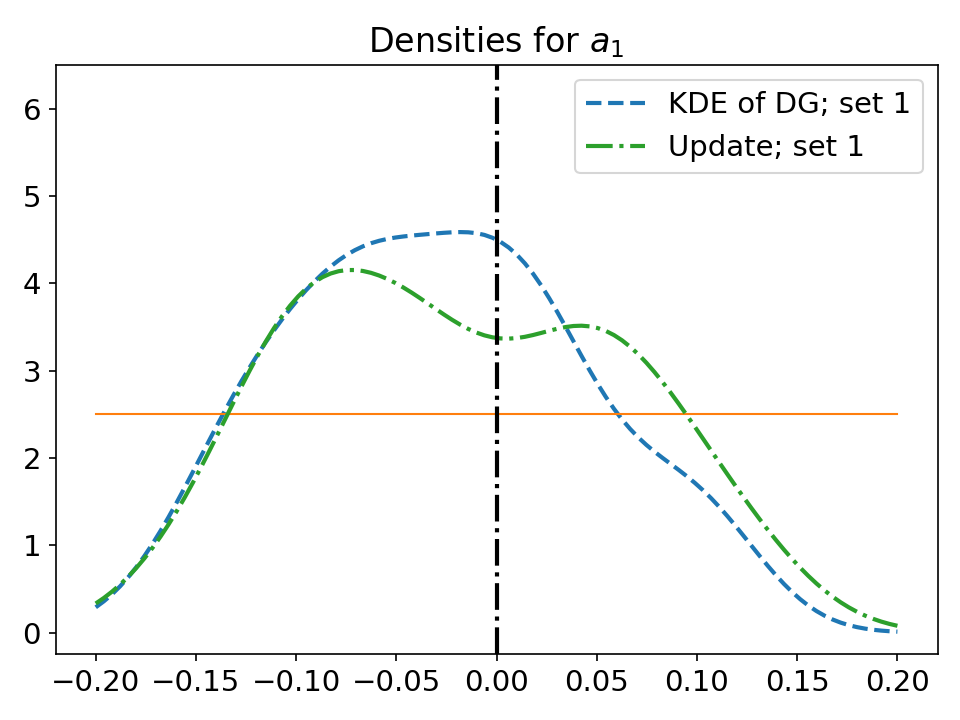}
}\quad
\subfloat[{Data-generating set 2.}]{%
\includegraphics[width=0.3\textwidth]{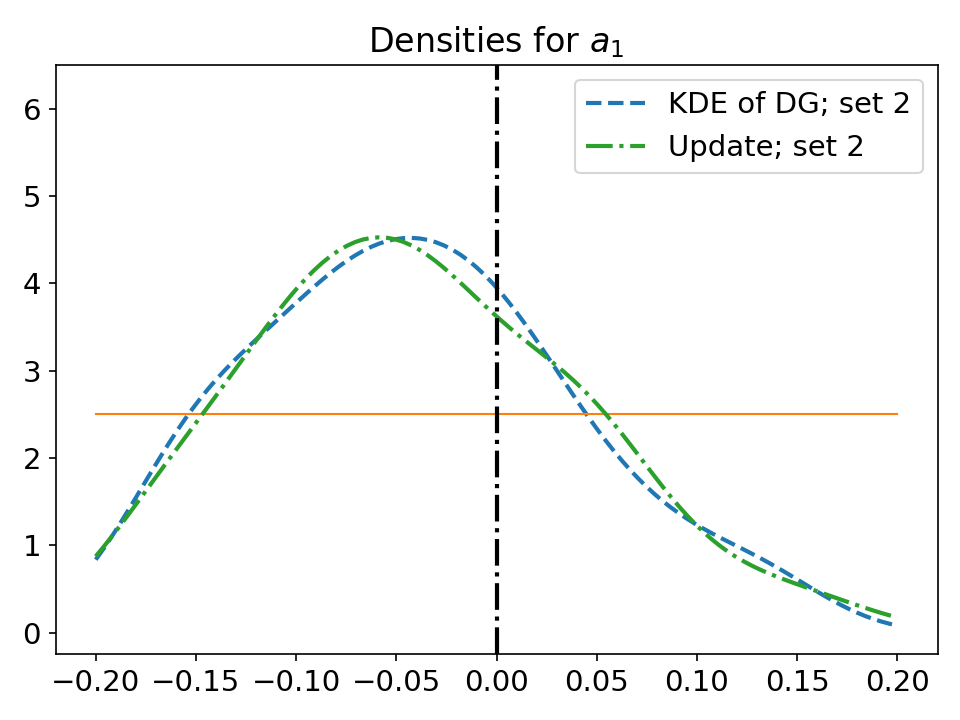}
}
\quad
\subfloat[{Data-generating set 3.}]{%
\includegraphics[width=0.3\textwidth]{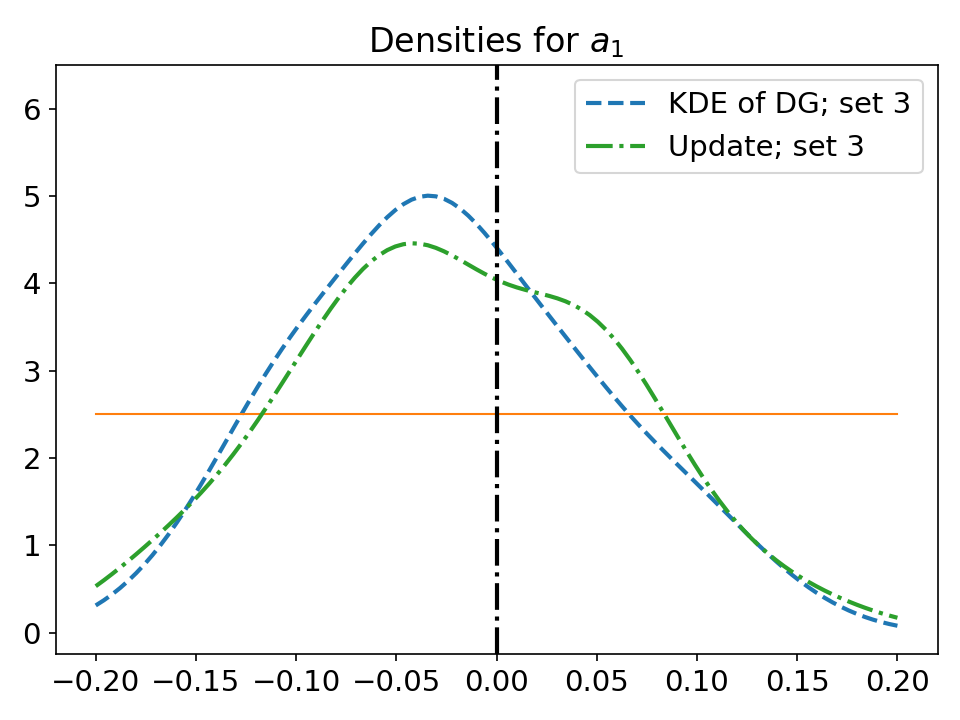}
}
\caption{Marginal densities for $a_1$ for shell of revolution example from three different data-generating sets of 75 samples. The horizontal line indicates the uniform initial density, the dashed line indicates the KDE of the DG distribution computed from the samples associated with the given observation data set, and the dashed-dotted line indicates the KDE of the updated density. The vertical dashed-dotted line at $a_1=0$ indicates where the shells change from elliptic to hyperbolic in shape.}\label{fig:ex1_a1_dens}
\end{figure}

The \texttt{LUQ} object is then separately applied to each of the three distinct observation sets. 
Given the clustering and relatively small number of samples utilized in the DCI solution overall and the clusters specifically, we compute a ``weighted mean diagnostic'' defined by the overall sample average of the updated density given in~\eqref{eq:updated-cluster} to assess the predictability assumption and quality of construction of the updated density as a whole.
For observation sets 1, 2, and 3, the associated diagnostics are $0.934, 0.937$ and $0.928$, which are all within 10\% of unity and indicate that both the predictability assumption is satisfied and the updated density is well approximated.

Having numerically verified the predictability assumption, we now evaluate how well the densities for the individual parameters are estimated in each of the observation sets.
Figures~\ref{fig:ex1_a0_dens} and~\ref{fig:ex1_a1_dens} show the associated uniform initial densities (horizontal line), KDE approximation of the DG distribution constructed from the samples associated with the observation set (dashed line), and KDE of the updated density (dashed-dotted line) for $a_0$ and $a_1$, respectively.
For the $a_1$ density plots, we also provide a vertical dashed-dotted line at $a_1=0$ as a visual indicator of where the shapes change from elliptic (on the left) to hyperbolic (on the right). 
It is worth emphasizing that there is clear variation between the KDEs of the DG distributions within the $a_0$ and $a_1$ plots due to the relatively small sample size (75 samples) utilized for each of the observation sets.
The marginal density plots qualitatively indicate that each of the updated densities, which are all computed by re-weighting the {\em same} 200 initial samples in every plot, are reasonable approximations to the associated KDE of the DG distribution that generated the observable data.
Moreover, the TV metrics are all under 0.1 for each marginal of each parameter across all sets of observable data, which as the illustrative examples have demonstrated, is a value we expect for a ``good'' KDE approximation of a given distribution. 
Specifically, the TV metrics (rounded to three decimal places) between the KDEs of the updated and DG densities for sets 1, 2, and 3 for $a_0$ are $0.086$, $0.096$, and $0.099$ and for $a_1$ are $0.083$, $0.028$, and $0.060$, respectfully. 

% \begin{figure}
% \centering
% \subfloat[{Marginal for $a_0$.}]{%
% \includegraphics[width=0.45\textwidth]{graphics/Ex-1-1e4init-3e3obs-noise-free-a0-dens.png}
% }\quad
% \subfloat[{Marginal for $a_1$.}]{%
% \includegraphics[width=0.45\textwidth]{graphics/Ex-1-1e4init-3e3obs-noise-free-a1-dens.png}
% }
% \caption{Marginal densities for shell of revolution example with no noise in data using 1e4 initial samples and 3e3 data-generating samples.}\label{fig:ex1_noise_free}
% \end{figure}

% {\bf Placeholder results below. Results need to be redone. Emphasize sensitivity of measurements? If no clustering done (going from elliptic to hyperbolic but analyzing all at once) then the 2nd and 3rd PCs are ``tight together'' and we can iterate through all three QoI to get a decent result. However, we should try to do some clustering with pre-determined labels.}

% \begin{figure}[htbp]
% \centering
% \subfloat[{Marginal for $a_0$.}]{%
% \includegraphics[width=0.45\textwidth]{graphics/prelimresults-a0-pdfs.png}
% }\quad
% \subfloat[{Marginal for $a_1$.}]{%
% \includegraphics[width=0.45\textwidth]{graphics/prelimresults-a1-pdfs.png}
% }
% \caption{Marginal densities for shell of revolution example.}\label{fig:shell_marginals}
% \end{figure}	

\subsection{Trommel Screen: MultiPanel Construction with Material Variation}
The second example is a trommel screen with three sections each with different regular penetration patterns but constant 
35\% hole coverage as shown in the introduction in Figure~\ref{fig:trommelA}.
The computational domain is a square $\Omega = [-\pi,\pi]\times[0,2\pi]$.
The boundaries at $x = \pm \pi$ are clamped. The bottom ($y=0$) and 
the top ($y=2\pi$) are periodic. All holes are free, that is, 
there are no kinematical constraints.
As with the previous example, we refer to \ref{sec:shell_models} for more details related to the underlying elasticity equations and solution of the model. 

In this example, due to different penetration patterns in each of the three sections and variations in the manufacturing, the Young's modulus $E$ is considered uncertain in each section. 
We let $a_0$, $a_1$, and $a_2$ denote, respectively, the values of Young's modulus in the left-, middle-, and right-sections of each trommel screen.
We assume that due to the larger holes in the the left-section that the material is weakest with $a_0$ taking values in $[0.576, 0.704]$.
The values of $a_1$ and $a_2$ are assumed, respectively, to take values between $[0.72, 0.88]$ and $[0.9, 1.1]$.
These intervals represent 10\% variation around nominal design values of $0.6, 0.8,$ and $1.0$, respectively. 

Loading is assumed to be a linear combination of
concentrated loads acting 
on specific points $(x_k,y_k)$.
The model for the concentrated load is simply
\begin{equation}\label{eq:cload}
	f(x,y) = \exp(-100((x - x_k)^2 + (y - y_k)^2)).
\end{equation}
We set $y_k = \pi$ (constant), 
and let $x_k = -2\pi/3 + (k-1) 2\pi/3$, $k=1,2,3$. 
Large debris that is separated by the left-section of the trommel will have impact points associated with $k=1$ values.
Moderate sized debris separated by the middle-section of the trommel will have impacts points associated with $k=2$ values.
Finally, small debris separated by the right-section of the trommel will have impact points associated with $k=3$. 

\subsubsection{Example highlights}

Here, we assume that three separate experiments are carried out across a random sample of trommel screens in controlled experiments (with known parameter values) and a different number of trommel screens in field studies (with unknown parameter values).
In each experiment, either ``large'', ``moderate'', or ``small'' debris are separated by the trommel with sensors recording the displacement from debris impact inside the trommel to finite precision, which represents noise in terms of ``loss of information/precision'' in the data.
In each case, these datasets are available synchronously meaning that the relative order of obtaining the spatial displacement data is considered unknown.
We subsequently utilize this data within the LUQ framework by bypassing the filtering step for each dataset (since noise is due to truncation in precision), learn a distinct number of QoI for each experiment, and then iterate through the DCI procedure to arrive at reasonable updated estimates of the DG distributions for each parameter.
Diagnostics and statistical analysis of data guided the iteration order and number of QoI learned for each experiment, but the interested reader can alter all of these choices and observe the subsequent results and diagnostics. 

\subsubsection{DCI and LUQ Setup}

\begin{table}
\begin{center}
\begin{tabular}{|c|c|}\hline
Input & Details \\ \hline\hline 
DG distribution $(a_0)$ & $a_0\sim\text{Beta}(2,6)$ on $[0.576,0.705]$  \\ \hline
DG distribution $(a_1)$ & $a_1\sim\text{Beta}(3,4)$ on $[0.72, 0.88]$ \\ \hline
DG Distribution $(a_2)$ & $a_2\sim\text{Beta}(2,4)$ on $[0.9, 1.1]$ \\ \hline
Init.~distrs. & Uniform on respective domains \\ \hline 
DG sample set & 50 samples from DG distributions \\ \hline
Init.~sample set & 500 samples from init.~distributions \\ \hline
Observable \& Predicted data & Sensor data (see Figure~\ref{fig:trommelB}) truncated to 2 digits each sample \\ \hline
\end{tabular}
\end{center}
\caption{Summary of setup details for DCI and LUQ frameworks in trommel example.}\label{tab:ex2-inputs}
\end{table}

Table~\ref{tab:ex2-inputs} summarizes the relevant setup details for both the DCI and LUQ frameworks in this example. 
We assume that the measurements for each experiment are truncated at two decimal places for both the observed and predicted data.
The goal is to learn the distributions of $a_0, a_1$, and $a_2$ from this finite precision data obtained from the separate experiments.
For the purposes of this example, the DG distributions are $a_0\sim\text{Beta}(2,6)$, $a_1\sim\text{Beta}(3,4)$, and $a_2\sim\text{Beta}(2,4)$ appropriately scaled and shifted to their respective domains.
The initial distributions are assumed to be uniform for each parameter.

We use $500$ initial samples and $50$ DG samples to generate the predicted and observed datasets.
To three decimal places, the variances in the ``full precision'' predicted data associated with the large, moderate, and small debris are $0.101$, $0.181$, and $0.015$, which represents significant loss of information from the finite precision measurements associated with truncating all data to two decimal places.
Since no filtering nor classification is required, a \texttt{LUQ} object is instantiated for the finite-precision predicted data associated with each experiment to learn a 2-dimensional, 3-dimensional, and 1-dimensional QoI map for the large, moderate, and small debris experiments, respectively. 
The observed data from each experiment are subsequently transformed into samples associated with their learned QoI maps.

\subsubsection{Results}

\begin{table}
\begin{center}
\begin{tabular}{|c|c|}\hline
Output & Details \\ \hline\hline 
Diagnostic: $\mathbb{E}_\text{init}(r(Q(\lambda)))\approx 1$ &
$0.974$ (iter 1, 2 QoI) $\quad$ $0.938$ (iter 2, 3 QoI) $\quad$ $0.937$ (iter 3, 1 QoI)
\\ \hline
TV between $a_0$ update/DG marginals & 0.090 (iter 1, 2 QoI) $\quad$ 0.064 (iter 2, 3 QoI) $\quad$ 0.064 (iter 3, 1 QoI) \\ \hline
TV between $a_1$ update/DG marginals & 0.062 (iter 1, 2 QoI) $\quad$ 0.018 (iter 2, 3 QoI) $\quad$ 0.020 (iter 3, 1 QoI)  \\ \hline
TV between $a_2$ update/DG marginals & 0.313 (iter 1, 2 QoI) $\quad$ 0.119 (iter 2, 3 QoI) $\quad$ 0.090 (iter 3, 1 QoI)  \\ \hline
\end{tabular}
\end{center}
\caption{Summary of output details for DCI in shell of revolution example.}\label{tab:ex2-outputs}
\end{table}

\begin{figure}
\centering
\subfloat[{$a_0$} densities]{%
\includegraphics[width=0.3\textwidth]{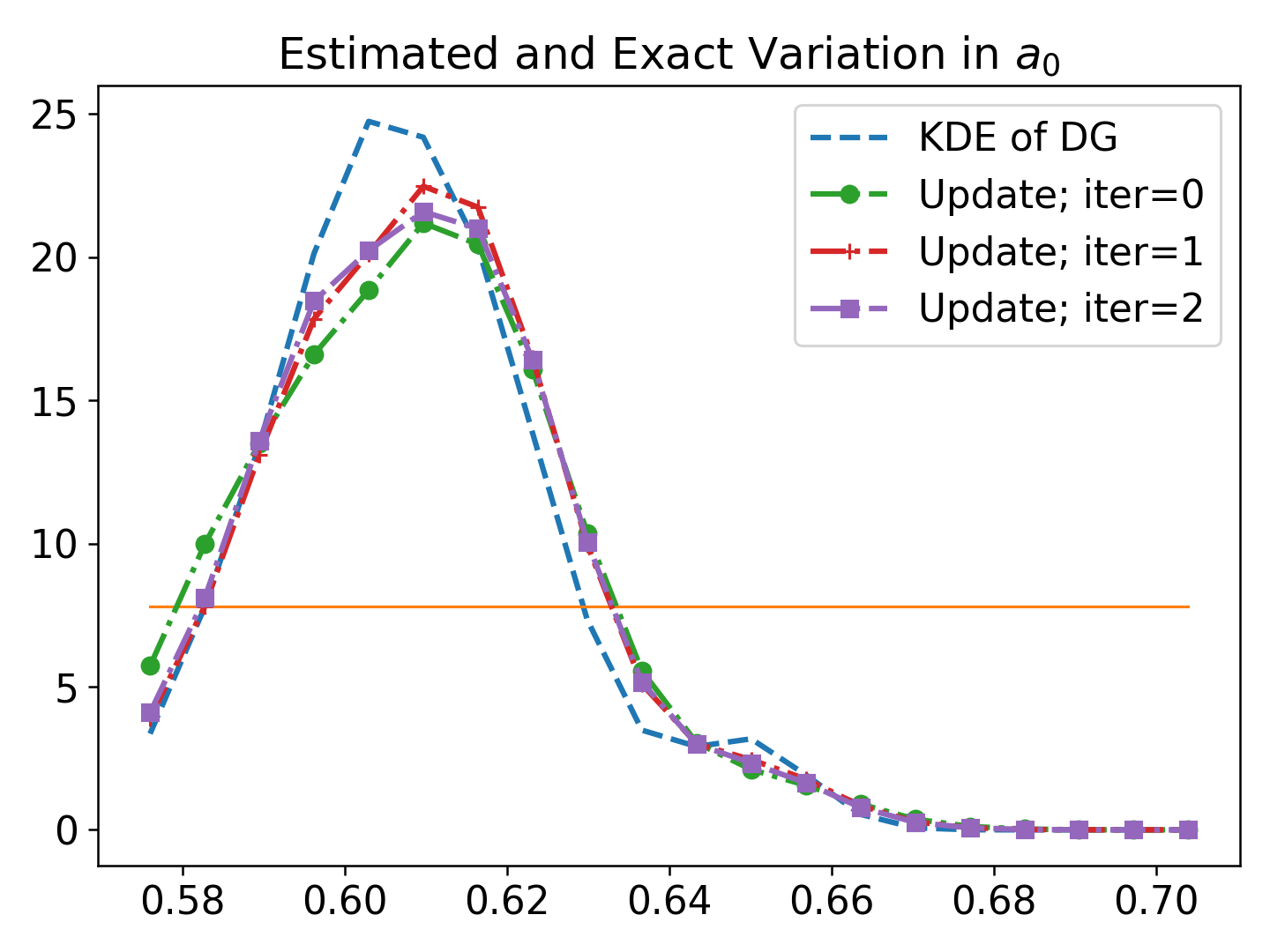}
}\quad
\subfloat[{$a_1$}]{%
\includegraphics[width=0.3\textwidth]{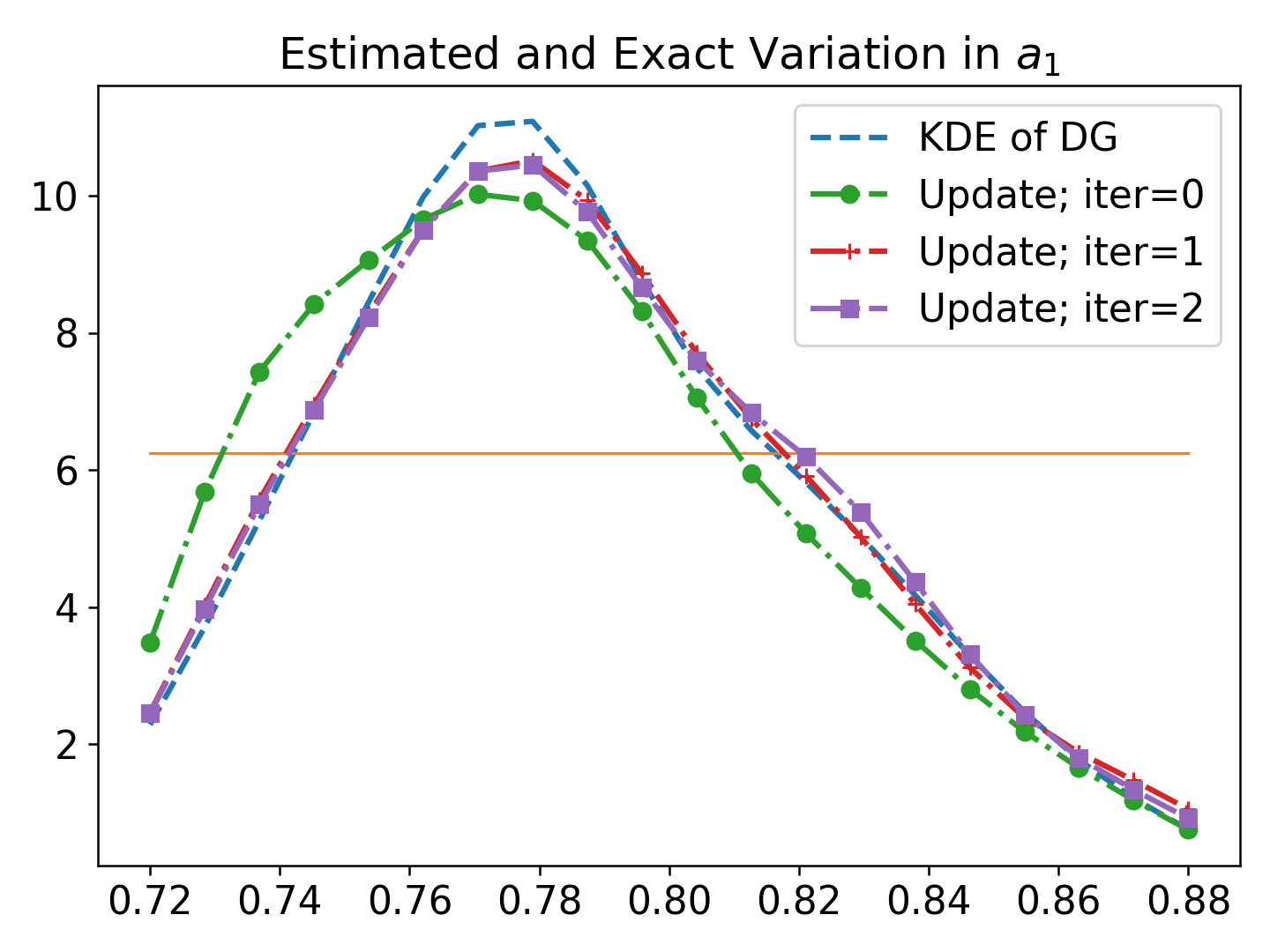}
}
\quad
\subfloat[{$a_2$}]{%
\includegraphics[width=0.3\textwidth]{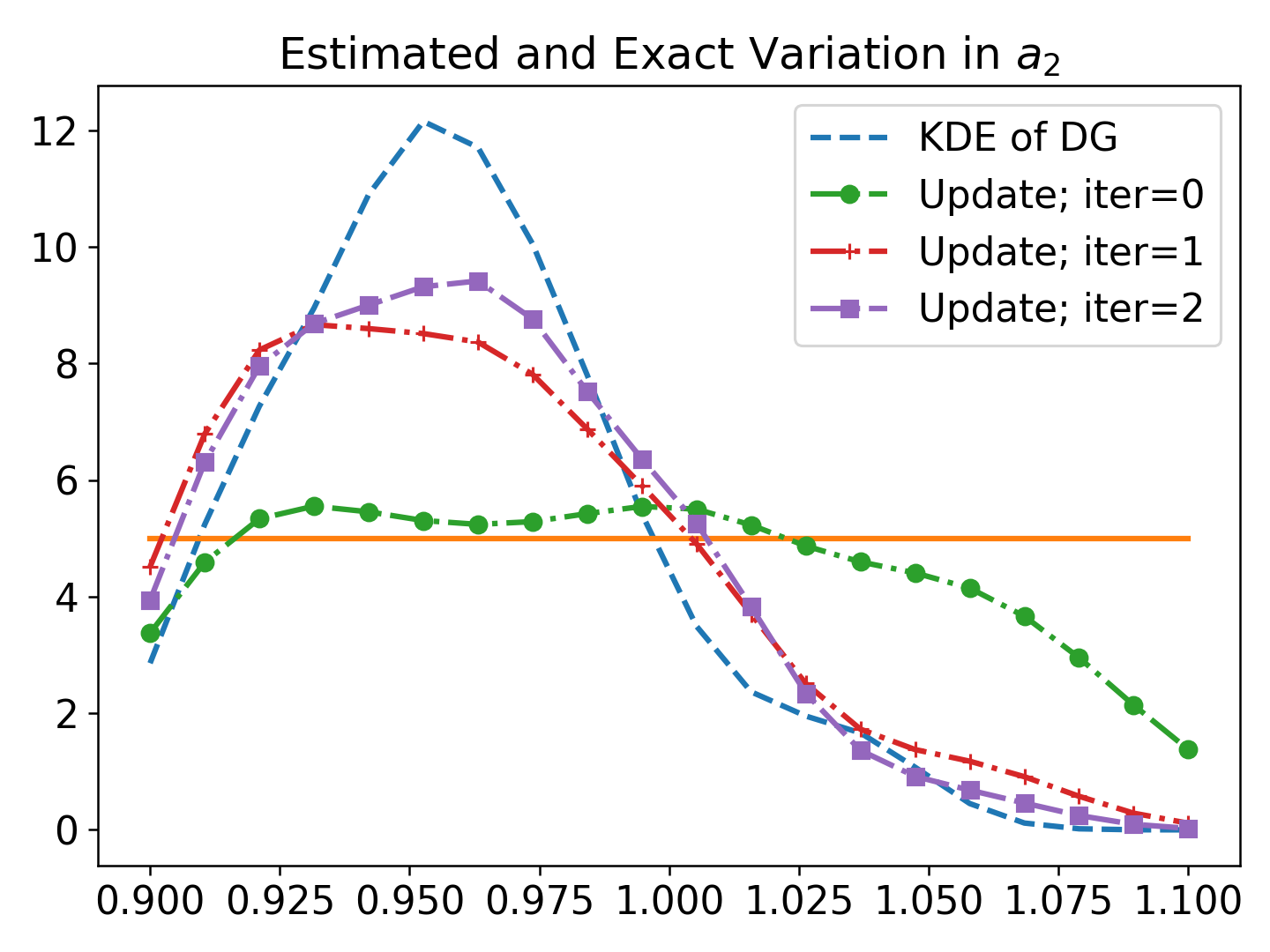}
}
\caption{Marginal densities for the trommel screen parameters associated with left- ($a_0$), middle- ($a_1$), and right-sections ($a_2)$ of the screen shown in Figure~\ref{fig:trommelA}.
The updated marginals are iteratively obtained by a sequence of experiments involving ``large'', ``moderate'', and ``small'' debris that cause impacts at various points on the trommel screen shown in Figure~\ref{fig:trommelB} and are filtered by the left-, middle-, and right-sections, respectively, of the trommel.}\label{fig:ex2_densities}
\end{figure}

We iterate through the large, moderate, and small debris experiments to sequentially update the initial distributions.
This is similar to the temporal iteration described in Section~\ref{sec:DCI_iterative} where an updated distribution obtained from one experiment serves as the initial distribution for the next experiment.
The results are summarized in Figure~\ref{fig:ex2_densities} and Table~\ref{tab:ex2-outputs}.
We observe that the first iteration, for which we learn a 2-dimensional QoI map and is associated with the large debris filtered through the left-section of the trommel appears to mostly learn the $a_0$ DG distribution while also learning some aspects of the $a_1$ DG distribution and providing no significant update to the $a_2$ initial distribution.
A general improvement across all three parameters is observed as we iterate across the experiments.
By the final experiment, the TV metric between the KDE estimate of the updated marginal and DG marginal for each parameter is under $0.1$ indicating a reasonable approximation across all parameters.
However, it is clearly the case that the updated marginal for the $a_2$ parameter is the worst estimate of its associated DG distribution.
This is easily explained as the $a_2$ parameter is associated with the ``stiffest'' part of the trommel and the final experiment that involves impact loads on this part of the trommel involves the small debris that produces the smallest variation in the data and is most affected by the finite precision of the measurements.
The interested reader can simply improve the precision in the predicted and observed data in the provided scripts to see how this improves these estimates (and allows for more QoI to be learned and utilized for futher improvement as well).
This is related to a future work exploring optimal experimental design questions within the LUQ and DCI frameworks. 

\section{Conclusions}
\label{sec:conclusions}

Uncertainties in the computational models of engineered or physical systems governed by principles of mechanics served as the motivation for this work.
We consider both aleatoric (i.e., irreducible) and epistemic (i.e., reducible) sources of uncertainty.
To quantify these uncertainties, we combine two frameworks.
The Learning Uncertain Quantities (LUQ) framework defines a formal three-step machine-learning enabled process for transforming datasets into samples of a learned Quantities of Interest (QoI) map to enable solution of a stochastic inverse problem posed in a data-consistent (DC) framework. 

In this work, we develop a robust filtering step in LUQ that can learn the most useful quantitative information present in spatio-temporal datasets.
Reproducing Kernel Hilbert Space (RKHS) theory is leveraged to mathematically analyze the learned QoI map and develop a quantitative sufficiency test for evaluating the filtered data.
An iterative DC-based inversion scheme is also developed and considered including a new diagnostic to identify both the quality and impact of inversion at each iteration.

The practical utility of these contributions is demonstrated through the application of three problems: an illustrative example involving random excitation of waves, the simulated manufacturing of thin shells of revolution with random shape profiles and uncertain stiffness, and the simulated manufacturing of trommel screens that are utilized in industrial applications involving the mechanical separation of materials of different sizes.

\section{Acknowledgments}
T.~Butler and T.~Roper's work is supported by the National Science Foundation under Grant No.~DMS-2208460.
T.~Butler's work is also supported by NSF IR/D program, while working at National Science Foundation. 
However, any opinion, finding, and conclusions or recommendations expressed in this material are those of the author and do not necessarily reflect the views of the National Science Foundation.

\bibliographystyle{elsarticle-num-names}
\bibliography{ShellSFEM, bibliography}

\appendix

\section{Software and Reproducibility of Results}\label{sec:software}

Version 2.0.0 of the Python package \texttt{LUQ}  \cite{pyLUQ2.0.0} was developed as part of this work and is available at https://github.com/CU-Denver-UQ/LUQ.
The software requires \texttt{numpy} ($\geq$1.17.0 and $<$1.23.0), \texttt{scipy} ($\geq$1.7.0 and $<$1.8), \texttt{matplotlib} ($\geq$3.1.0), and \texttt{scikit-learn} ($>=$1.0.2).
The example data, numerics, and figures presented in this paper can be reproduced using the Jupyter notebooks and/or Python scripts found within the \texttt{LUQ} GitHub repository.
The illustrative example is found in the example folder named ``wave'' and is separated by parts to align with the presentation in this manuscript. 
The shell examples are found in the example folder named ``shells'' and require the package \texttt{mat73} for reading the data into memory.
See the README file on github for more information.
We note that the examples are reproducible machine-to-machine, but due to back-end multi-threading mechanisms within the used packages, there may exist minor deviations between machines and in comparison to those reported in the manuscript.

\section{Shell Models}
\label{sec:shell_models}

\subsection{Shell Geometry}
\label{sec:shellgeometry}

Our reference structures are thin shells of revolution.
They can formally be characterized as domains in $\mathbb{R}^3$ of type
\begin{equation}
	\Omega = \left\{ \mathbf{x} + z \mathbf{n(x)} \ | \ \mathbf{x} \in \Gamma, -d/2 < z < d/2\right\},
\end{equation}
where $d$ is the (constant) thickness of the shell, 
$\Gamma$ is a (mid)surface of revolution, 
and $\mathbf{n(x)}$ is the unit normal to $\Gamma$.
The geometry model is based on principal curvature coordinates, where only four parameters
are needed to specify
the curvature and metric on $\Gamma$, namely
the radii of principal curvature $R_1$, $R_2$, and the Lam\'e parameters,
$A_1$, $A_2$, which relate coordinates changes to arc lengths. 
The displacement vector field of the midsurface $\mathbf{u} = \{u,v,w\}$ 
is the set of projections to directions
\begin{equation}
\mathbf{e}_1 = \frac{1}{A_1}\frac{\partial \Psi}{\partial x_1},\quad \mathbf{e}_2 = \frac{1}{A_2}\frac{\partial \Psi}{\partial x_2},\quad \mathbf{e}_3 = \mathbf{e}_1\times \mathbf{e}_2,
\end{equation}
where $\Psi(x_1,x_2)$ is a suitable parametrization of the surface of revolution,
$\mathbf{e}_1, \mathbf{e}_2$ are the unit tangent vectors along the principal curvature lines, and
$\mathbf{e}_3$ is the unit normal.
In other words 
\[
\mathbf{u} = u\,\mathbf{e}_1+ v\,\mathbf{e}_2+ w\,\mathbf{e}_3.
\]

\subsubsection{Profile functions and parametrization}
When a plane curve is rotated (in three dimensions) about a line in the plane of the curve,
it sweeps out a surface of revolution. Consider a plane curve rotating about the $x$-axis, 
the so-called profile function in the $xy$-plane, $y = f(x)$, which induces a surface $\Gamma_f$.

Let 
$I=[\alpha,\beta]\subset \mathbb{R}$ be a bounded closed interval, and let  
$f(x): \: I \rightarrow \mathbb{R}^+$ be a regular function.
The shell midsurface $\Gamma_f$ is parameterized by means of the mapping
\begin{equation}\label{map1}
\begin{aligned}
& \boldsymbol{\Psi}_f \ : \ I\times[0,2\pi] \ \longrightarrow \ \mathbb{R}^3 \\
& \boldsymbol{\Psi}_f(x_1,x_2) = (x_1, f(x_1)\cos{x_2} , f(x_1)\sin{x_2})
\ .
\end{aligned}
\end{equation}
For $\Gamma_f$ we have
\begin{equation}
A_1(x) = \sqrt{1+[f '(x)]^2}, \quad A_2(x) = f (x), \quad
R_1(x) = - \frac{A_1(x)^3}{f ''(x)}, \quad R_2(x) = A_1(x) A_2(x).
\end{equation}

%Let 
%$J=[\alpha,\beta]\subset \mathbb{R}$ be a bounded closed interval with $\alpha > 0$, and let  
%$g(x): \: J \rightarrow \mathbb{R}$ be a regular function.
%In this case the shell midsurface $\Gamma_g$ is parametrised by means of the mapping
%
%\begin{equation}\label{map2}
%\begin{aligned}
%& \boldsymbol{\Psi}_g \ : \ J\times[0,2\pi] \ \longrightarrow \ \mathbb{R}^3 \\
%& \boldsymbol{\Psi}_g(x_1,x_2) = (x_1 \cos{x_2}, g(x_1), x_1 \sin{x_2})
%\ .
%\end{aligned}
%\end{equation}
%For $\Gamma_g$ we have
%\begin{equation}
%A_1(x) = \sqrt{1+[g '(x)]^2}, \quad A_2(x) = x,
%\end{equation}
%and
%\begin{equation}
%R_1(x) = \frac{A_1(x)^3}{g ''(x)}, \quad R_2(x) = A_1(x) A_2(x).
%\end{equation}
%

\subsection{Two-Dimensional Model}
The shell defined as $\Omega \subset \mathbb{R}^3$ above is a
three-dimensional body for which linear elasticity
could be considered ``exact'' for small deformations. Classical
dimension reduction models are reasonably accurate for these shells \cite{pms}.
We consider a model allowing for continuous solutions,
namely the Reissner-Mindlin model (often attributed to Naghdi).
The displacement field $\mathbf{u}$ has two additional components $\theta$, $\psi$, 
that is, it has five components $u$, $v$, $w$, $\theta$, $\psi$, each of which
is a function of two variables on the mid-surface of the shell.
The first two components represent the tangential displacements of the
mid-surface, $w$ is the transverse deflection, and whereas $\theta$, $\psi$ are dimensionless
rotations. In this section we assume that the shell
consists of homogeneous isotropic material with Young modulus $E$ and
Poisson ratio $\nu$. 

The total energy of the shell has the following expression in our dimension reduction model:
\begin{equation}\label{eqn:totalenergy}
\mathcal{F}(\ub) = \frac{1}{2} D (a(\ub,\ub)+d^2\,b(\ub,\ub))-q(\ub),
\end{equation}
where $D = E\,d/(12(1-\nu^2))$ is a scaling factor (see for instance \cite{np2}), $q$ is the external load potential, 
and $a(\ub,\ub)$ and $b(\ub,\ub)$ represent the portions of total deformation energy 
that are stored in membrane and transverse shear deformations and bending deformations, 
respectively.
The latter quadratic forms are independent of $d$ and defined as
\begin{eqnarray}
a(\ub,\ub) &=& a_m(\ub,\ub) + a_s(\ub,\ub) \nonumber\\
&=& 12 \int_{\w}
    \Bigl[ \nu (\beta_{11}(\ub)+\beta_{22}(\ub))^2 
+ (1-\nu) \sum_{i,j=1}^2 \beta_{ij}(\ub)^2\Bigr]\ A_1 A_2 \ d\gamma + \nonumber\\
& & 6(1-\nu)\int_{\w}
    \Bigl[(\rho_1(\ub)^2 + \rho_2(\ub))^2\Bigr]
    \ A_1 A_2 \ d\gamma, \\
b(\ub,\ub) &=& \int_{\w} \Bigl[\nu (\kappa_{11}(\ub)+\kappa_{22}(\ub))^2 
+(1-\nu) \sum_{i,j=1}^2 \kappa_{ij}(\ub)^2 \Bigr] \ A_1 A_2 \ d\gamma,
\end{eqnarray}
where $\beta_{ij}$, $\rho_i$, and $\kappa_{ij}$ stand for the membrane, transverse shear, and
bending strains, respectively. The strain-displacement relations are
linear and involve at most first derivatives of the displacement
components and hence are suitable for standard finite element methods. 
%The principal curvature coordinates, where there are only four parameters,
%the radii of principal curvature $R_1$, $R_2$, and the so-called Lam\'e parameters,
%$A_1$, $A_2$, which relate coordinate changes to arc lengths, are needed to specify
%the curvature and the metric on $\Gamma$. Here
%\begin{equation}
%A_1(x) = \sqrt{1+[f '(x)]^2}, \quad A_2(x) = f (x),
%\end{equation}
%and
%\begin{equation}
%R_1(x) = - \frac{A_1(x)^3}{f ''(x)}, \quad R_2(x) = A_1(x) A_2(x).
%\end{equation}

%\begin{remark}
%  In the following, we shall omit the constant factor $D$ from
%  the energy expressions. Consequently, all displacements can be
%  considered to be scaled with a factor $D^{-1}$ and energies
%  with a factor $D^{-2}$.	
%\end{remark}

Using the identities above, the bending, membrane, and shear strains~\cite{malinen}, 
$\kappa_{ij}, \beta_{ij},$ and $\rho_{i}$, respectively, can be written as 
\begin{eqnarray*}
\kappa_{11} &=& \frac{1}{A_1}\frac{\D \theta}{\D x}
         +  \frac{\psi}{A_1 A_2}\frac{\D A_1}{\D y},  \\
\kappa_{22} &=& \frac{1}{A_2}\frac{\D \psi}{\D y}
         +  \frac{\theta}{A_1 A_2}\frac{\D A_2}{\D x},  \\
\kappa_{12} &=& \kappa_{21}
         =  \frac{1}{2} \left[
            \frac{1}{A_1}\frac{\D \psi}{\D x}
         +  \frac{1}{A_2}\frac{\D \theta}{\D y}
         -  \frac{\theta}{A_1 A_2}\frac{\D A_1}{\D y}
         -  \frac{\psi}{A_1 A_2}\frac{\D A_2}{\D x} \right. \\
     & & -  \frac{1}{R_1}\left(\frac{1}{A_2}\frac{\D u}{\D y}
         -  \frac{v}{A_1 A_2}\frac{\D A_2}{\D x}\right)  \\
     & & - \left. \frac{1}{R_2}\left(\frac{1}{A_1}\frac{\D v}{\D x}
         -  \frac{u}{A_1 A_2}\frac{\D A_1}{\D y}\right)
        \right], 
\end{eqnarray*}

\begin{eqnarray*}
\beta_{11} &=& \frac{1}{A_1} \frac{\D u}{\D x}
          + \frac{v}{A_1 A_2} \frac{\D A_1}{\D y}
          + \frac{w}{R_1},  \\
\beta_{22} &=& \frac{1}{A_2} \frac{\D v}{\D y}
          + \frac{u}{A_1 A_2} \frac{\D A_2}{\D x}
          + \frac{w}{R_2},  \\
\beta_{12} &=& \beta_{21}
         =  \frac{1}{2} \left(
            \frac{1}{A_1} \frac{\D v}{\D x}
          + \frac{1}{A_2} \frac{\D u}{\D y}
          - \frac{u}{A_1 A_2} \frac{\D A_1}{\D y}
          - \frac{v}{A_1 A_2} \frac{\D A_2}{\D x}
          \right),
\end{eqnarray*}
\begin{eqnarray*}
\rho_{1}  &=& \frac{1}{A_1} \frac{\D w}{\D x}
          - \frac{u}{R_1} -\theta, \\
\rho_{2}  &=& \frac{1}{A_2} \frac{\D w}{\D y}
          - \frac{v}{R_2} -\psi.
\end{eqnarray*}
\begin{remark}
  When the shell parametrisations defined above
  are used, all terms of the form
  $\partial A_i / \partial y$ are identically zero.
\end{remark}

The energy norm $||| \cdot |||$ is defined in a natural way in terms
of the deformation energy:
\begin{equation}
\E(\ub) := |||\ub|||^2 = a(\ub,\ub)+d^2\,b(\ub,\ub).
\end{equation}

Similarly for bending, membrane, and shear energies:
\begin{equation}
B(\ub) := d^2\, b (\ub,\ub), \quad
M(\ub) := a_m (\ub,\ub), \quad
S(\ub) := a_s (\ub,\ub).
\end{equation}

The load potential has the form
\[
q(\mathbf{v}) = \int_{\w} \mathbf{f}(x,y)\,\cdot \mathbf{v}\,  A_1 A_2 \ dx\, dy.
\]
We are interested in problems where 
the load acts in the transverse
direction of the shell surface, i.e., $\mathbf{f}(x,y) =
[0,0,f_w(x,y),0,0]^T$. It can be shown that if for the load $\mathbf{f} \in
[L^2(\w)]^5$ holds, the variational problem has a
unique weak solution $\ub \in [H^1(\w)]^5$. The corresponding result
is true in the finite dimensional case, when the finite element method is employed.

The existence of the asymptotic displacement field corresponding 
to the limit $t \to 0$ is possible if a suitable scaling of 
the load amplitude is done in the limit process.
For discussion on details, see Malinen and Pitk\"aranta \cite{malinen2}. 

\subsection{One-dimensional Model}\label{sec:1dmodel}
The shell model can be further reduced to a one-dimensional one, if
the load is periodic in the angular direction and the Young modulus is constant in the angular direction,
that is, $E= E(x)$.
Due to the periodic structure
one can choose between two periodic transverse loadings acting on the $w$-component, 
either $(f_1)_w(x,y) = F(x) \cos(k \, y)$ or
$(f_2)_w(x,y) = F(x) \sin(k \, y)$, 
$k=0,1,2,\ldots$, and then the corresponding solution fields
$\mathbf{u}(x,y)$ are either
\begin{equation}\label{eq:ansatz}
\mathbf{u}_1(x,y) = \left(
  \begin{array}{c}
  u(x) \cos(k \, y) \\
  v(x) \sin(k \, y) \\
  w(x) \cos(k \, y) \\
  \theta(x) \cos(k \, y) \\
  \psi(x) \sin(k \, y) \\
  \end{array}
\right) \text{ or }
\mathbf{u}_2(x,y) = \left(
  \begin{array}{c}
  u(x) \sin(k \, y) \\
  v(x) \cos(k \, y) \\
  w(x) \sin(k \, y) \\
  \theta(x) \sin(k \, y) \\
  \psi(x) \cos(k \, y) \\
  \end{array}
\right).
\end{equation}
\begin{remark}
  The two ansatz simply represent the trigonometric basis in the angular direction.
\end{remark}
Using the ansatz $\mathbf{u}_1(x,y)$ above the energies are defined over an interval $I$.
Here the three energies are defined separately, with $B$, $M$, $S$, denoting
bending, membrane, and shear, respectively.
\begin{eqnarray}
d^2 \A_{B}^{1D}(\u,\u)
&=& d^2 \int_I E(x) \Bigl[\nu (\k_{11}(\u)+\k_{22}(\u))^2
+(1-\nu) \sum_{i,j=1}^2 \k_{ij}(\u)^2 \Bigr] \ A_1 A_2 \ dx , \label{eqn:taipumaenergia1D} \\
\A_{M}^{1D}(\u,\u) &=& 12 \int_I E(x)
    \Bigl[ \nu (\b_{11}(\u)+\b_{22}(\u))^2 + (1-\nu) \sum_{i,j=1}^2 \b_{ij}(\u)^2\Bigr]
    \ A_1 A_2 \ dx, \label{eqn:kalvoenergia1D}\\
\A_{S}^{1D}(\u,\u) &=&
    6(1-\nu)\int_I E(x)
    \Bigl[(\r_1(\u)^2     + \r_2(\u))^2\Bigr]
    \ A_1 A_2 \ dx \label{eqn:leikkausenergia1D}.
\end{eqnarray}
The strains can be written in terms of
the wave number $k$:
\begin{eqnarray*}
\k_{11} &=& \frac{1}{A_1}\frac{\D \theta}{\D x},  \\
\k_{22} &=& \frac{k \, \psi}{A_2}
         +  \frac{\theta}{A_1 A_2}\frac{\D A_2}{\D x},  \\
\k_{12} &=& \frac{1}{2} [
            \frac{1}{A_1}\frac{\D \psi}{\D x}
         -  \frac{k \, \theta}{A_2}
         -  \frac{\psi}{A_1 A_2}\frac{\D A_2}{\D x}  \\
     & & +  \frac{1}{R_1}(\frac{k \, u}{A_2}
         +  \frac{v}{A_1 A_2}\frac{\D A_2}{\D x})
         -  \frac{1}{R_2}\frac{1}{A_1}\frac{\D v}{\D x}
        ] = \k_{21},  \\
     & & \\
\b_{11} &=& \frac{1}{A_1} \frac{\D u}{\D x}
          + \frac{w}{R_1},  \\
\b_{22} &=& \frac{k \, v}{A_2}
          + \frac{u}{A_1 A_2} \frac{\D A_2}{\D x}
          + \frac{w}{R_2},  \\
\b_{12} &=& \frac{1}{2} (
            \frac{1}{A_1} \frac{\D v}{\D x}
          - \frac{k \, u}{A_2}
          - \frac{v}{A_1 A_2} \frac{\D A_2}{\D x}
        ) = \b_{21}, \\
        & & \\
\r_{1}  &=& \frac{1}{A_1} \frac{\D w}{\D x}
          - \frac{u}{R_1} -\theta, \\
\r_{2}  &=& - \frac{k \, w}{A_2}
          - \frac{v}{R_2} -\psi.
\end{eqnarray*}
The load potential becomes
\[
\Q^{1D}(\v) = \int_{\alpha}^{\beta} F(x) \, w \,A_1 A_2 \, dx.
\]
Notice that we have omitted  the coefficient $\pi$ ($2 \pi$, when $k=0$) from the load potential and the energy definitions
(\ref{eqn:taipumaenergia1D})--(\ref{eqn:leikkausenergia1D}).
This has to be taken into account whenever one- and two-dimensional
energies are compared.

%\cite{*}
\end{document}